\documentclass[Afour,sageh,times]{sagej}

\usepackage{moreverb,url}
\usepackage[colorlinks,bookmarksopen,bookmarksnumbered,citecolor=red,urlcolor=red]{hyperref}
\usepackage{verbatim}
\usepackage{amsmath}

\usepackage[utf8]{inputenc}
\usepackage{graphicx}
\usepackage{subfigure}
\usepackage{float}
\usepackage{bm}
\usepackage{amssymb}
\usepackage{threeparttable}
\usepackage{multirow}
\usepackage{booktabs}
\usepackage{algorithm}
\usepackage{algorithmic}
\usepackage{tabularray}

\usepackage{graphics} 
\usepackage{epsfig} 
\usepackage[T1]{fontenc}
\usepackage{color}
\usepackage[table]{xcolor}
\UseTblrLibrary{booktabs}

\newcommand\BibTeX{{\rmfamily B\kern-.05em \textsc{i\kern-.025em b}\kern-.08em
T\kern-.1667em\lower.7ex\hbox{E}\kern-.125emX}}

\def\volumeyear{2024}

\begin{document}

\runninghead{Chen et al.}

\title{Upper-Limb Rehabilitation with a Dual-Mode Individualized Exoskeleton Robot: A Generative-Model-Based Solution}

\author{Yu Chen\affilnum{1}, Shu Miao\affilnum{1}, Jing Ye\affilnum{2}, Gong Chen\affilnum{2}, Jianghua Cheng\affilnum{3}, Ketao Du\affilnum{3}, and Xiang Li\affilnum{1}}

\affiliation{\affilnum{1}Department of Automation, Tsinghua University, China
\\
\affilnum{2}Shenzhen MileBot Robotics Co., Ltd, China\\
\affilnum{3}Department of Rehabilitation, South China Hospital, Medical School, Shenzhen University, China
}

\corrauth{Xiang Li,
Department of Automation, Central Main Building, Tsinghua University, Beijing, China 100084}
\email{xiangli@tsinghua.edu.cn}

\begin{abstract}
Several upper-limb exoskeleton robots have been developed for stroke rehabilitation, but their rather low level of individualized assistance typically limits their effectiveness and practicability.
Individualized assistance involves an upper-limb exoskeleton robot continuously assessing feedback from a stroke patient and then meticulously adjusting interaction forces to suit specific conditions and online changes.
This paper describes the development of a new upper-limb exoskeleton robot with a novel online generative capability that allows it to provide individualized assistance to support the rehabilitation training of stroke patients.
Specifically, the upper-limb exoskeleton robot exploits generative models to customize the fine and fit trajectory for the patient, as medical conditions, responses, and comfort feedback during training generally differ between patients.
This generative capability is integrated into the two working modes of the upper-limb exoskeleton robot: an active mirroring mode for patients who retain motor abilities on one side of the body and a passive following mode for patients who lack motor ability on both sides of the body.
In addition, the upper-limb exoskeleton robot has three other attractive features. First, it has six degrees of freedom (DoFs), namely five active DoFs and one passive DoF, to assist the shoulder and the elbow joints and cover the full range of upper-limb movement.
Second, most of its active joints are driven by series elastic actuators (SEAs) and a cable mechanism, which absorb energy and have low inertia.
These compliantly driven high DoFs provide substantial flexibility and ensure hardware safety but require an effective controller.
Thus, based on the singular perturbation approach, a model-based impedance controller is proposed to fully exploit the advantages of the hardware.
Third, the safety of the upper-limb exoskeleton robot is guaranteed by its hardware and software.
Regarding hardware, its SEAs are tolerant to impacts and have high backdrivability. Regarding software, online trajectory refinement is performed to regulate the assistance provided by the upper-limb exoskeleton robot, and an anomaly detection network is constructed to detect and relax physical conflicts between the upper-limb exoskeleton robot and the patient.
The performance of the upper-limb exoskeleton robot was illustrated in experiments involving healthy subjects and stroke patients.
\end{abstract}


\maketitle

\section{Introduction}
\begin{figure*}[!h]
  \centering
    \includegraphics[width=0.9\linewidth]{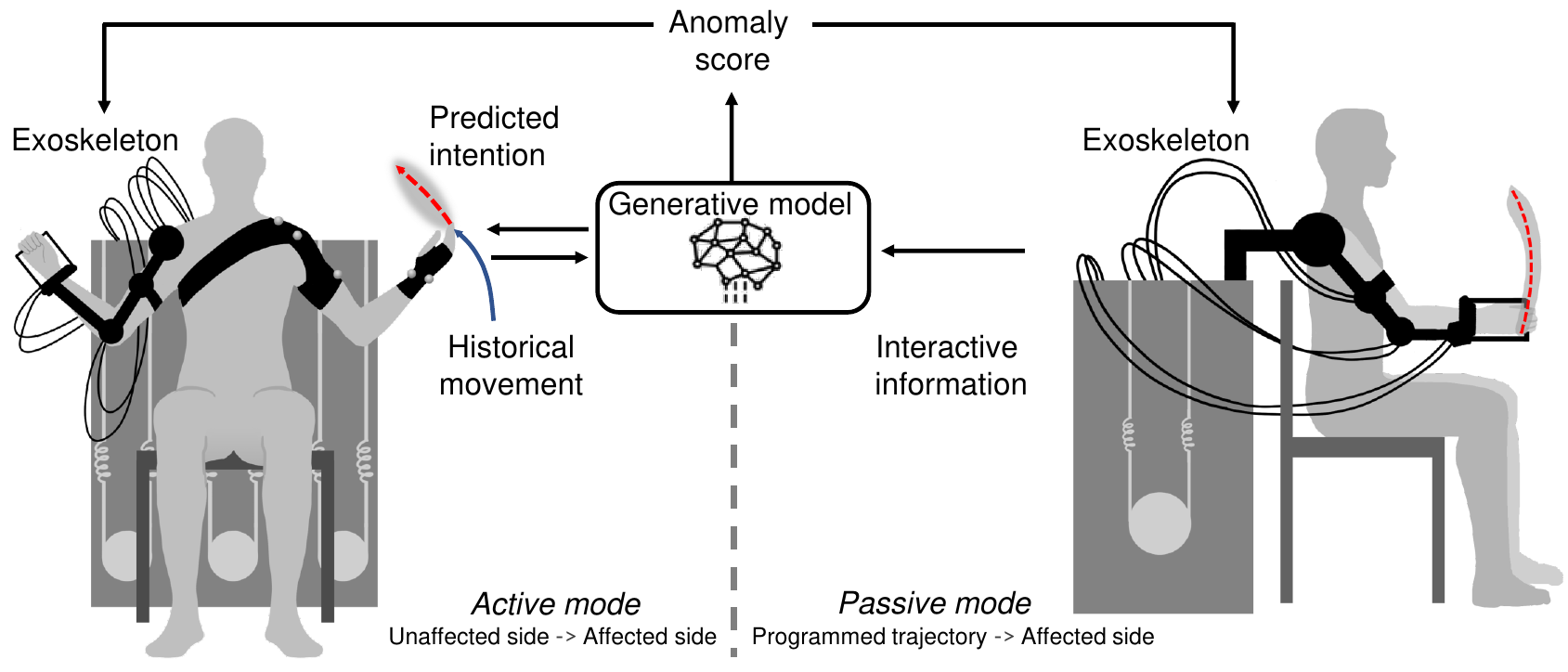}
  \caption{Dual-mode upper-limb exoskeleton rehabilitation training. In active mirroring mode, the reference trajectory of the affected side is determined by the motion intention of the unaffected side (the red dashed line). Conversely, in passive following mode, the reference trajectory is pre-programmed (the red dashed line). The generative model in active mirroring mode can predict the motion intentions of the unaffected side and utilize interactive feedback to generate an anomaly score, which is used for trajectory refinement in both modes.}
  \label{two_mode} 
\end{figure*}
Exoskeleton robots offer several advantages in stroke rehabilitation. In particular, they enable precise control of movement, consistent and repeatable therapy sessions, and collection of real-time data on patient progress \citep{huang2009robotic,jezernik2004automatic,said2022systematic}. 
Furthermore, they offer high-intensity training while reducing physical burdens on therapists, and they are adaptable to the needs and recovery stages of individual patients \citep{gull2020review}.
This paper focuses on rehabilitation using upper-limb exoskeleton robots, a training task that is the focus of several prototypical or commercially available products.
Among these products, the Harmony \citep{kim2017upper} stands out, as it is equipped with an anatomical shoulder mechanism designed to augment and facilitate the arm’s natural movements.
Additionally, ANYexo \citep{zimmermann2023anyexo} enhances the upper limb mobility of patients and exhibits sufficient flexibility to encompass a wide array of daily activities.
Furthermore, the Armeo Power (Hocoma) \citep{lee2020comparisons} enables early-stage stroke patients to start intensive arm therapy.
However, current upper-limb exoskeleton robots may not fully achieve the main goal of upper-limb rehabilitation, i.e., recovery of the manipulability of the human body with high degrees of freedom (DoFs), as they have too few DoFs to match the movement of a healthy subject and a rather low capacity for individualization. Individualization is a key feature of an upper-limb exoskeleton robot used in rehabilitation, as it enables the provision of assistance that is customized to the condition (e.g., stroke duration and medical background) of the patient.
Such assistance helps the patient to regain movement ability that is original and natural, i.e., devoid of abnormal patterns.
Hence, insufficient individualization results in low-quality rehabilitation training. 

Generally, either {\em active mirroring} or {\em passive following} training modes are applied in the rehabilitation of stroke patients, according to their medical condition and as illustrated in  Figure \ref{two_mode}. 
Active mirroring training is used for patients who have mild impairments and an unaffected side of the body and thus retain motor abilities.
In this mode, a generative model known as the intention predictor estimates the patient’s motion intentions based on historical motion data.
Subsequently, these intentions are mirrored by an upper-limb exoskeleton robot on the affected side of the patient’s body.
This process aims to align motion intentions between the unaffected and affected sides of the body at a neural level, thereby increasing patient engagement.
Passive following training is used for patients with severely impaired motor abilities.
In this mode, the patient follows a pre-defined trajectory facilitated by an upper-limb exoskeleton robot, which provides structured guidance for movement rehabilitation.
The aforementioned modes can also be utilized together with manual training for patients with different medical conditions or in different stages of stroke. For example, using the passive mode at the early stage and then the active one, when the healthy side can move naturally.
Such dual-mode rehabilitation is important for ensuring that an upper-limb exoskeleton robot can be used by multiple patients with various levels of physical ability.

As patients wear an upper-limb exoskeleton robot to carry out training procedures, ensuring patient safety in the presence of tight and continuous physical interactions is a primary concern.
Much progress has been made in this regard, in terms of both hardware and software. In terms of hardware, safety is increased by using compliant actuators such as series elastic actuators (SEAs) \citep{pratt1995series} to absorb impact forces during interactions. Additionally, bio-inspired cable-driven actuators \citep{xu2023bio} are used to transition between different actuation modes, and safety is further guaranteed by backdrivable actuators \citep{zhu2021design}.
In terms of software, safety regulation can be achieved through various control strategies, including impedance control \citep{li2018iterative}, velocity field-based control \citep{martinez2018velocity}, and data-driven ergonomic control \citep{clark2022learning}, all of which ensure that an upper-limb exoskeleton robot system operates safely.
However, most hardware and software approaches are applied after a safety problem has occurred. 
A better approach is to predict potential safety problems and prevent their occurrence. 

Accordingly, this paper describes the development of a new upper-limb exoskeleton robot for rehabilitation training of stroke patients that is superior to previously reported upper-limb exoskeleton robots in terms of the three characteristics described below.
\begin{enumerate}
    \item {\em Safety}: 
    The new upper-limb exoskeleton robot contains SEAs in most of its active joints, so that the joints absorb impact forces caused by collisions and unexpected contacts during upper-limb interaction.
    This design enhances the precision of torque control, thereby providing smoother assistance in rehabilitation training than previously reported upper-limb exoskeleton robots.
    Moreover, an anomaly detection neural network informed by interactive feedback is implemented to assess the safety and naturalness of the movement of the upper-limb exoskeleton robot in real time.
    This network enables the early detection of anomalies, allowing for their prevention or mitigation during rehabilitation tasks.
    Furthermore, an online trajectory refinement module and an impedance controller are incorporated into the upper-limb exoskeleton robot to ensure safety throughout the planning and execution phases of rehabilitation.
    \item {\em Effectiveness}: 
    The upper-limb exoskeleton robot is configured with one passive joint and five active joints, endowing it with a high number of DoFs and thus enabling the workspace to accommodate a broad spectrum of daily activities.
    Given the inherent uncertainty and randomness associated with human motion intentions and upper-limb movements, the system efficiently generates trajectories by exploiting a probabilistic model of motion for explicit consideration during the sampling process so as to match the nature of human intention.
    The effectiveness of the upper-limb exoskeleton robot was validated through clinical trials, which demonstrated that the experimental group, which underwent passive following training, recovered motor abilities more quickly than the control group.
    \item {\em Friendliness}:
    The active joints of the upper-limb exoskeleton robot are capable of sensing force and thus can detect interaction torque.
    Therefore, when the upper-limb exoskeleton robot is operated in transparent mode for demonstration purposes, it can be maneuvered effortlessly by the patient.
    Moreover, given the substantial load sustained by the shoulder joint during upper-limb movements, a direct-drive motor is incorporated into this joint in the upper-limb exoskeleton robot.
    This motor delivers enhanced driving torque and increases backdrivability by eliminating nonlinear friction forces.
    Furthermore, to reduce the effects of inertia during movement and thus increase comfort, a cable-driven mechanism is utilized in joints that must have a broader range of motion in space than other joints.
\end{enumerate}

The key novelty of this study is its design and implementation of a generative-model-based refinement framework that is capable of generating highly individualized trajectories and ensuring safety.
Specifically, the framework operates in either of the two aforementioned distinct modes, namely {\em active mirroring} mode and {\em passive following} mode.

The {\em active mirroring} mode exhibits the newly developed key features described below.
\begin{itemize}
\item[-] A novel diffusion model-based motion intention predictor, which uses historical motion data of the patient’s unaffected side to estimate the patient’s upper-limb motion intentions.
This allows for the prediction of future trajectories and establishes the mean and variance of the patient’s motion intentions.
\item[-] A preemptive tuning algorithm that ensures that a predicted trajectory remains within a safe region by exploiting the predicted distribution to mitigate potential risks.
\end{itemize}
The {\em passive following} exhibits the newly developed key features described below.
\begin{itemize}
\item[-] A diffusion model-based anomaly detection network capable of evaluating safety and naturalness based on an anomaly score. This score is utilized to guide the online trajectory refinement and evaluation for each rehabilitation task.
\item[-] A probabilistic movement primitives (ProMPs)-based approach to trajectory generation that captures the distribution of the patient’s motion intention. In addition, based on each sampling result’s performance, this approach iteratively optimizes the assistive trajectory.
\end{itemize}
Both modes customize the assistance and generate an individualized trajectory for the patient, enabling natural and original motion patterns to be recovered by the patient.
In particular, the intention predictor and anomaly detector utilize generative models to capture patterns in the latent space of large datasets.
Moreover, the intention predictor and anomaly detector have distinct functions. The intention predictor accurately predicts the patient’s original motion intentions. This prediction serves as a self-mirrored trajectory that is finely tailored to the needs of the patient’s affected side and thus aligns closely with the patient’s motion intentions.
Simultaneously, the anomaly detector quantitatively identifies the differences between the patient’s rehabilitation movements and those of healthy individuals. These differences are used to guide the adjustment of rehabilitation exercises, thereby enhancing the individualization of the treatment.

In a previous study \citep{chen2023safe}, we developed an individualization framework for passive following training using a variational autoencoder (VAE)-based anomaly detector. In the current study, we expand this approach to a dual-mode upper-limb exoskeleton robot by incorporating generative model technology, i.e., by utilizing a diffusion model. We present results from a series of experiments and comparisons that confirm the safety and performance of the new upper-limb exoskeleton robot.
In addition, we present results from a clinical trial that validate its effectiveness and friendliness. 



\section{Related Works}
This section reviews related work on upper-limb exoskeleton robots, trajectory generation, and interaction control. Unlike those that have been previously reported, the upper-limb exoskeleton robot reported in this paper has an efficient generative capability, which enables it to effectively provide individualized assistance.

\begin{table*}
\centering
\caption{Comparison of Different Upper-Limb Exoskeleton Robots}
\begin{tblr}{
  hlines,
  vline{2} = {-}{},
  hline{1,6} = {-}{0.08em},
}
EXOSKELETON                                      & MOVEMENT            & NO.OF DOFs                     & SENSOR                    & ACTUATION \\
\cite{wu2016development}  & {Shoulder,\\Elbow }            & {Four-Active\\One-Passive   } & Encoders                    & Direct                  \\
\cite{zimmermann2023human} & {Shoulder,\\Elbow,\\Forearm  } & Nine-Active                 & {Encoders\\Force sensors  } & Compliant                  \\
\cite{mao2012design}      & {Shoulder,\\Elbow,\\Forearm  } & Five-Active                 & {Encoders\\Force sensors  } & Cable-driven                  \\
Ours                                        & {Shoulder,\\Elbow,\\Forearm  } & {Five-Active\\One-Passive  }  & {Encoders\\Force sensors  } & Cable-driven and compliant                 
\end{tblr}
\label{Struct_comp}
\end{table*}

\subsection{Upper-Limb Exoskeleton Robots}
Upper-limb rehabilitation necessitates a large range of motion. Therefore, upper-limb exoskeleton robots must possess a sufficient number of DoFs to accommodate dynamic and complex upper-limb movements.
The redundancy provided by multiple DoFs can enhance patient comfort \citep{kim2012redundancy}. 
However, many upper-limb exoskeleton robots have been primarily designed for the rehabilitation of single joints, such as the shoulder \citep{nasr2023optimal}, elbow \citep{chen2019elbow}, or wrist \citep{martinez2013design}. 
In contrast, our newly developed upper-limb exoskeleton robot is designed to fulfill the requirements of most upper-limb rehabilitation tasks and thus features one passive joint and five active joints.
It also features advanced sensory capabilities and a lightweight mechanical design.
Table \ref{Struct_comp} compares our upper-limb exoskeleton robot with previously reported designs.

Unlike traditional upper-limb exoskeleton robots, which are directly motor-driven, our design incorporates a cable-driven mechanism.
Thus, it has a significantly lower weight and inertia effect than traditional upper-limb exoskeleton robots \citep{perry2007upper}. 
The cable-driven mechanism also offers the advantages of flexibility, low-backlash gearing, and backdrivable transmission \citep{jau1988anthropomorhic}. 
However, the friction generated during cable transmission introduces nonlinear characteristics that complicate joint control \citep{wang2022extracting}.
Compared with other joints, the shoulder joint requires a higher driving torque and thus exhibits higher friction. Thus, this joint requires additional enhancements to optimize its performance.

During exoskeleton robot-supported rehabilitation, there is tight and frequent human–robot interaction and thus compliant actuation is essential.
In particular, the use of compliant actuation ensures flexibility, mechanical safety, and good interaction performance \citep{tiboni2022sensors, kalita2021development}.
As such, many upper-limb exoskeleton robots have used SEAs \citep{ebrahimi2017control, pan2022nesm, li2021bear}.
Accordingly, we incorporate SEAs within the cable-driven mechanism of our upper-limb exoskeleton robot to improve its interaction performance.

Previously developed upper-limb exoskeleton robots have demonstrated good flexibility. 
However, few have simultaneously incorporated cable-driven mechanisms and compliant actuator designs (to offer sufficient numbers of DOFs for expansive rehabilitation training) and force sensors (to offer enhanced sensory capabilities).
Furthermore, the combination of cable drive and SEA structures creates a system with high-order nonlinear dynamics, thereby presenting substantial control challenges.






\subsection{Trajectory Generation}
The assistance provided by an upper-limb exoskeleton robot can be represented by the assistive (i.e., desired) trajectory that it follows to interact with the stroke patient.
For the two modes of training considered in this study, the assistive trajectory is generated in different ways.

\begin{figure*}[!h]
  \centering
    \includegraphics[width=0.9\linewidth]{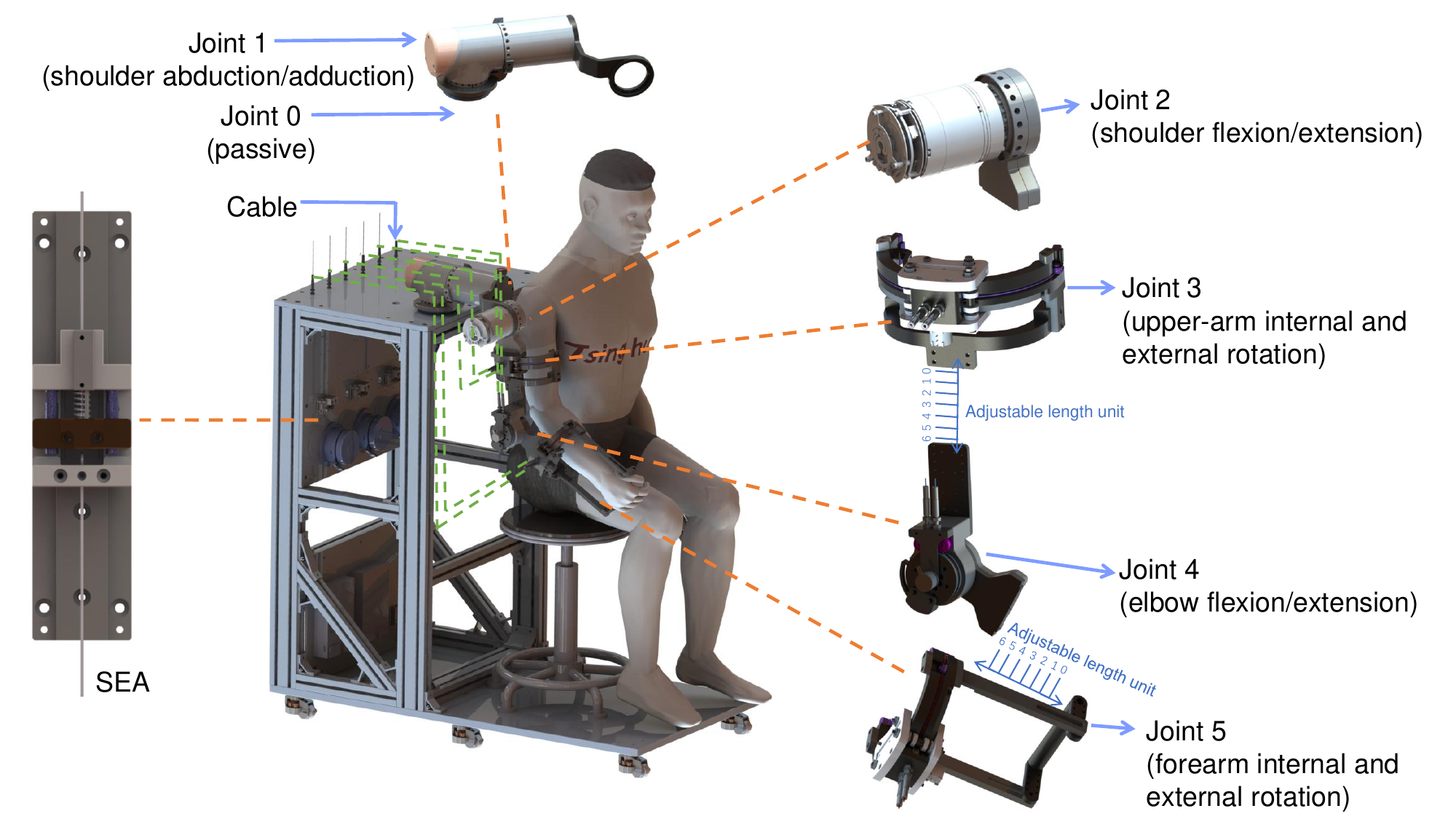}
  \caption{Overview of the developed upper-limb exoskeleton robot, which is cable-driven and consists of five active joints (Joints 1–5) and one passive joint (Joint 0).}
  \label{wholeexo} 
\end{figure*}

In \textit{active mirroring} mode, the aim of assistance is to replicate the motion of the unaffected side of the patient’s body, and the assistive trajectory is generated as a sequence of predicted human motion intentions.
Various methods have been developed to determine human motion intentions from sensor information, such as that generated by inertial measurement units \citep{zhu2020novel}, force-sensing resistors \citep{huang2015control}, and surface electromyography \citep{lenzi2012intention}.
Additionally, Gaussian process regression was employed to estimate human motion intention from human–robot interaction data \citep{long2018human}.
Furthermore, a multi-sensor fusion-based method was proposed to adaptively update an assistance profile to enhance mirror training rehabilitation \citep{li2022multi}.
Moreover, learning-based approaches that exploit neural networks’ capability to incorporate multimodal data and their excellent expressive capabilities have been successfully implemented in mirror training rehabilitation \citep{xu2020multi, li2023subject}.
However, the above-mentioned methods have primarily been designed to focus on immediate motion intentions and thus struggle to predict motion intention trends over time. 
Given that the affected side of the body of a stroke patient has limited motion ability, failing to predict motion intention may result in crucial factors such as safety constraints being overlooked, thereby posing risks to the patient during rehabilitation training.

In \textit{passive tracking} mode, the aim of assistance is to enable the patient to follow a pre-defined trajectory.
In this mode, the simplest method for trajectory generation involves discretizing the trajectory into multiple waypoints and then performing linear interpolation of discrete points and trajectory constraints within a planning framework \citep{sommerhalder2023trajectory}.  
Trajectories may also be parameterized using polynomials and dynamic movement primitives to meet the requirements of different assistive tasks \citep{kagawa2015optimization, lanotte2021adaptive, qiu2020exoskeleton}.
Additionally, assistance can be provided through torque profiles generated in a human-in-the-loop manner \citep{zhang2017human} or determined via reinforcement learning \citep{zhang2022imposing}.
However, the above-mentioned methods typically optimize feasible trajectories at only the kinematic level or by using metabolic measurements, which are labor-intensive to obtain.
Thus, these methods fail to simultaneously effectively incorporate safety constraints and personalize a trajectory, despite the former being the primary concern for an exoskeleton robot, and the latter being important for ensuring that recovered motion is natural and original.

\subsection{Interaction Control}
Hardware development for exoskeleton robots driven by compliant actuators is well advanced.
However, controller development for such exoskeleton robots remains challenging, because their cable-driven mechanism forms a high-order system with significant nonlinearity.
To mitigate this nonlinearity, which is caused by the coupling of compliant actuators and rigid joints, several control strategies have been devised for accurate and flexible interaction control. For instance, the backstepping control method \citep{pan2017adaptive, li2018iterative} has been devised to hierarchically deliver desired control commands, enabling position or impedance control.
Additionally, singular perturbation theory \citep{spong1987modeling} eliminates the need for high-order motion information.
Thus, a method based on singular perturbation has been developed that utilizes the intrinsic differences between two subsystems on exoskeleton robots equipped with SEAs.
This method has been used to generate a multimodal control scheme \citep{li2017multi} and for adaptive trajectory tracking \citep{han2023human}.
Moreover, a recent advancement is a learning-based method \citep{sambhus2023real} that further refines interaction control capabilities.

Exoskeleton robots featuring cable-driven mechanisms can have lower weights than those featuring other mechanisms.
However, the substantial friction generated during cable transmission and movement may hinder effective interaction control.
Multiple methodologies have been developed to alleviate friction and other disturbances in human-robot interactions. 
For instance, \cite{Li2018IterativeLI} employed an iterative learning control algorithm that compensates for disturbances, including friction, at both motor and joint ends by accurately fitting disturbances.
\cite{li2023human} used a pulley model to obtain an analytical expression for the friction and the tension output from a Boden cable.
This model facilitated adaptive learning of friction parameters through the system’s dynamic structure to effectively counteract the above-mentioned disturbances.
Additionally, \cite{wang2022extracting} demonstrated that friction parameters can be obtained via nonlinear fitting techniques.

However, despite significant advancements in the management of SEA systems and the mitigation of disturbances in cable transmission, it remains challenging to integrate these capabilities to enable flexible interaction control.
Moreover, in cable-driven exoskeleton robots, disturbance signals and human-machine interaction forces are often overlapping.
Thus, there remains a need for a systematic consideration of all of the aforementioned aspects combined with a rigorous proof of the stability of a closed-loop system, as this would enable a full exploration of the advantages of hardware (i.e., high numbers of DoFs, and cable-driven and compliant actuation).

\section{Overall Structure}
This study examined a prototype cable-driven upper-limb exoskeleton robot that was co-developed by Tsinghua University and Shenzhen MileBot Robotics Co., Ltd. A computer-aided design model of the upper-limb exoskeleton robot is illustrated in Figure \ref{wholeexo}.
The upper-limb exoskeleton robot is equipped with one passive and five active joints that allow refinement of the upper-arm and forearm lengths within ranges of 20 and 40 cm, respectively.
The passive DoF, labeled as Joint 0, accommodates the eccentric movements of the shoulder joint and thereby eliminates the limitations of the upper-limb motion range and increases the space for patient motion training.
The active joints are numbered 1 to 5 and serve specific functions, as described below.
 \begin{enumerate}
 \item[-] Joint 1: shoulder abduction and adduction
 \item[-] Joint 2: shoulder flexion and extension
 \item[-] Joint 3: upper-arm internal and external rotation
 \item[-] Joint 4: elbow flexion and extension
 \item[-] Joint 5: forearm internal and external rotation
 \end{enumerate}

\begin{figure}[!h]
  \centering
    \includegraphics[width=3.0in]{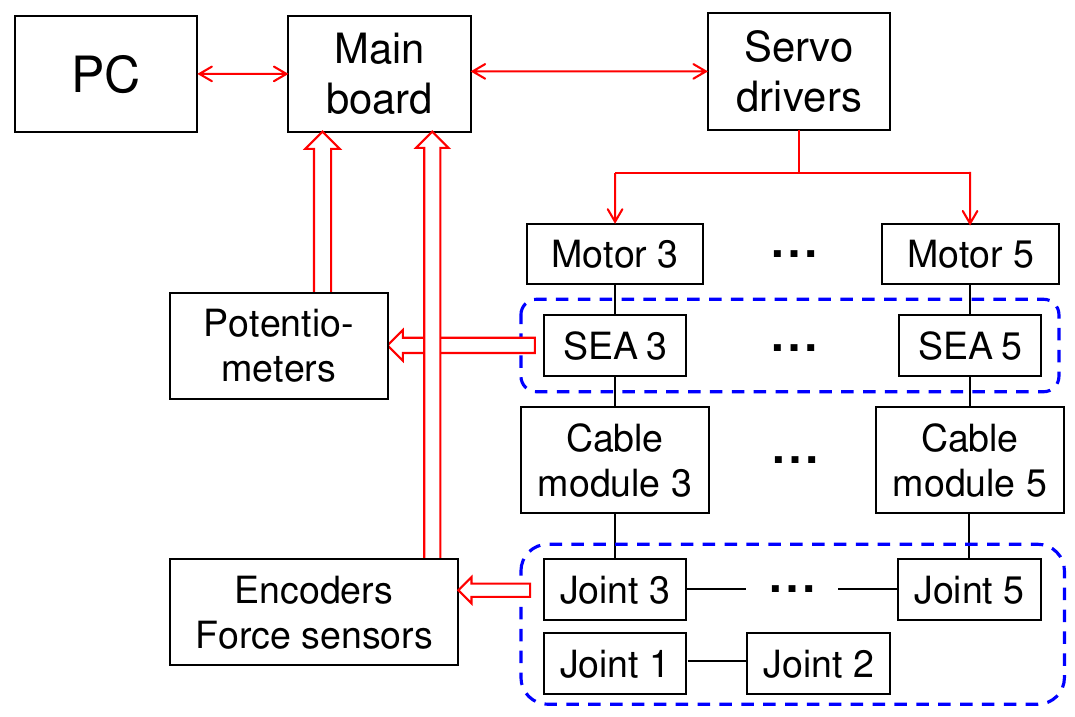}
  \caption{Block diagram of the upper-limb exoskeleton robot, where the red arrow represents the electrical transmission, and the black line represents the mechanical connection}
  \label{structBD} 
\end{figure}

A block diagram of the upper-limb exoskeleton robot is shown in Figure \ref{structBD}.
Joints 1 and 2 consist of direct-drive joint modules (RJSIIT-17-RevB2 and RJSIIT-17-RevB5, respectively), each of which is outfitted with force sensors and supported by the frame. Joints 3 to 5 utilize harmonic deceleration servo motors (AK80-64) that are configured as cable-driven SEAs and equipped with torque sensors (TK17-191151) and encoders (3590S-2-104L for the internal and external rotation joints, and QY2204-SSI for the elbow joint). 
Additionally, these joints incorporate two potentiometers for measuring spring compression during motion. 
The detailed technical specifications of the upper-limb exoskeleton robot are provided in Table \ref{specifications}.
Motor drivers are implemented to control Joints 3–5, and together with two joint modules, facilitate communication with the main board through a controller area network.
The main board receives commands from a personal computer (PC) via a universal asynchronous receiver/transmitter.

The upper-limb exoskeleton robot employs a hierarchical control system.
At the low level, the main board executes motion control and environmental perception, enabling precise force control and real-time acquisition of sensory data. 
At the high level, a Linux system on the PC operates a motion planning module and performs computationally intensive tasks, such as dynamic calculation and network inference.
All programs are integrated into Robot Operating System (ROS) \citep{quigley2009ros} and run concurrently.


\begin{table}[!h]
\centering
\caption{Specifications of the Upper-Limb Exoskeleton}
\begin{tblr}{
  hline{1,7} = {-}{0.08em},
  hline{2} = {-}{},
}
Property          & Value                         \\
Continuous torque & 49 N$\cdot$m @ Joint 1 and 2            \\
                  & 48 N$\cdot$m @ Joint 3,4, and 5            \\
Backdrivability   & {Less than 0.7 N$\cdot$m @ 10 $^{\circ}$/s} \\
Weight            & 7.78 kg excluding the frame     \\
Control frequency & 1000Hz                        
\end{tblr}
\label{specifications}
\end{table}

\begin{figure}[!h]
  \centering
    \includegraphics[width=1\linewidth]{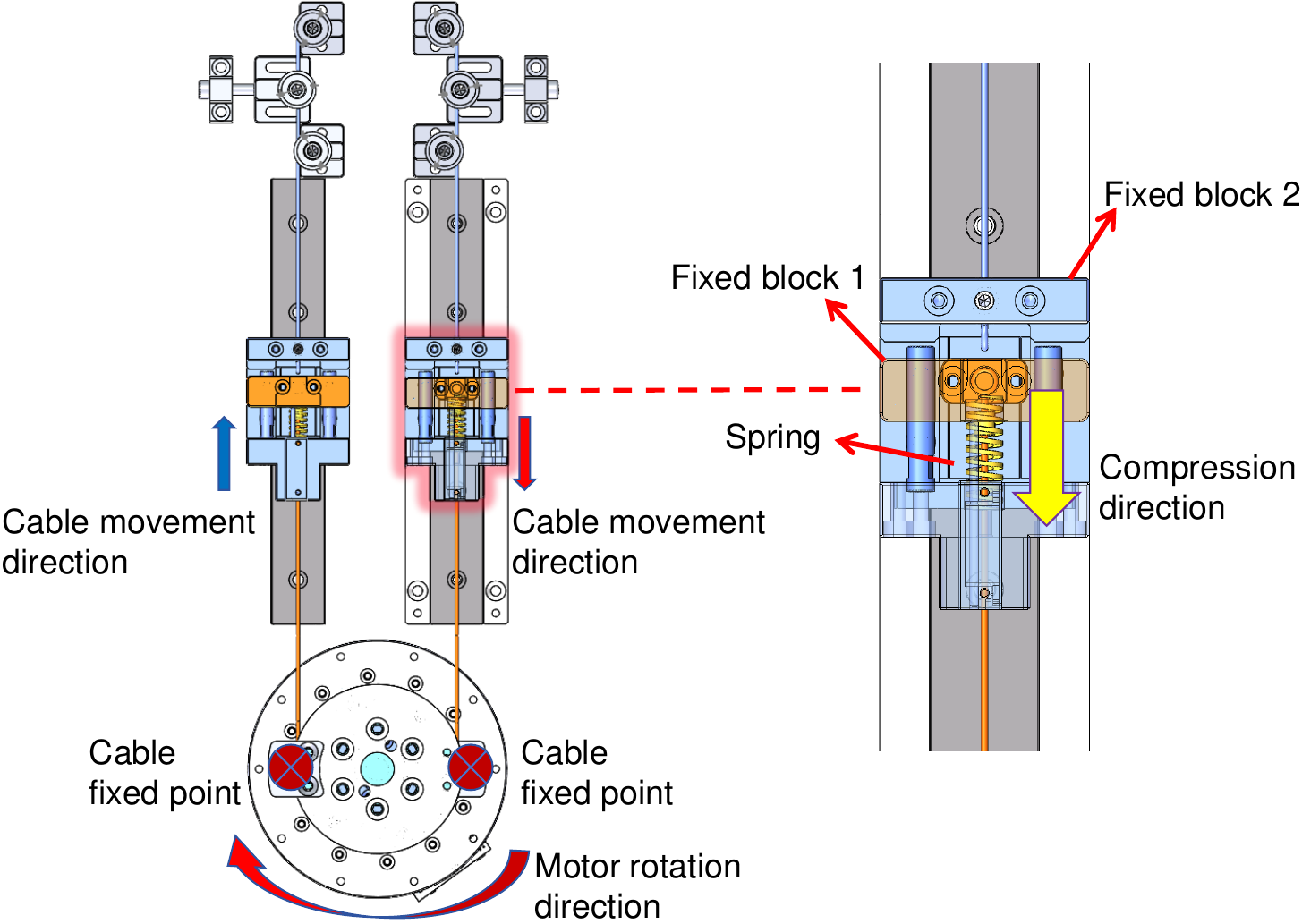}
  \caption{The operating principle of the self-developed SEA}
  \label{sea} 
\end{figure}

The joints that undergo significant spatial movement, namely Joints 3, 4, and 5, utilize our self-developed SEAs equipped with cable-driven mechanisms.
A working principle diagram of a SEA is provided in Figure \ref{sea}.
Each joint motor is outfitted with two sets of pulleys, with each set corresponding to a different rotational direction. 
Each end of the motor output shaft is affixed with a steel cable that is connected to a potentiometer to constitute the SEA.
The potentiometer assembly includes two fixed blocks, namely fixed block 1 and fixed block 2, and a spring component. The cable from the motor end (depicted in red) is connected via a spring to fixed block 1, while another cable (depicted in blue) connects the moving joint components to fixed block 2.
As the motor applies torque, fixed block 1 moves along a sleeve attached to fixed block 2, thereby compressing the spring and enabling the measurement of tension within the cable. Upon spring compression, the blocks form a unified structure that moves in synchrony with and in the same direction as the joint, thereby facilitating overall movement.

\begin{figure*}[!t]
  \centering
    \includegraphics[width=6in]{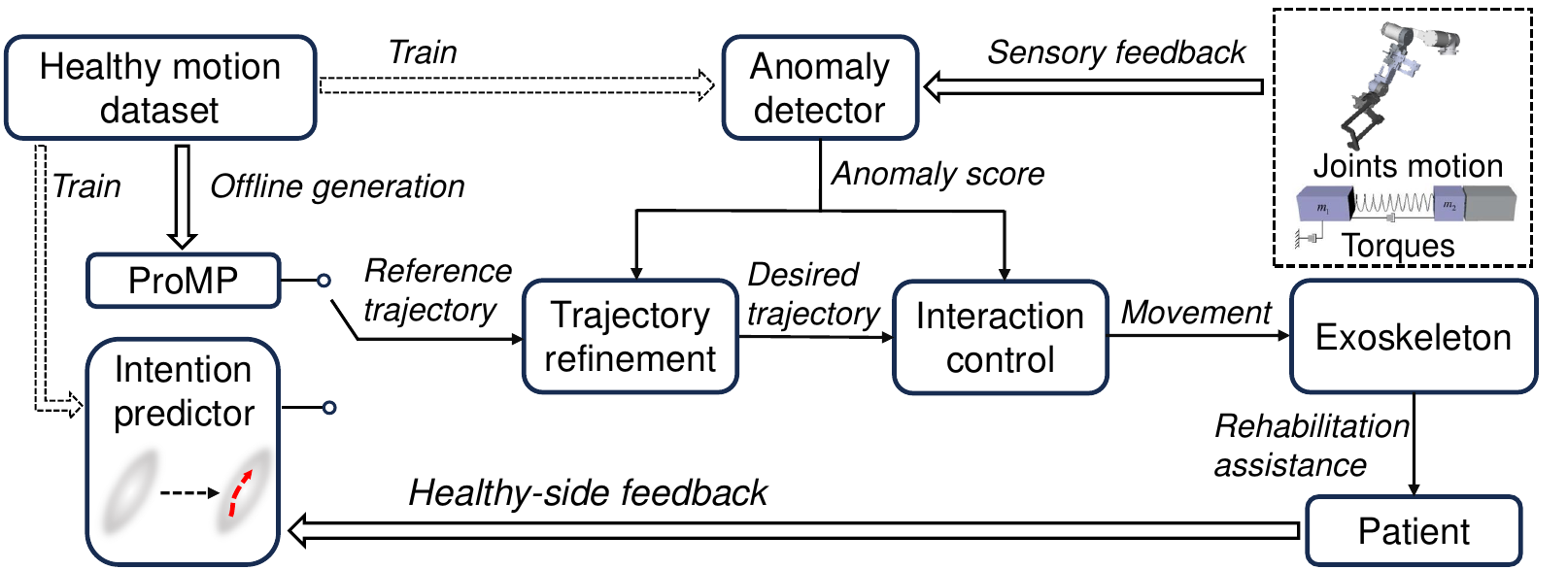}
  \caption{Workflow of dual-mode training. In active mirroring mode, the predicted motion intention serves as the reference trajectory. In passive following mode, the reference trajectory is generated offline by fitting ProMPs to the data obtained from the demonstrations of healthy individuals.
  In both modes, the overarching goal is to deliver the most personalized assistance possible. Additionally, the motion data from healthy individuals are used to train the intention predictor and anomaly detector, and the anomaly score is used to guide trajectory refinement and the adjustment of impedance parameters.}
  \label{passiveBD} 
\end{figure*}

For safety, the upper-limb exoskeleton robot is equipped with an emergency stop button that physically disconnects power during urgent situations. Additional safety measures are incorporated at the software level, including measures that limit the range of motion, prevent self-collision, restrict joint velocity, and automatically cease operation in cases of excessive interaction force or torque.

A dynamic model of a SEA-driven upper-limb exoskeleton robot can be defined as follows \citep{albu2007unified}: 
\begin{align}
\bm M(\bm q)\ddot{\bm q} + \bm C(\dot{\bm q}, \bm q)\dot{\bm q} + \bm g(\bm q) &= \bm S_1 \bm u + \bm S_2^{\mathsf{T}} \bm K (\bm\theta- \bm S_2 \bm q)+ \notag \\
&\quad \bm \tau_{e} +\bm S_2^{\mathsf{T}} \bm \tau_{f},\label{jointsys}\\
\bm B\ddot{\bm \theta} + \bm K (\bm\theta-\bm S_2 \bm q) &= \bm S_2\bm u,\label{seasys}
\end{align}
where  $\bm q\hspace{-0.05cm}\in\hspace{-0.05cm}\Re^{5}$ is the vector of joint positions; $\bm \theta \in \Re^{3}$ is the vector of motor-rotor-shaft positions; and $\bm M(\bm q)\in \Re^{5\times 5}$, $\bm C(\dot{\bm q}, \bm q)\dot{\bm q}\in \Re^{5\times 5}$, and $\bm g(\bm q)\in \Re^{5}$ are the inertia matrix, centripetal and Coriolis torques, and gravitational torques of the robot, respectively.
The selection matrices $\bm S_1=\text{diag}(1, 1, 0, 0, 0)\in \Re^{5\times 5}$ and $\bm S_2 = [\bm 0, \bm I_3]\in \Re^{3\times 5}$ are used to separate the cable-driven joints. In addition,
$\bm K\in \Re^{3\times 3}$ is the stiffness matrix, $\bm B\in \Re^{3\times 3}$ is the inertia matrix of the motor, 
$\bm \tau_{e} \in \Re^{5}$ is the physical interaction torque vector, $\bm \tau_{f} \in \Re^{3}$ is the disturbance due to the cable transmission and the friction of the joint system, and $\bm u \in \Re^{5}$ is the control torque applied to the actuator.

The dynamic model of the overall system is described by (\ref{jointsys}) and (\ref{seasys}) and exhibits different time-scales. 
Specifically, the SEA subsystem operates on a fast time-scale, while the upper-limb exoskeleton robot subsystem functions on a slow time-scale.
To control this system effectively, the control input is formulated according to singular perturbation theory \citep{spong1987modeling}, as follows:
\begin{eqnarray}
&\bm u=\bm u_f +\bm u_s,\label{controlSection3}
\end{eqnarray}
where $\bm u_f$ is the fast time-scale control term for stabilizing the model defined by (\ref{seasys}), and $\bm u_s$ is the slow time-scale control term for stabilizing the model defined by (\ref{jointsys}). A representative form of $\bm u_f$ is expressed as follows:
\begin{eqnarray}
&\bm u_f=-\bm S_2^{\mathsf{T}}\bm K_v(\dot{\bm\theta}-\bm S_2 \dot{\bm q}),\label{uf}
\end{eqnarray}
where $\bm K_v\in\Re^{3\times 3}$ is a diagonal and positive-definite matrix. 

Substituting (\ref{controlSection3}) and (\ref{uf}) into (\ref{seasys}) yields
\begin{eqnarray}
\bm B\ddot{\bm\theta}+\bm K(\bm\theta- \bm S_2 \bm q)+\bm K_v(\dot{\bm\theta}-\bm S_2 \dot{\bm q})= \bm S_2 \bm u_s,
\end{eqnarray}
which can be rewritten as follows by defining $\bm \tau_o=\bm K(\bm\theta-\bm S_2 \bm q)$:
\begin{eqnarray}
\bm B\ddot{\bm \tau}_o +\bm K_v\dot{\bm \tau}_o +\bm K \bm\tau_o=\bm S_2 \bm u_s-\bm B \bm S_2 \ddot{\bm q}.
\label{substitute_SP1sec}
\end{eqnarray}

By introducing $\bm K = \bm K_1/\varepsilon ^{2}$ and $\bm K_v = \bm K_2/\varepsilon $, with $\varepsilon$ being a small positive parameter, (\ref{substitute_SP1sec}) can be written as follows:
\begin{eqnarray}
&\varepsilon^2\bm B\ddot{\bm \tau}_o+\varepsilon \bm K_2\dot{\bm \tau}_o+\bm K_1\bm \tau_o=\bm K_1(\bm S_2 \bm u_s-\bm B \bm S_2 \ddot{\bm q}).
\label{substitute_SP3sec}
\end{eqnarray}

When $\varepsilon \hspace{-0.05cm}=\hspace{-0.05cm}0$, the solution of (\ref{substitute_SP3sec}) is $\Bar{\bm \tau}_o=\bm S_2 \bm u_s-\bm B \bm S_2 \ddot{\bm q}$.

If the fast time-scale is set as $\gamma=\frac{t}{\varepsilon }$, $\Bar{\bm \tau}_o$ is achieved at $\gamma\rightarrow\infty$. 
$\Bar{\bm \tau}_o$ remains constant at $\varepsilon =0$.
Next, we introduce a new variable $\bm \eta=\bm \tau_o-\Bar{\bm \tau}_o$ to rewrite (\ref{substitute_SP3sec}) on the fast time-scale, as follows:
\begin{eqnarray}
\bm B(\frac{\rm d^2\bm \eta}{\rm d\gamma^2})+\bm K_2(\frac{\rm d\bm \eta}{\rm d\gamma})+\bm K_1\bm \eta = \bm 0
\label{boundary_Layer},
\end{eqnarray}
which defines the \textit{boundary-layer system}.

Substituting the solution of (\ref{boundary_Layer}) into (\ref{jointsys}), affords a \textit{quasi-steady-state system} that captures the slow dynamics of the overall system. These dynamics can be expressed as follows:
\begin{align}
    (\bm M(\bm q)+\bar{\bm B})\ddot{\bm q}+\bm C(\dot{\bm q}, \bm q)\dot{\bm q}+\bm g(\bm q)=&\bm u_s+\bm\tau_e + \bm S_2^{\mathsf{T}} \bm\tau_f,\label{substitute_SP4sec}
\end{align}
where $\bar{\bm B} = \bm S_2^{\mathsf{T}}\bm B \bm S_2$ is the projected motor inertia matrix. 

According to singular perturbation theory, the stability of the overall system is guaranteed if the \textit{boundary-layer system} and the \textit{quasi-steady-state system} are both exponentially stable. A stability analysis of this system is provided in the Appendix.

The overall dynamic model described by (\ref{jointsys}) and (\ref{seasys}) is a high-order system, where (\ref{jointsys}) represents the rigid-joint side, and (\ref{seasys}) represents the compliant actuator side. It is non-trivial to stabilize and control such a system.

An upper-limb exoskeleton robot needs to guide the patient to perform repetitive motions via close interaction to help the patient to regain motor function.
Thus, the upper-limb exoskeleton robot is controlled to track a time-varying trajectory in accordance with the following impedance model: 
\begin{eqnarray}
&\bm M_{d}(\ddot{\bm q}-\ddot{\bm q}_{d})+ \bm C_{d}(\dot{\bm q}-\dot{\bm q}_{d})+\bm K_{d}( \bm q-\bm q_{d})=\bm \tau_{e},\label{desiredImpedance}
\end{eqnarray}
where $\bm q_{d} \in \Re^{n}$ is the vector of the desired trajectory, and $\bm M_d, \bm C_d, \bm K_d\in \Re^{n\times n}$ are the desired inertia, the desired damping, and the desired stiffness matrices, respectively, which are diagonal and positive-definite. Tracking the desired trajectory in accordance with the impedance model allows the patient to deviate from the trajectory and hence provides a certain level of compliance to improve safety.

The proposed dual-mode training framework is illustrated in Figure \ref{passiveBD}. 
Both the anomaly detector and intention predictor are designed using trained generative models based on a dataset of motions of healthy individuals.
These models are designed to capture the motion intentions of the unaffected side of the patient and evaluate the comfort and naturalness of the current patient–robot interaction.
Individualization is facilitated by employing two methods, i.e., employing the anomaly score to guide the refinement of a desired trajectory that complies with dynamic constraints, and employing ProMPs to integrate various assistive trajectories into a personalized assistance distribution.
Additionally, the safety of the motion process is enhanced by adjusting the impedance parameters according to the anomaly score, thereby ensuring a secure and effective rehabilitation environment.

\section{Online Trajectory Refinement}
This section introduces the new trajectory generation method for the upper-limb exoskeleton robot.
This method is based on the generative models and can be refined online to suit the patient while maintaining safety and individualized features, and hence maintaining effectiveness.
First, in both training modes (i.e., active mirroring and passive following modes), we introduce a general integrated vector $\bm x_d = [\bm q_d^{\mathsf{T}},\dot{\bm q}_d^{\mathsf{T}}]^{\mathsf{T}}$, which encapsulates a sequence of discretized trajectory points within the joint space, thereby parameterizing the trajectory.
The following quadratic programming problem is formulated to describe the planning process:

\begin{align}
\label{opt_all}
\min_{\bm x_d, \bm u_d, s} &\sum_{i=t}^{t+N_p}\left[\|\bm q_{d}^{(i)} - \bm q_{r}^{(i)}\|_{\bm Q}^2 + \|\bm u_{d}^{(i)}\|_{\bm R}^2 + (s^{(i)})^2\right],\\
 s.t.\quad &{\bm x}_d^{(t+1)} = \begin{bmatrix}\bm I & \bm I\Delta t\\\bm 0 & \bm I\end{bmatrix}\bm x_d^{(t)} + \begin{bmatrix}\bm 0\\\bm I\Delta t\end{bmatrix}\bm u_d^{(t)}\notag\\
 &s^{(t+1)}=s^{(t)} + \begin{bmatrix}\bm 0\\-(\frac{\partial f}{\partial \bm \tau_e})^{\mathsf{T}}\bm K_{a}\Delta t\end{bmatrix}\bm x_d^{(t)}\notag\\
&\qquad\qquad-(\frac{\partial f}{\partial \bm \tau_e})^{\mathsf{T}}\bm C_{a}\bm u_d^{(t)}\Delta t\notag\\
&\bm x_d \in \mathcal{X}, \bm u_d \in \mathcal{U} \notag
\end{align}
where $N_p$ is the predictive horizon, $\bm Q$ and $\bm R$ are symmetric positive-definite weighting matrices, $\bm u_d$ is the acceleration of the desired trajectory, $\bm C_{a}$ and $\bm K_{a}$ are the varying impedance parameters to be defined later, and $f(\cdot)$ is a functional representation of the anomaly detection network.
In addition, the superscript indicates the time-step, and $\mathcal{X}$ and $\mathcal{U}$ are the sets of trajectory space constraints and acceleration constraints, respectively.
Thus, (\ref{opt_all}) enables various constraints and safety measures to be included in both training modes.

\subsection{Active Training Mode}
In cases of mild motor impairment, such as that observed in early-stage stroke, patients retain functional motion capabilities on their unaffected side.
Thus, the upper-limb exoskeleton robot must accurately estimate the motion intentions of the patient’s unaffected side to provide mirrored support for the patient’s affected side.
This mirroring process enables an affected limb to emulate the movement of an affected limb, thereby enhancing the recovery of motor function.

The intention estimation system inputs a series of $N$ historical observations, defined as $\bm o_{(i)} = \{ \bm q_{h(i)}^{(t)}\in \Re^{5} | t = -N_{o},-N_{o}+1,\cdots,0 \}, \forall i \in \{1,2,\cdots, N \}$, where $N_{o}$ is the number of past time steps, and $\bm q_{h(i)}^{(t)}$ is the position of the unaffected side of the body at time step $t$.
The system also generates a prediction of the motion intention, expressed as $\bm p_{(i)} = \{ {\bm q}_{h(i)}^{(t)} \in \Re^{5} | t = 1,2,\cdots, N_{p} \}$.
To estimate the potential distribution of plausible trajectories, we introduce a probabilistic predictor based on denoising diffusion probabilistic models (DDPMs) \citep{ho2020denoising}, as DDPMs can effectively describe the uncertain nature of human motion intention.
For clarity, we omit the subscript $(i)$ in the following subsections and refer to the past and future trajectories as $\bm o$ and $\bm p$, respectively.\\

\begin{figure}[!h]
    \centering
    \includegraphics[width=1\linewidth ]{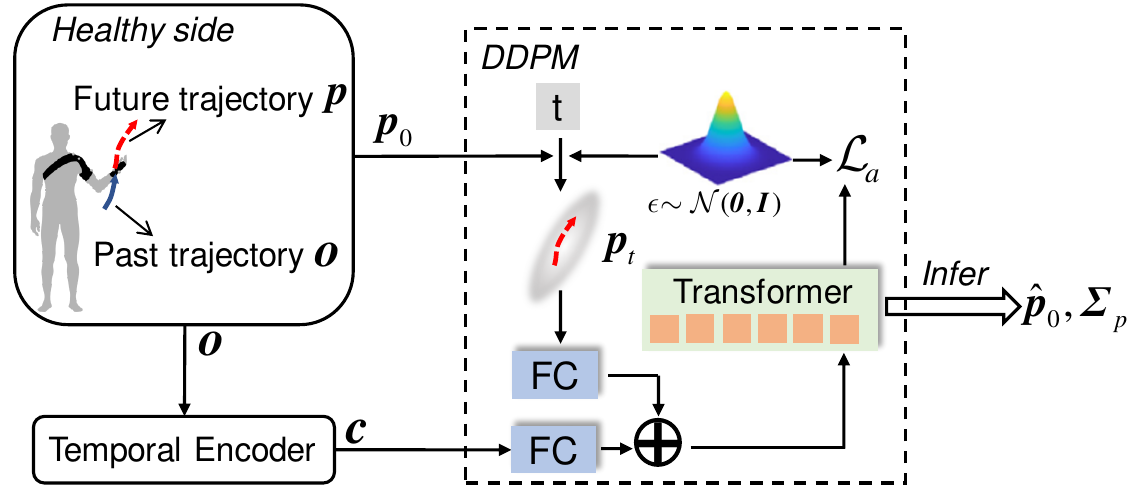}
    \caption{Overall structure of the intention predictor: In the training phase, the past trajectory $\bm o$ is encoded into a context vector $\bm c$, which is used in conjunction with the future trajectory $\bm p$ to train the diffusion model.
    In the inference phase, only the context vector is employed to guide the denoising process, which is used to predict the future trajectory.}
    \label{OUR_pred}
    \vspace{-0.3cm}
\end{figure}

\noindent\textbf{Intention Predictor}: The role of the intention predictor is to produce probabilistic forecasts of future trajectories.
As such, the intention predictor is informed by an initial, ambiguous predicted trajectory $\bm p_T$, which represents a noise-perturbed estimate of the future trajectory $\bm p_0$ after $T$ diffusion steps, where $T$ is the pre-defined maximum number of diffusion steps.
The predictor is designed to infer the reverse diffusion process, articulated as the sequence $(\bm p_T, \bm p_{T-1}, \cdots, \bm p_0)$. 
This sequence is mathematically structured as a Markov chain that is characterized by Gaussian transition probabilities and refines the prediction by reducing the initial uncertainty.
Furthermore, the intention predictor incorporates historical observations encoded into a context vector $\bm c$, which is synthesized via a neural network parametrized by $\phi$. 
This context vector informs the trajectory generation process.
The principle of the intention predictor is depicted in Figure \ref{OUR_pred}.
Within this framework, the upper-limb exoskeleton robot is endowed with the ability to precisely estimate the future trajectory based on a simple sampling from the initial noisy prediction. Importantly, a future trajectory represents the patient’s motion intention and is not based on direct measurements of the patient’s unaffected limb.
Thus, this approach generates a self-mirrored trajectory that closely aligns with the patient’s original movement patterns, thereby representing a personalized trajectory.

The posteriors associated with diffusion and reverse processes are given as follows:
\begin{align}
    q(\bm p_{t}|\bm p_{t-1}) &= \mathcal{N}(\bm p_{t};\sqrt{1-\beta_{t}}\bm p_{t-1},\beta_{t}\bm I),\label{diffusion_process}\\
    q_{\psi}(\bm p_{t-1}|\bm p_{t},\bm c) &= \mathcal{N}(\bm p_{t-1};\bm\mu_{\psi}(\bm p_{t},\bm c),\Tilde{\beta}_{t}\bm I),\label{reverse_process}
\end{align}
where $\beta_1,\beta_2,\cdots,\beta_T$ is the variance schedule used to adjust the injected noise.
The adjusted variance, 
$\Tilde{\beta}_{t}$, is defined as $\Tilde{\beta}_{t}=\frac{1-\bar{\alpha}_{t-1}}{1-\bar{\alpha}_t}\beta_t$, where this formulation is derived by adopting the definitions $\alpha_t = 1-\beta_t$ and $\bar{\alpha}_t = \prod_{i=1}^{t}\alpha_i$.

The objective of the predictor is to maximize the log-likelihood, denoted as $\mathbb{E}[\log q_{\psi}(\bm p_{0})]$. 
Given that direct computation of this metric is infeasible, a variational lower bound is utilized as a surrogate for the training loss:
\begin{align}
    \mathcal{L}_{vlb}(\phi,\psi)=&\sum_{t=2}^{T}D_{KL}\left(q(\bm p_{t-1}|\bm p_{t})\| q_{\psi}(\bm p_{t-1}|\bm p_{t},\bm c)\right)\notag\\
    &-\log q_{\psi}(\bm p_{0}|\bm p_{1},\bm c),
\end{align}
where $D_{KL}(\cdot)$ signifies the Kullback–Leibler divergence function.

In the proposed methodology, the conditional probability $q(\bm p_{t-1}|\bm p_{t})$ is reformulated as $q(\bm p_{t-1}|\bm p_{t},\bm p_0)$. 
This reformulation allows the following closed-form expression to be devised:
\begin{align}
    q(\bm p_{t-1}|\bm p_{t},\bm p_0)=&\mathcal{N}(\bm p_{t-1};\Tilde{\bm\mu}_{t}(\bm p_t,\bm p_0),\Tilde{\beta}_{t}\bm I).\label{condition_posterior}
\end{align}

By iteratively employing parameterization techniques in the diffusion process (\ref{diffusion_process}), the mean of the posterior distribution in (\ref{condition_posterior}) and the reverse process (\ref{reverse_process}) can be articulated as follows:
\begin{align}
    \Tilde{\bm\mu}_{t}(\bm p_t,\bm p_0)=&\frac{1}{\sqrt{\alpha_t}}(\bm p_t - \frac{\beta_t}{\sqrt{1-\bar{\alpha}_t}}\bm \epsilon_t),\\
    \bm\mu_{\psi}(\bm p_{t},\bm c)=&\frac{1}{\sqrt{\alpha_t}}(\bm p_t - \frac{\beta_t}{\sqrt{1-\bar{\alpha}_t}}\bm \epsilon_{\psi}(\bm p_t,\bm c)),
\end{align}
where $\bm \epsilon \thicksim \mathcal{N}(\bm 0, \bm I)$, and the loss function applied in active mirroring mode is further simplified as follows 
\begin{align}
    \mathcal{L}_a(\phi,\psi) = \mathbb{E}_{t,\bm p_0,\bm \epsilon}[\|\bm \epsilon - \bm \epsilon_{\phi,\psi}(\bm p_t,\bm o)\|^2].
\end{align}

Once the reverse process is trained, multiple trajectories are sampled by utilizing the predictor to infer from the initial noise. 
For simplicity and conservatism, it is assumed that the future of each joint is predicted in an independent manner. 
Consequently, the entire inference and statistical process is encapsulated in the following equation:
\begin{align}
    \hat{\bm p}_0, \hat{\bm\Sigma}_p =& f_a(\bm o),
\end{align}
where $\hat{\bm p}_0 = \{ \hat{\bm q}_{h}^{(t)} \in \Re^{5} | t = 1,2,\cdots, N_{p} \}$ represents the estimated motion intention, and $\hat{\bm\Sigma}_p = \{ diag(\hat{\bm\sigma}_{h}^{(t)}) \in \Re^{5\times5} | t = 1,2,\cdots, N_{p} \}$ represents the standard deviation of the prediction.
\\
\begin{algorithm}[t]
\caption{Preemptive Tuning for the Reference Trajectory}
\begin{algorithmic}
\STATE \bf {Given}: \(\hat{\bm p}_0, \hat{\bm\Sigma}_p,\bm q, \varepsilon\)  
\STATE $\bm q_r^{(0)} = {\rm Saturate}(\hat{\bm q}_{h}^{(0)},\bm q_{\min},\bm q_{\max})$  
\FOR{$t = 1$ \TO $N_p$}  
    \FOR{$i = 1$ \TO $5$}  
        \STATE $\delta_i^{(t)} = \min(\|\hat{q}_{hi}^{(t-1)} - q_{\min,i}\|, \|q_{\max,i} - \hat{q}_{hi}^{(t-1)}\|)$
        \IF{$\frac{\hat{\sigma}_{hi}^{(t)}}{\hat{\sigma}_{hi}^{(t)} + (\delta_i^{(t)})^2} \leq \varepsilon$}
            \STATE $q_{ri}^{(t)} = \hat{q}_{hi}^{(t)}$
        \ELSE
            \STATE $q_{ri}^{(t)} = q_{ri}^{(t-1)}$
        \ENDIF
    \ENDFOR
\ENDFOR
\STATE Output: $\bm q_r^{(0)}$ to $\bm q_r^{(N_p)}$
\end{algorithmic}
\label{Preemptive}
\end{algorithm}

\noindent\textbf{Preemptive Tuning}: Given the distribution of a predicted future trajectory, it is imperative to preemptively tune the reference trajectory to ensure compliance with established constraints.
Specifically, for each joint $i$, denoted by subscript, the feasible trajectory must hold the following probabilistic inequality:
\begin{align}
    \mathcal{P}({q}_{hi}^{(t)}\notin 
 \mathcal{Q}_i)<&\varepsilon,
\end{align}
where $\mathcal{Q}_i$ is the pre-defined permissible motion range $[{q}_{\min,i}, {q}_{\max,i} ]$.
It is stipulated that $\hat{q}_{hi}^{(0)} = q_i$, where $q_i$ is the current position of joint $i$ of the upper-limb exoskeleton robot.
In accordance with Cantelli’s inequality, the predicted distribution is characterized by:
\begin{align}
    &\mathcal{P}({q}_{hi}^{(t)}\geq \hat{q}_{hi}^{(t)} + \delta_i^{(t)})\leq\frac{\hat{\sigma}_{hi}^{(t)}}{\hat{\sigma}_{hi}^{(t)} + (\delta_i^{(t)})^2},\\
    &\delta_i^{(t)} = \min(\|\hat{q}_{hi}^{(t-1)}-{q}_{\min,i}\|,\|{q}_{\max,i}-\hat{q}_{hi}^{(t-1)}\|).
\end{align}
Now, we introduce the following tuning law for prediction:
\begin{align}
    q_{ri}^{(t)} = \begin{cases}
    \hat{q}_{hi}^{(t)}, & \text{if } \frac{\hat{\sigma}_{hi}^{(t)}}{\hat{\sigma}_{hi}^{(t)} + (\delta_i^{(t)})^2} \leq \varepsilon, \\
    q_{ri}^{(t-1)}, & \text{otherwise}.
    \end{cases}
\end{align}
Here, $q_{ri}^{(t)}$ is the reference trajectory of joint $i$ at time-step $t$.
It updates to the predicted motion intention $\hat{q}_{hi}^{(t)}$ when the movement trend is within the established boundaries.
However, if a potential boundary violation is detected, it maintains trajectory from the previous time-step to prevent the limit being crossed.
The aforementioned process is also detailed in Algorithm \ref{Preemptive}.
The adjusted reference trajectory $\bm q_r$ offers an accurate estimate of the intended motions of the unaffected side of the body, guaranteeing that safety considerations are adequately addressed. 
By resolving (\ref{opt_all}), it is feasible to generate a trajectory for the upper-limb exoskeleton robot that is not only smooth and characterized by minimal acceleration but also ensures the safety of the desired future trajectory.



\subsection{Passive Training Mode}
In cases of significant motor impairment, the upper-limb exoskeleton robot is engaged in passive following training. This mode of training is essential for assisting the patient to follow a pre-defined trajectory.
Moreover, to promote rehabilitation, it is crucial to individualize the assistance in the regulation of assist-as-needed. The workflow of passive following training is depicted in Figure \ref{passiveBD}. In this mode, motion data from the upper limbs of healthy subjects are collected to facilitate the training of two distinct modules.\\

\begin{figure}[!h]
    \centering
    \includegraphics[width=0.9\linewidth ]{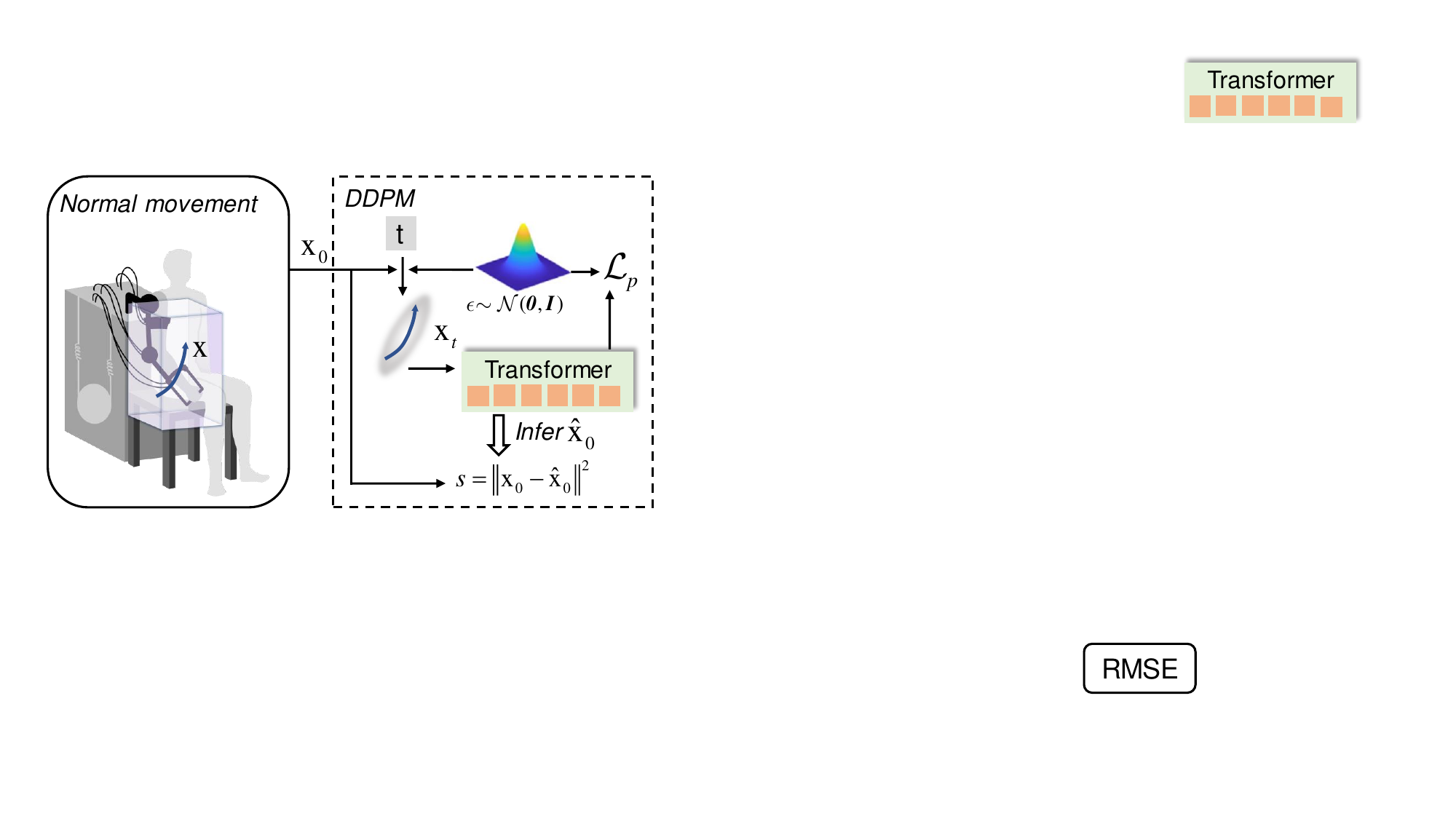}
    \caption{Illustration of the anomaly detector: The transparent rectangular prism delineates the feasible workspace, encompassing the range within which upper-limb motion is categorized as normal and safe.
    These movements incorporate position, velocity, and interaction torque information, which are collectively depicted as a curved blue arrow and represented by $\mathbf{x}$.}
    \label{ano_net}
    \vspace{-0.3cm}
\end{figure}

\noindent\textbf{Anomaly Detector}: To individualize assistance, a real-time criterion is needed for evaluating the assistive trajectory. We use an anomaly detector to quantify the comfort of wear and the effectiveness of rehabilitation.
This anomaly detector is based on a diffusion model architecture and identifies irregular patterns in upper-limb movements, as illustrated in Figure \ref{ano_net}.
Subsequently, the anomaly detector computes a score that quantifies the deviation between the current human-robot interaction and a natural interaction condition, thereby guiding the individualization of assistance.
In this approach, sensory feedback at a given timestep $i$, captured through a sliding window mechanism, is fed into the anomaly detector, denoted as $\bm{\mathbf{x}}^{(i)}\in\hspace{-0.05cm}\Re^{L_{s}N_c}$, where $L_s$ is the width of the sliding window, and $N_c$ is the number of data channels.

As mentioned, the anomaly detector is based on a diffusion model, wherein the reverse diffusion process of duration $T^p$ is delineated as the sequence $(\bm{\mathbf{x}}_{T^p}, \bm{\mathbf{x}}_{T^p-1}, \cdots, \bm{\mathbf{x}}_0)$.
The associated diffusion and reverse diffusion processes are defined as follows:
\begin{align}
    q(\bm{\mathbf{x}}_{t}|\bm{\mathbf{x}}_{t-1}) &= \mathcal{N}(\bm{\mathbf{x}}_{t};\sqrt{1-\beta_{t}^p}\bm{\mathbf{x}}_{t-1},\beta_{t}^p\bm I),\label{diffusion_process}\\
    q_{\Psi}(\bm{\mathbf{x}}_{t-1}|\bm{\mathbf{x}}_{t}) &= \mathcal{N}(\bm{\mathbf{x}}_{t-1};\bm\mu_{\Psi}(\bm{\mathbf{x}}_{t}),\Tilde{\beta}_{t}^p\bm I),\label{reverse_process_AD}
\end{align}
where $\beta_1^p,\beta_2^p,\cdots,\beta_{T^p}^p$ are variance schedules utilized to modulate the level of noise injected during the process.
The adjusted variance, $\Tilde{\beta}_{t}^p$, is calculated using the formula $\Tilde{\beta}_{t}^p=\frac{1-\bar{\alpha}^a_{t-1}}{1-\bar{\alpha}_t^p}\beta_t^p$, and by exploiting the definitions $\alpha_t^p = 1-\beta_t^p$ and $\bar{\alpha}_t^p = \prod_{i=1}^{t}\alpha_i^p$.

In alignment with the concept outlined in the \textit{intention predictor}, the loss function used to train the diffusion model in passive following mode is specified as follows
\begin{align}
    \mathcal{L}_p(\Psi) = \mathbb{E}_{t,\bm{\mathbf{x}}_0,\bm \epsilon}[\|\bm \epsilon - \bm \epsilon_{\Psi}(\bm{\mathbf{x}}_t)\|^2].
\end{align}

Once the reverse process has been effectively learned from the dataset comprising upper-limb movements, the anomaly detector proficiently filters noise from contaminated sensory data to yield a clarified output.
This capability facilitates the generation of a refined sensory input through the anomaly detector.
this input is employed to compute the anomaly score, as delineated in Algorithm \ref{AD_ALG}, where $\bm \epsilon_p, \bm z_p\thicksim \mathcal{N}(\bm 0, \bm I)$ and $\nu \in [1, T^p]$ is a constant parameter. 
For the sake of brevity, the methodology for calculating the anomaly score is encapsulated by the following function:
\begin{align}
s = f(\bm q, \dot{\bm q}, \bm \theta, \dot{\bm \theta},\bm \tau_e).
\label{AD_function}
\end{align}

\begin{algorithm}[ht]
\caption{Anomaly Detection}
\begin{algorithmic}[1]
\REQUIRE $\bm q, \dot{\bm q}, \bm \theta, \dot{\bm \theta},\bm \tau_e$
\STATE Initialize sliding window queue
\FOR{each new data point}
    \STATE $\bm{\mathbf{x}}_{now}^{(i)} \gets (\bm q, \dot{\bm q}, \bm \theta, \dot{\bm \theta},\bm \tau_e)$
    \STATE Enqueue $\mathbf{x}_{now}^{(i)}$ to sliding window queue
    \IF{sliding window queue is full}
        \STATE $\mathbf{x}_0 \gets \text{GetWindowData}$
        \STATE $\bm \epsilon_p \gets \text{Sample}$
        \STATE $\hat{\mathbf{x}}_\nu \gets \mathbf{x}_0 \sqrt{\bar{\alpha}^p_\nu} + \bm \epsilon_p\sqrt{1-\bar{\alpha}^p_\nu}$
        \FOR{$t=\nu,\cdots,1$}
            \STATE $\bm z_p \gets \text{Sample}()$
            \STATE $\hat{\mathbf{x}}_{t-1} \gets \frac{1}{\sqrt{\alpha^p_t}}(\hat{\mathbf{x}}_{t}-\frac{1-\alpha^p_t}{\sqrt{1-\bar{\alpha}^p_t}}\bm \epsilon_{\Psi}(\hat{\mathbf{x}}_t))+\Tilde{\beta}_{t}^p\bm z_p$
        \ENDFOR
        \STATE $s \gets {||\mathbf{x}_0- \hat{\mathbf{x}}_0||}^{2}$
        \STATE Output: $s$
        \STATE Dequeue from sliding window queue
    \ENDIF
\ENDFOR
\end{algorithmic}
\label{AD_ALG}
\end{algorithm}

The anomaly detector integrates diffusion models and thus is adept at capturing the inherent spatiotemporal patterns and stochastic motion tendencies of upper-limb movements through analysis of sensor data.
Consequently, anomaly scores can be computed in real time for human–robot interactions.
These scores serve as indicators of the comfort levels and the naturalness of the assistance provided by the upper-limb exoskeleton robot.\\

\noindent\textbf{Reference Generation}: 
To customize assistance during passive following training, historical upper-limb trajectories, coupled with online trajectory refinement as outlined in (\ref{opt_all}) are integrated to develop a probabilistic model. 
In this context, ProMPs are employed to encode a set of trajectories into a probabilistic framework \citep{paraschos2013probabilistic}, which is capable of generating similar references through sampling.
The application of ProMPs for trajectory sampling is particularly suited for analyzing the repetitive movements encountered in passive following training.
This suitability is attributable to the probabilistic model’s effective accommodation of sensor noise, human uncertainty, and individual biases.

To implement the ProMPs, we express the trajectory by means of the weight vector $\bm \omega \in \Re^{Dn\times1}$, where $D$ is the number of basis functions, and $n$ is the number of active joints, such that
\begin{align}
 \bm{y}_t &=  \left[ 
\begin{array}{ccc}
\bm q_{1,t}^{\mathsf{T}} & \cdots & \bm q_{n,t}^{\mathsf{T}}
\end{array} 
\right]^{\mathsf{T}}=\begin{bmatrix}\bm\Phi_t\\\dot{\bm\Phi}_t\end{bmatrix}\bm\omega+\bm\epsilon_y,\\
\bm q_{i,t} &= \left[ 
\begin{array}{cc}
q_{i,t}  & \dot{q}_{i,t}
\end{array} 
\right]^{\mathsf{T}},\\
p(\bm\tau_y|\bm w) &= \prod_t\mathcal{N}(\bm{y}_t|\bm\Phi_t \bm \omega, \bm \Sigma_y),
\end{align}
where $\bm q_{i,t}\in\Re^{2}$ represents the composite vector of the $i^{th}$ joint at time step $t$, $\bm\epsilon_y\sim \mathcal{N}(\bm 0,\bm \Sigma_y)$ represents zero-mean i.i.d. Gaussian noise, $\bm\tau_y$ is the trajectory over the demonstration, and $\bm\Phi_t \in \Re^{n\times Dn}$, chosen as a Gaussian form, is the time-variant basis matrix.

Given the assumption that $\bm\omega$ follows a normal distribution, $\bm\omega \sim \mathcal{N}(\bm\omega|\bm \mu_{\omega}^{(k)},\bm \Sigma_{\omega}^{(k)})$, a new trajectory at time step $t$ can be modeled as follows:
\begin{align}
p(\bm y_t;\bm \mu_{\omega}^{(k)},\bm \Sigma_{\omega}^{(k)}) = \int\mathcal{N}(\bm{y}_t|\bm\Phi_t \bm \omega, \bm \Sigma_y)\mathcal{N}(\bm\omega|\bm \mu_{\omega}^{(k)},\bm \Sigma_{\omega}^{(k)})d\bm\omega.\label{proModel}
\end{align}
Therefore, the reference trajectory is given as follows:
\begin{align}
\bm q_r(t) = [q_{i,t},\cdots, q_{n,t},]
\end{align}

To facilitate the generation of personalized assistance, the following cost function is introduced to evaluate the performance of the provided reference trajectory:
\begin{align}
\mathcal{S}(\bm q_r) = \int_{T_r} \{\|\bm q_{d}(\bm q_r) - \bm q\|_{\bm Q}^2+s^2\}dt,
\label{cost_qr}
\end{align}
where $T_r$ signifies the duration of the assistance, and $\bm q_{d}$ denotes the desired trajectory obtained from online refinement via (\ref{opt_all}). 

Given that each sampled trajectory can be evaluated by its cost, it becomes feasible to attribute an information-theoretic weight to them, reflecting the performance across all $k$ trajectories \citep{williams2017information}:
\begin{align}
w^{(i)} = \frac{1}{\eta_k}\exp(-\frac{1}{\lambda_p}\mathcal{S}(\bm q_r^{(i)})),
\label{promp_weight}
\end{align}
where $\bm q_r^{(i)}$ is the reference trajectory from the $i^{th}$ sampling, $\eta_k$ is a normalization constant, and $\lambda_p$ is a small positive constant. 
A detailed analysis of the adopted weight setting is included in the Appendix.

Considering the performance of the trajectory, the parameters $\bm \mu_{\omega}^{(k)}$ and $\bm \Sigma_{\omega}^{(k)}$ are deduced from the $k$ collected trajectories, as follows:
\begin{align}
\bm \omega^{(i)} &= (\bm\Phi^{\mathsf{T}}\bm\Phi)^{-1}\bm\Phi^{\mathsf{T}}\bm Y^{(i)},\label{ProMP_omega_i}\\
\bm \mu_{\omega}^{(k)} &= \sum_i w^{(i)}\bm \omega^{(i)},\label{promp_mu}\\
\bm \Sigma_{\omega}^{(k)} &= \sum_i w^{(i)}(\bm \omega^{(i)} - \bm \mu_{\omega}^{(k)})(\bm \omega^{(i)} - \bm \mu_{\omega}^{(k)})^{\mathsf{T}}\label{promp_sigma},
\end{align}
where $\bm\Phi \in \Re^{N_r n\times Dn}$ is a matrix comprised of block diagonal matrices $\bm\Phi_t$, stacked vertically in accordance with sampling number $N_r$, and $\bm Y^{(i)}\in \Re^{N_r n}$ corresponds to the $i^{th}$ gathered trajectory.
This process facilitates the construction of the probabilistic model.

The adjusted trajectory is integrated into the upper-limb exoskeleton robot to assist the patient.
Subsequently, the patient’s actual movement during the training sessions is captured and utilized to iteratively refine the probabilistic model (\ref{proModel}). 
Specifically, 
$\bm \mu_{\omega}^{(k)} \leftarrow \bm \mu_{\omega}^{(k+1)}$ and $\bm \Sigma_{\omega}^{(k)} \leftarrow \bm \Sigma_{\omega}^{(k+1)}$, which is then used
to facilitate the planning of subsequent trajectories, as illustrated in Algorithm \ref{ProMP_ALG}. 
$N_s$ is the number of times that free exploration is performed based on the pre-collected demonstrations.
The refinement of assistance is governed by the anomaly score, reflecting prior healthy movement behavior. 
This score serves as an indicator of movement comfort and naturalness during rehabilitation, ensuring that the trajectory adjustments intrinsically enhance the quality of assistance.

The optimal assistance distribution is identified via a coarse-to-fine approach.
Initially, the sample distribution is established based on pre-defined demonstrations, which primarily guide the exploratory phase using the initial samples.
Subsequent iterations improve the assistance distribution by progressively narrowing the sampling space and utilizing previously explored optimal trajectory values. 
The probability model focuses on the best-performing trajectories to provide optimal assistance, which effectively addresses the various uncertainties that the patient may encounter during the task.
Incorporation of the patient’s motion data ensures that subsequent trajectories become increasingly individualized and thus align increasingly more with the patient’s medical needs.
Hence, this method enables the upper-limb exoskeleton robot to exploit online interactions for the dual purpose of enhancing interaction safety and improving the efficacy of passive following training.

\begin{algorithm}[ht]
\caption{Reference Trajectory Generation}
\begin{algorithmic}[1]
\REQUIRE $\bm Y^{(k+1)}$
\IF{$k\leq N_s$}
\STATE Initialize $\bm \mu_{\omega}^{(k+1)}, \bm \Sigma_{\omega}^{(k+1)}$ with pre-collected data.
\ELSE
\STATE Calculate cost $\mathcal{S}(\bm q_r^{(k+1)})$ using Eq(\ref{cost_qr})
\FOR{all collected $k+1$ trajectories}
\STATE Calculate cost $w^{(i)}, \bm \omega^{(i)}$ using Eq(\ref{promp_weight}) and Eq(\ref{ProMP_omega_i})
\ENDFOR
\STATE Calculate distribution $\mathcal{N}(\bm\omega|\bm \mu_{\omega}^{(k+1)},\bm \Sigma_{\omega}^{(k+1)})$ using Eq(\ref{promp_mu}) and Eq(\ref{promp_sigma})
\ENDIF
\STATE $\bm \omega \gets \text{Sample}(\mathcal{N}(\bm\omega|\bm \mu_{\omega}^{(k+1)},\bm \Sigma_{\omega}^{(k+1)}))$
\STATE $\bm q_r(t) \gets \bm\Phi_t \bm\omega$
\end{algorithmic}
\label{ProMP_ALG}
\end{algorithm}

\section{Interaction Control}

The joint below the shoulder joints is cable-driven, which decreases the inertia associated with movement and thus increases comfort. 
Given the presence of these joints, there is substantial friction, which can hinder the movement of joints and therefore should be considerably compensated for.

To achieve this, the upper-limb exoskeleton robot first performs movements in the absence of the patient, where $\bm\tau_e\hspace{-0.05cm}=\hspace{-0.05cm}\bm 0$. In addition, the disturbance torque can be parameterized as follows \citep{de1995new}:
\begin{align}
\bm\tau_f &= (\bm a_f + \bm b_f \odot e^{-\bm c_f \odot\dot{\bm q}} + \bm d_f\odot\dot{\bm q})\odot\bm {sgn}(\dot{\bm q})\notag \\
& \approx (\bar{\bm a}_f + \bar{\bm b}_f\odot\dot{\bm q} + \bar{\bm c}_f\odot\dot{\bm q}\odot\dot{\bm q})\odot\bm {sgn}(\dot{\bm q})= \bm Y(\dot{\bm q})\bm \zeta,\label{friction_model}
\end{align}
where $\bm a_f, \bm b_f, \bm c_f, \bm d_f$ are the unknown parameters, $\bar{\bm a}_f, \bar{\bm b}_f, \bar{\bm c}_f$ are derived from the Taylor expansion as simplifications for the model,
$\odot$ denotes the Kronecker product, $\bm {sgn}(\cdot)$ represents a sign function, $\bm Y(\cdot)$ represents a regressor matrix, and $\bm\zeta$ represents the vector of model parameters. 
The approximation presented in (\ref{friction_model}) is reasonable because the velocities of the joints of the upper-limb exoskeleton robot remain rather low during a given rehabilitation process.

As the upper-limb exoskeleton robot is equipped with force sensors on all of its active joints, friction can be directly measured and recorded in the absence of a patient.
Together with the recorded joints’ velocities, the parameters of friction model are learned via polynomial fitting. The estimated friction is represented as follows
\begin{align}
\hat{\bm\tau}_f &= \bm Y(\dot{\bm q})\hat{\bm \zeta}.\label{est_friction}
\end{align}

Once the friction is estimated, a variable impedance model is proposed. This model must be capable of identifying the human–robot interaction condition and addressing the conflict during rehabilitation.
These capabilities enable the model to regulate the action of the upper-limb exoskeleton robot.

To regulate the impedance model, a weighting function \citep{zhang2023multi} is introduced to consider the anomaly score. This function is mathematically defined as follows:
\begin{eqnarray}
&w(s) = \lambda_1 \tanh(-\frac{s}{\chi_1} + \chi_2) +\lambda_2, \label{weight_fun}
\end{eqnarray}
where $\lambda_1$ and $\lambda_2$ are positive constants that determine the range and median of the weighting function, respectively, 
$\chi_1$ is a constant that normalizes the anomaly score into a specified small range, and $\chi_2$ is the offset of the weighting function from the origin of the coordinates along the positive horizontal axis.
Based on this function, the desired impedance model is redefined as follows:
\begin{eqnarray}
&\bm C_d(\dot{\bm q}-\dot{\bm q}_d)+\bm K_d(\bm q-\bm q_d)=\frac{1}{w(s)}\bm\tau_e.\label{model}
\end{eqnarray}

Multiplying both sides of (\ref{model}) by $w(s)$ yields
\begin{eqnarray}
&\bm C_{a}(t)(\dot{\bm q}-\dot{\bm q}_d)+\bm K_{a}(t)(\bm q-\bm q_d)=\bm\tau_e,\label{model2}
\end{eqnarray}
where $\bm C_a(t)\stackrel{\triangle}{=}w(s)\bm C_d$ and $\bm K_a(t)\stackrel{\triangle}{=}w(s)\bm K_d$, are the time-varying apparent impedance parameters. These parameters are utilized to deduce the relationship between the desired trajectory and the interaction torque, as outlined in (\ref{opt_all}). 
The function of the mechanism is such that an increase in the anomaly score leads to a decrease in $w(s)$, thereby reducing the magnitude of the impedance parameters. 
This reduction ensures that there is an increase in the passiveness with which the upper-limb exoskeleton robot responds to any detected conflicts.
Conversely, when the anomaly score is low, the impedance parameters revert to their original values, thereby maintaining the level of assistance provided by the upper-limb exoskeleton robot.

Subsequently, an impedance vector is introduced, as follows
\begin{align}
\bm z&=\dot{\bm q}-\dot{\bm q}_r\nonumber \notag \\
&=\dot{\bm q}-\dot{\bm q}_d+\bm C_d^{-1}\bm K_d(\bm q-\bm q_d)-\frac{1}{w(s)}\bm C_d^{-1}\bm\tau_e,\label{vectorz}
\end{align}
where 
\begin{eqnarray}
&\dot{\bm q}_r=\dot{\bm q}_d-\bm C_d^{-1}\bm K_d(\bm q-\bm q_d)+\frac{1}{w(s)}\bm C_d^{-1}\bm\tau_e
\end{eqnarray}
is a reference vector. According to (\ref{vectorz}), the convergence of $\bm z\rightarrow\bm 0$ implies the realization of the desired impedance model (\ref{model}).

The overall control input is designed as defined in (\ref{controlSection3}), with the fast time-scale control term defined as in (\ref{uf}). 
Next, the slow time-scale control term is established by using the estimated friction $\hat{\bm\tau}_f$ to stabilize the dynamics expressed in (\ref{substitute_SP4sec}) and achieve the desired impedance model, as follows:
\begin{align}
    \bm u_s=&-\bm K_z\bm z-\bm S_2^{\mathsf{T}}\hat{\bm\tau}_f-{\bm\tau}_e-k_g\cdot {\bm{sgn}}(\bm z) \notag \\
    &+(\bm M(\bm q)+\bar{\bm B})\ddot{\bm q}_r+\bm C(\dot{\bm q}, \bm q)\dot{\bm q}_r
+\bm g(\bm q),\label{slow_controller}
\end{align}
where ${\bm{sgn}}(\cdot)$ is the sign function and is defined as follows:
\begin{eqnarray}
{\rm sgn} (z) = 
\left\{ 
\begin{array}{*{20}{l}}
 1,&& z >0\\
 0,&& z= 0\\
 - 1,&& z<0
\end{array} 
\right.
\end{eqnarray}
where $k_g$ is a positive constant, and 
$\bm K_z\in\Re^{n\times n}$ is a diagonal and positive-definite matrix.

The proposed variable impedance controller, as delineated in (\ref{controlSection3}), (\ref{uf}), and (\ref{slow_controller}), can be demonstrated to be exponentially stable, as shown by the stability analysis provided in the Appendix.


\section{Experiments}
\begin{figure}[!h]
  \centering
    \includegraphics[width=0.9\linewidth]{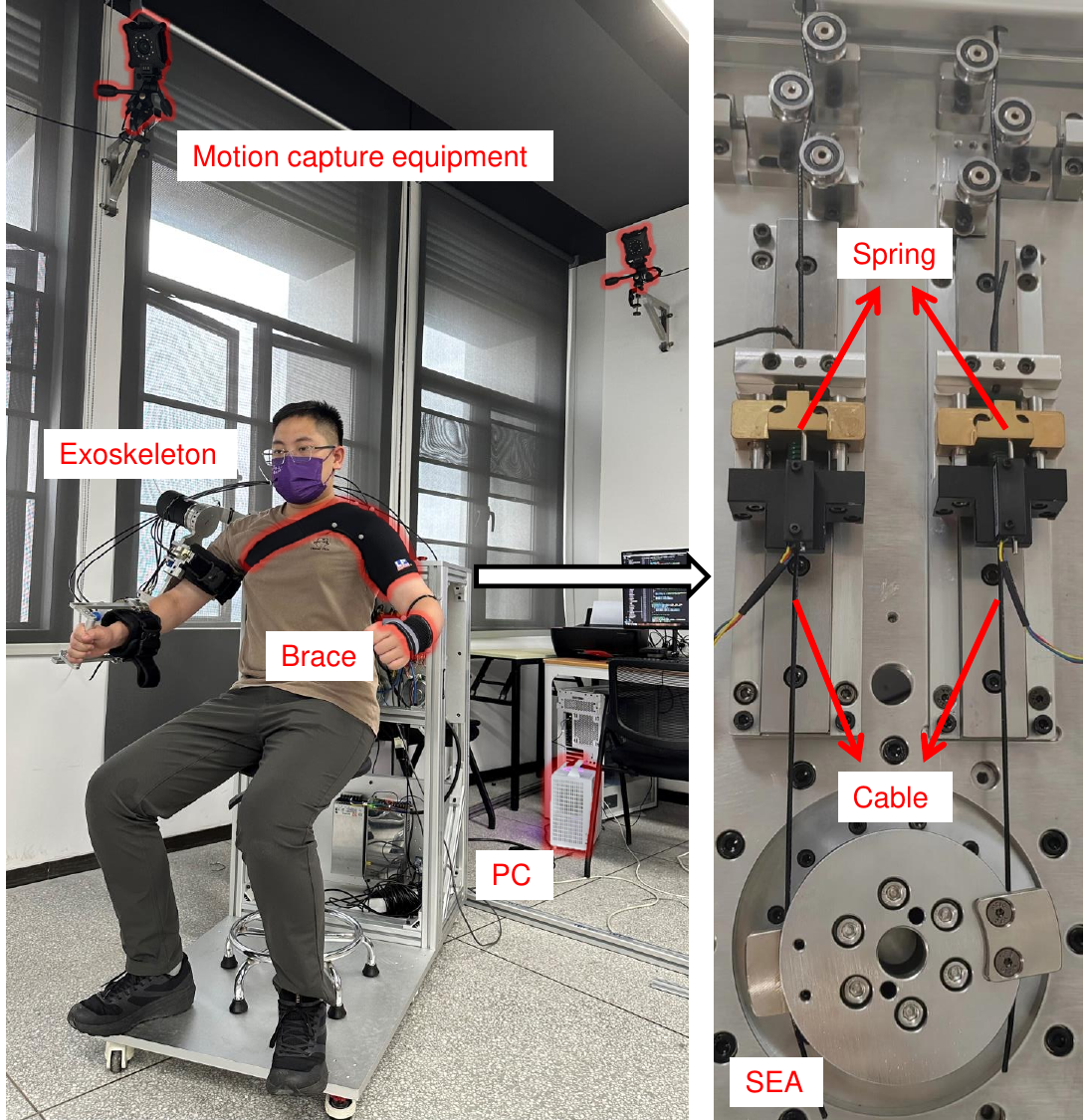}
  \caption{The experimental setup in active mirroring mode used motion capture equipment to obtain feedback from the unaffected side of the body. The SEA was constructed by connecting the motor output and the joint end with a spring and cable.}
  \label{exp_setup} 
\end{figure}

The proposed dual-mode trajectory refinement method was implemented within the upper-limb exoskeleton robot to assess the effectiveness of both rehabilitation modes.
Figure \ref{exp_setup} depicts the experimental configuration, in which the main board controlled the impedance of the upper-limb exoskeleton robot.
This board was interfaced via a serial port connection with a PC, which was outfitted with an Intel i5-13490F CPU and an RTX 4060Ti graphics card.
The PC executed the anomaly detection module, calculated the reference trajectory $\bm q_r$, and dynamically updated the desired trajectory $\bm q_d$ online. 
Subjects performed rehabilitation training in passive following mode or active mirroring mode.
In the latter mode, the subject wore a brace equipped with markers on the unaffected side of the body to facilitate engagement in rehabilitation exercises.
Optical motion capture equipment (Nokov) was used to accurately capture motion intentions.

Implementation of the variable impedance controller required knowledge of the dynamic model.
Thus, the dynamic parameters were computed analytically in real time using the open-source Orocos Kinematics and Dynamics Library\footnotemark. \footnotetext{\url{https://www.orocos.org/wiki/orocos/kdl-wiki.html}}
In this computation, the upper-limb exoskeleton robot is deconstructed into a sequence of links and joints to formulate a model defining its physical configuration, encompassing characteristics such as the length, mass, and inertia of each link.
Subsequently, the forward dynamics are derived using Newton–Euler equations based on this model.

The intention predictor and the anomaly detection module required motion data, which was collected from the upper-limb exoskeleton robot during operation by healthy subjects.
To facilitate free and natural movement, the upper-limb exoskeleton robot was set to operate in a transparent mode during the data collection phase \citep{tro_2024}. 
Specifically, the controller was designed as follows \citep{zimmermann2020towards}:
\begin{align}
    \bm u_{0} = &(\bm M(\bm q)+\bar{\bm B})\ddot{\bm q}_0+\bm C(\dot{\bm q}, \bm q)\dot{\bm q}\notag\\
    &+\bm g(\bm q)-\bm S_2^{\mathsf{T}}\hat{\bm\tau}_f-  {\bm\tau}_e + \bm u_f,\\
    \ddot{\bm q}_0 = &\frac{1}{\gamma_0}\bm (\bm M(\bm q)+\bar{\bm B})^{-1}{\bm\tau}_e,
\end{align}
where $\gamma_0$ represents a parameter controlling the magnitude of virtual mass, and $\ddot{\bm q}_0$ denotes the desired acceleration. In this transparent mode, healthy subjects or patients were able to maneuver the upper-limb exoskeleton robot effortlessly, without experiencing significant discomfort.

The next section presents the results of five experiments, which are described below.
\begin{enumerate}
\item [-]\textit{Intention Predictor}: 
This experiment aimed to validate the efficacy and accuracy of the proposed intention predictor. A diverse set of intention prediction methods were trained using the collected data, and then a comparative analysis was performed to demonstrate the superiority of the proposed intention predictor.
\item [-]\textit{Anomaly Detection}: 
This experiment aimed to evaluate the performance of the anomaly detection network.
Its detection accuracy was demonstrated in various simulated anomaly scenarios, including movements outside the normal range, stroke-induced convulsions, and human–robot interaction conflicts.
Furthermore, comparative studies were conducted to illustrate that the proposed anomaly detector, which is based on a diffusion model, exhibits detection accuracy that is significantly better than that exhibited by other anomaly detection methods.
\item [-]\textit{Interaction Control}: 
This experiment aimed to assess the dynamic capabilities of the system and its ability to reject disturbances.
Thus, a trajectory tracking task was conducted. In addition, the efficacy of the proposed variable impedance controller was validated in a scenario involving anomalies.
\item [-]\textit{Active Mirroring Training}: 
This experiment aimed to assess the effectiveness of the online trajectory refinement. 
Thus, ablation studies were conducted. In addition, how the anomaly score influences the trajectories generated by this refinement process was examined. 
Furthermore, an active mirroring mode was implemented, and the motion capture system was used to verify that the proposed method effectively constrains assistive trajectories and maintains safety throughout a rehabilitation process.
\item [-]\textit{Passive Following Training}: 
This experiment aimed to validate the improvements in rehabilitation facilitated by passive following training.
An ablation study was conducted to demonstrate that the online trajectory refinement significantly enhances movement naturalness and task performance throughout the training process.
Moreover, a clinical trial was performed with stroke patients to obtain evidence that this rehabilitation framework significantly aids in the recovery of motor functions.
\end{enumerate}

\section{Results}

Two able-bodied participants with no prior experience with upper-limb exoskeleton robots were recruited for motion data collection.
Table \ref{joint_range} presents the motion range of the upper-limb exoskeleton robot during data collection. This motion range covers the main spatial areas of upper-limb daily activities.


\begin{table}[!h]
\caption{Motion Range of Joints}
\centering
\begin{tabular}{cccccc} 
\toprule
Joint & 1  & 2 & 3 & 4 & 5  \\ 
\midrule
Min($^{\circ}$)   & -90   & -45 & -30 & -5 & -30         \\
Max($^{\circ}$)  & 30   & 115  & 30  & 120 & 30     \\
\bottomrule
\end{tabular}
\label{joint_range}
\end{table}

Table \ref{participants} presents an overview of the participants’ statistical attributes.
The participants signed a written informed consent form prior to the experimental sessions.
Next, the participants underwent a 6-minute pre-training phase to familiarize themselves with moving while wearing the upper-limb exoskeleton robot.
During the data collection phase, no specific movement guidelines were imposed.
Therefore, the participants were permitted to move the upper-limb exoskeleton robot at their preferred speed, i.e., rapidly or slowly, or to maintain it in a stationary position. 
The collected data were compiled into a dataset that was utilized for training the intention predictor and anomaly detection modules.

\begin{table}[!h]
\caption{Statistical information of the subjects}
\centering
\begin{tabular}{ccccc} 
\toprule
Subject & Gender  & Age(y) & Weight(kg) & Height(cm)  \\ 
\midrule
1   & Male   & 30   & 87        & 181        \\
2 & Male   & 24   & 70        & 172        \\
\bottomrule
\end{tabular}
\label{participants}
\end{table}

\begin{figure*}[!ht] 
	\centering 
	\vspace{-0.3cm} 
	\subfigtopskip=2pt 
	\subfigbottomskip=2pt 
	\subfigcapskip=-5pt 
	\subfigure[]{
		\includegraphics[width=0.45\linewidth]{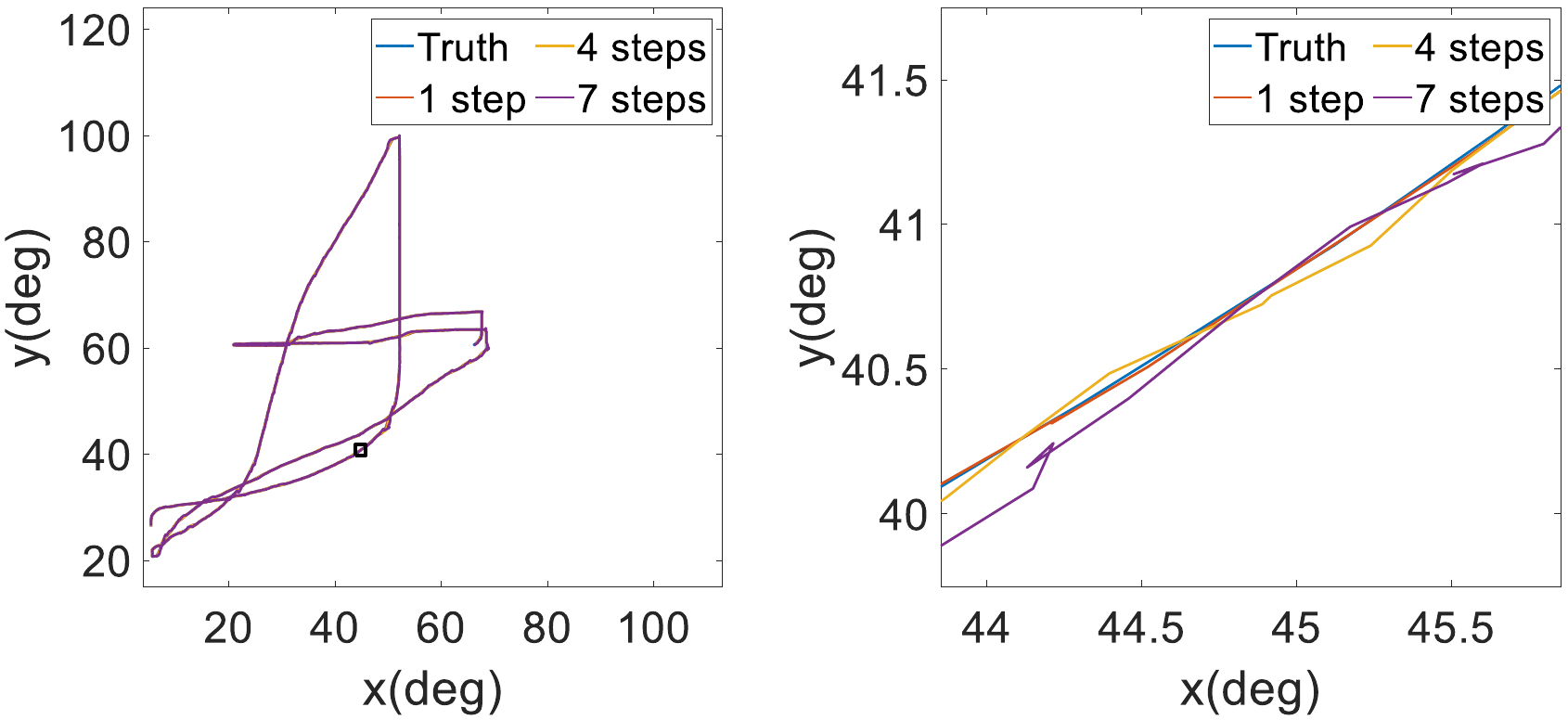}}
	\subfigure[]{
		\includegraphics[width=0.45\linewidth]{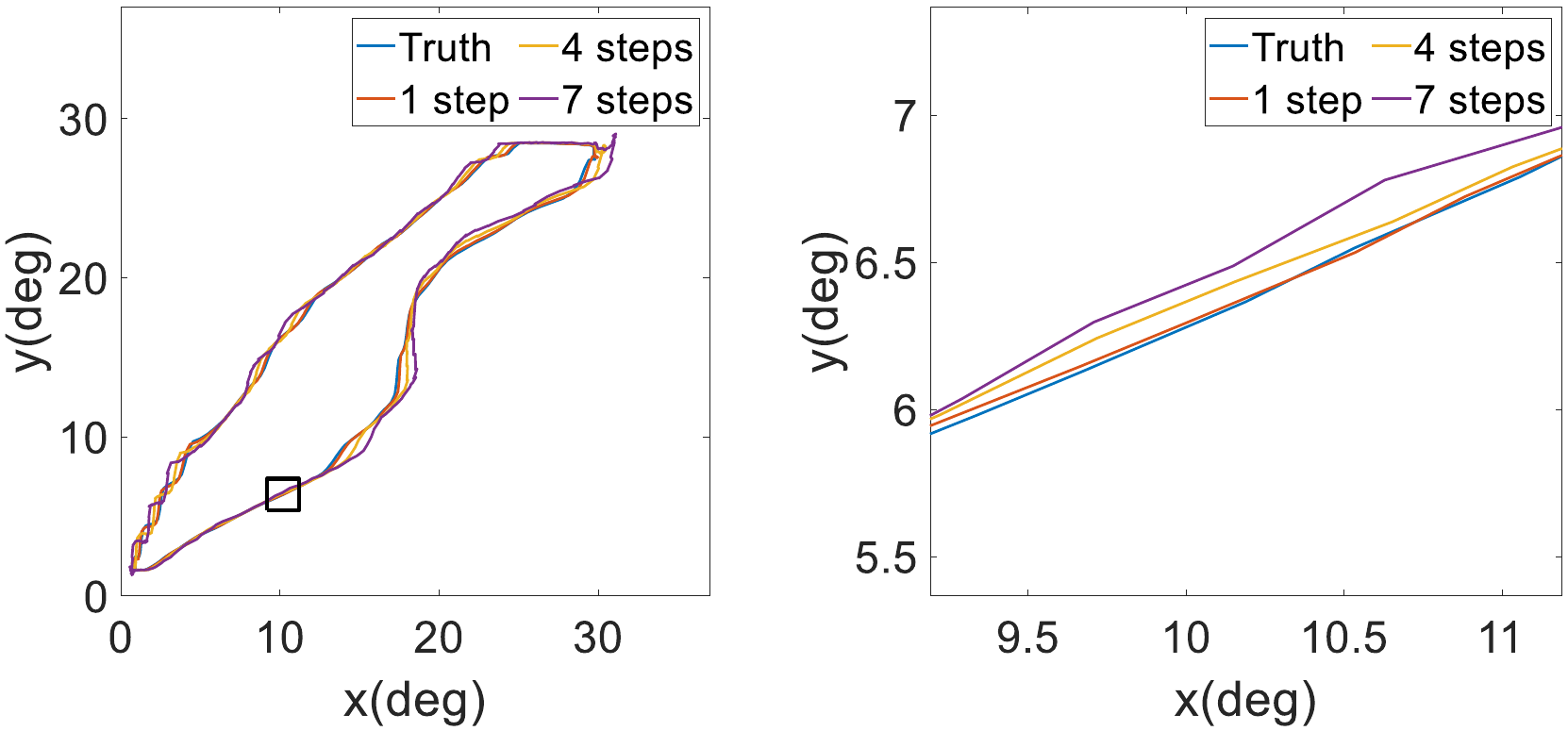}}
    \subfigure[]{
		\includegraphics[width=0.45\linewidth]{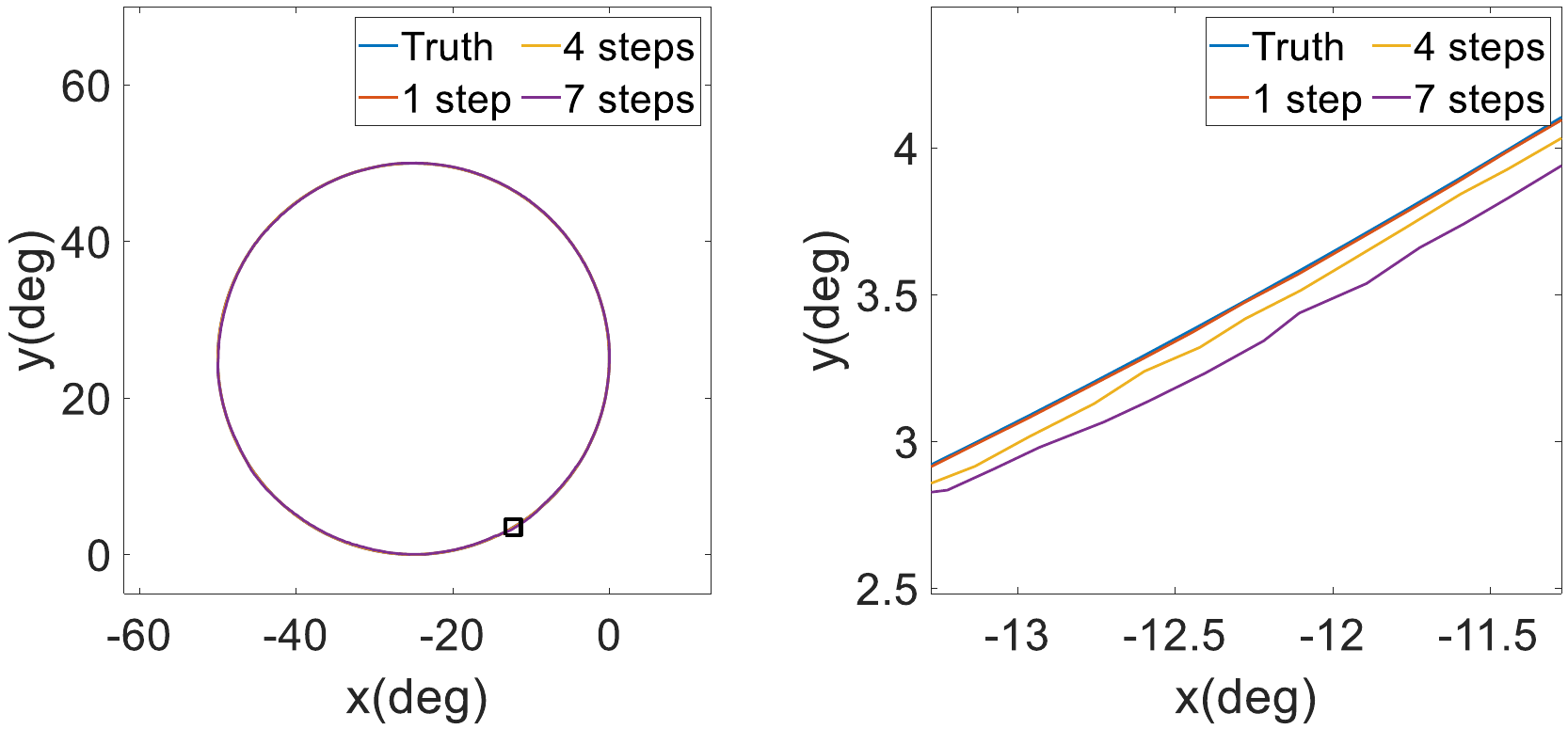}}
	\subfigure[]{
		\includegraphics[width=0.45\linewidth]{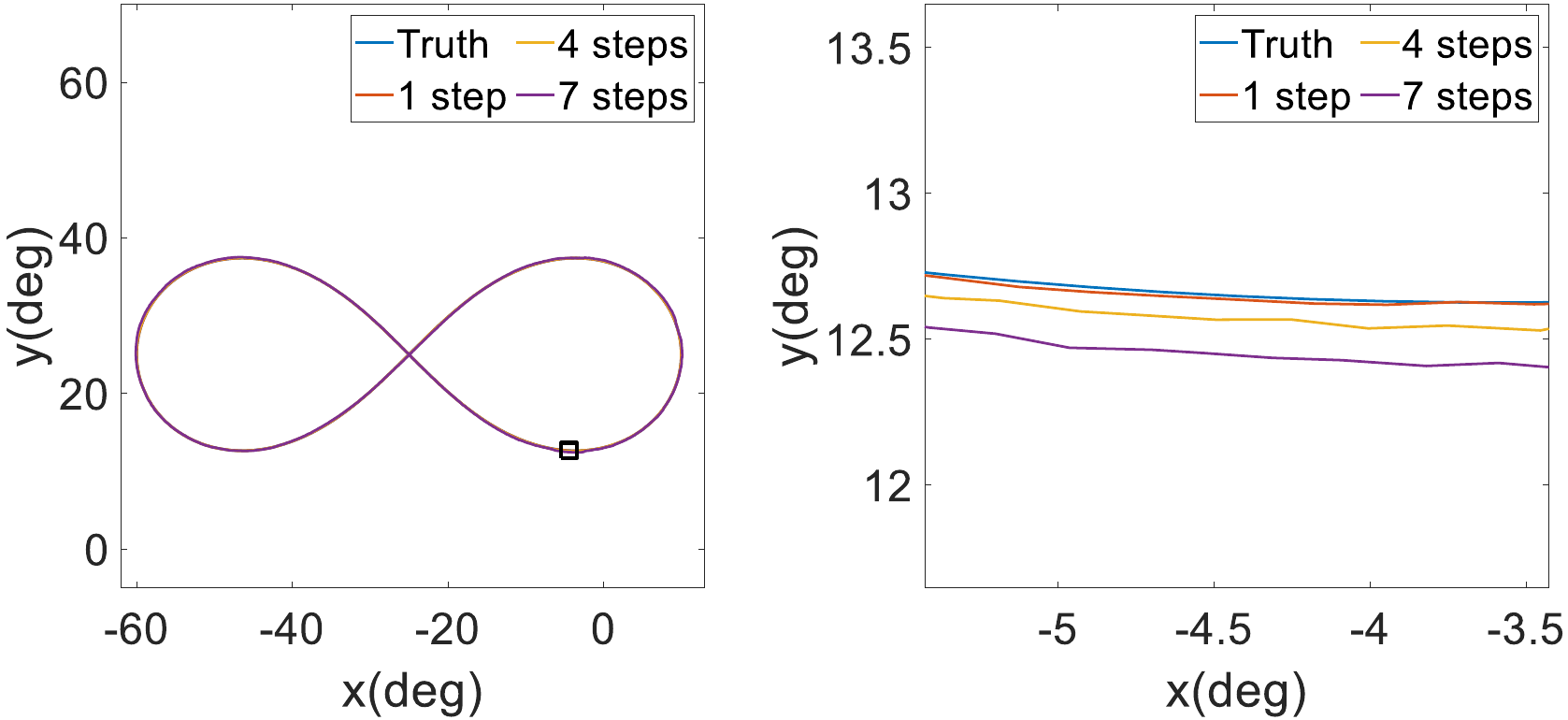}}
	\caption{Predicted trajectories for various steps across four distinct tasks. (a) Free move; (b) sinusoidal; (c) circular; and (d) lemniscate tasks.}
	\vspace{0cm}
	\label{1_4_7}
\end{figure*}

\subsection{Intention Predictor}
To achieve real-time estimation of human motion intention, the collected motion data were utilized to train the intention predictor.
To enhance the generalizability of the trained model, we partitioned the collected motion data into training, testing, and validation sets in an 8:2:1 ratio during the training process. 
We employed Trajectron++ \citep{salzmann2020trajectron++} as the backbone for extracting trajectory features and selected a transformer architecture to manage the diffusion process.
The parameters for the intention predictor were set as follows: $N_o = 5$, $N_p = 7$, and $T = 100$.
Denoising diffusion implicit models (DDIMs) \citep{song2020denoising} were employed in the inference process.
Intention prediction relies on historical observation data, and the predicted trajectory length was rather short.
Therefore, we selected forward integration and a convolutional neural network–long short-term memory (CNN-LSTM)-based neural network that previously demonstrated excellent performance in estimating robot joint behavior \citep{kim2019predicting} as the baselines for the experiments.

\begin{table}[!h]
\centering
\caption{Prediction Results for Different Methods}
\begin{tblr}{
  cell{5}{1} = {c=5}{},
  vline{2} = {1-4}{},
  hline{1-2,5} = {-}{},
  row{4} = {bg=gray!30},
}
Method              & FDE($^{\circ}$) & ADE($^{\circ}$) & MAE($^{\circ}$) & RMSE($^{\circ}$) \\
Forward-int. & 0.171   & 0.102    & 0.055   & 0.066                  \\
CNN-LSTM            & 0.165 $\uparrow$   & 0.099 $\uparrow$   & 0.056 $\downarrow$  & 0.067 $\downarrow$                  \\
Ours                & 0.096 $\uparrow$ & 0.061 $\uparrow$    & 0.033 $\uparrow$   & 0.040 $\uparrow$ \\
\end{tblr}
\label{validation_set}
\end{table}

Furthermore, we adopted final displacement error (FDE), average displacement error (ADE), mean absolute error (MAE), and root-mean-square error (RMSE) as the evaluation metrics to measure the accuracy of the predictions.
FDE and ADE were computed on a two-dimensional plane defined by Joint 2 and Joint 4, which possess the largest range of motion and exhibit the most frequent movements.
MAE and RMSE were averaged across all active joints. The predictions for the validation set are presented in Table \ref{validation_set}. 
These results suggest that in the considered real-time intention prediction task, the trajectory prediction performance of the CNN-LSTM-network-based method was on a par with that of the forward integration method.
However, our intention predictor based on a diffusion model significantly outperformed the forward integration method. 
Specifically, our intention predictor showed a 43.9$\%$ lower FDE, a 40.2$\%$ lower ADE, a 40.0$\%$ lower MAE, and a 39.4$\%$ lower RMSE than the forward integration method.


A comparative analysis was conducted to validate the applicability of our intention predictor across different movement conditions in four tasks.
Two tasks were based on motion data that were collected while the participants were wearing the upper-limb exoskeleton robot, including data on free movements similar to the training data and arm-swinging sinusoidal movements.
The other two tasks were based on simulated smooth trajectories, namely circular and lemniscate trajectories.
Not all tested trajectories were included in the pre-collected dataset employed in the training phase.
Table \ref{tasks} reports the performance of the intention predictor in the four tasks, averaged across five active joints.
It can be seen that our intention predictor exhibits the best performance in all metrics. 

\begin{table*}[!h]
\centering
\caption{Prediction Results for Different Tasks}
\begin{tblr}{
  cell{2}{1} = {r=3}{},
  cell{5}{1} = {r=3}{},
  cell{8}{1} = {r=3}{},
  cell{11}{1} = {r=3}{},
  cell{14}{1} = {c=6}{},
  vline{2-3} = {1-13}{},
  vline{3} = {3-4,6-7,9-10,12-13}{},
  hline{1-2,5,8,11,14} = {-}{},
  row{4,7,10,13} = {bg=gray!30},
}
Task       & Method              & FDE($^{\circ}$) & ADE($^{\circ}$) & MAE($^{\circ}$) & RMSE($^{\circ}$) \\
Free move  & Forward integration &   0.209 &   0.125 &   0.076   & 0.090\\
           & CNN-LSTM            & 0.163  &  0.098  &  0.060  &  0.073 \\
           & Ours                &   0.111 (46.9 $\%\uparrow$)  &  0.071 (43.2 $\%\uparrow$) &    0.042 (44.7 $\%\uparrow$) &   0.052 (42.2 $\%\uparrow$)\\
Sinusoidal & Forward integration  &  0.731 &   0.379 &    0.243 &    0.297\\
           & CNN-LSTM            &0.739 &   0.387  &  0.248 &   0.302 \\
           & Ours                 &  0.616 (15.7 $\%\uparrow$) &  0.331 (12.7 $\%\uparrow$) &    0.212 (12.8 $\%\uparrow$) &    0.258 (13.2 $\%\uparrow$) \\
Circular     & Forward integration &  0.078   &  0.037   &  0.024   & 0.029  \\
           & CNN-LSTM            &0.101 &   0.054  &  0.031 &    0.038 \\
           & Ours                 &   0.066 (15.4 $\%\uparrow$)  &   0.029 (21.6 $\%\uparrow$)  & 0.018 (25.0 $\%\uparrow$)    &   0.024 (17.2 $\%\uparrow$) \\
Lemniscate & Forward integration& 0.111    & 0.049    &  0.032   & 0.041  \\
           & CNN-LSTM            &0.095 &   0.045  &  0.028 &   0.036 \\
           & Ours                 & 0.086 (22.5 $\%\uparrow$)    &  0.035 (28.6 $\%\uparrow$)   &  0.022 (31.3 $\%\uparrow$)   & 0.029 (29.3 $\%\uparrow$) \\
\end{tblr}
\label{tasks}
\end{table*}


Figure \ref{1_4_7} illustrates trajectories from steps 1, 4, and 7 of trajectory prediction, together with the actual reference trajectory.
For clarity, we present the experimental results in a two-dimensional plane in which Joint 2 is the x-axis and Joint 4 is the y-axis.
Moreover, the right side of each subplot shows an expanded view of the trajectory prediction details within the black box.
It can be seen that the trajectory prediction at step 1 aligns well with the actual trajectory, while that the trajectory predictions at steps 4 and 7 deviate slightly from the real trajectory.
These results demonstrate that as the number of prediction steps increase, the prediction accuracy slightly decreases and corresponding prediction variance increases.
This trend is explicitly considered in the preemptive tuning algorithm.
Overall, the above-mentioned results confirm that the our intention predictor reliably forecasts upper-limb joint movements and certain regular motions by capturing the dynamic trends of trajectories based on historical observations.
The free-move task was directly related to human movements in practical applications, as it simulated real-life scenarios. In contrast, the sinusoidal, circular, and lemniscate tasks were designed to test the generalization performance of our intention predictor in contexts other than rehabilitation training.

\subsection{Anomaly Detection}
The collected motion data, inclusive of interaction information, were integrated to train the our anomaly detector, which operated in real time to evaluate safety and the naturalness of motion.
Therefore, these data served as a repository of safe and natural interaction feedback, enabling the anomaly detector to discern their latent relationship and subsequently identify abnormal interactions during human–robot interaction.
The parameters were $L_s = 100$ and $N_c = 21$, denoting the history of joint and motor motion data, in addition to interaction torque.
The diffusion model parameters during both training and inference phases were $T^p=100$ and $\nu = 60$. 
DDIMs were also employed during inference.

\begin{figure}[!h] 
	\centering 
	\vspace{-0.3cm} 
	\subfigtopskip=2pt 
	\subfigbottomskip=2pt 
	\subfigcapskip=-5pt 
	\subfigure[]{
		\includegraphics[width=0.31\linewidth]{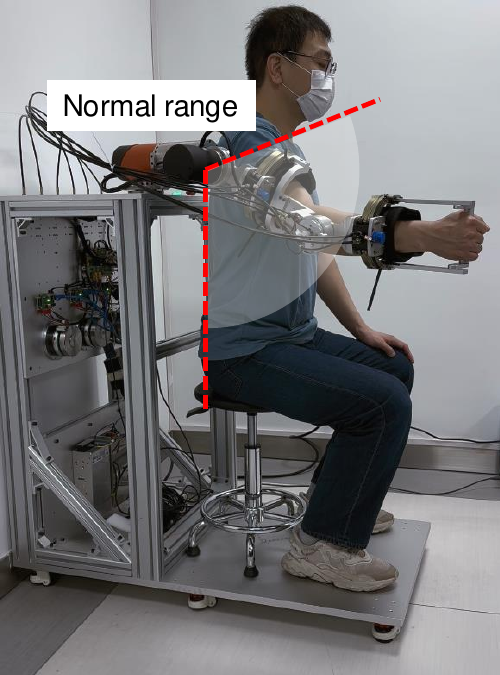}}
	\subfigure[]{
		\includegraphics[width=0.31\linewidth]
        {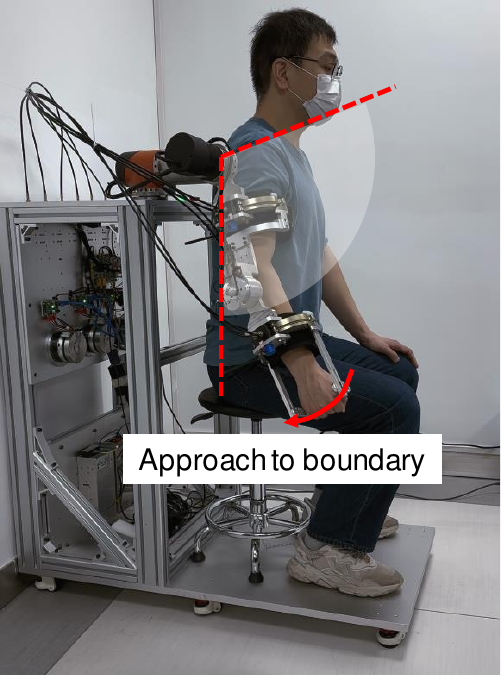}}
	\subfigure[]{
		\includegraphics[width=0.31\linewidth]{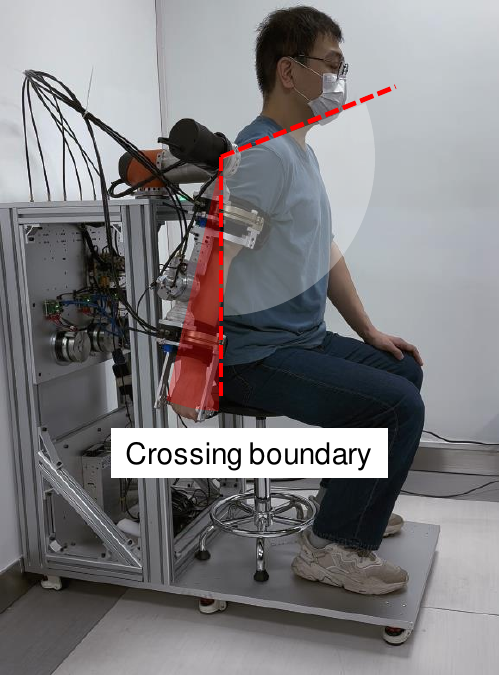}}
  \subfigure[]{
		\includegraphics[width=1\linewidth]
  {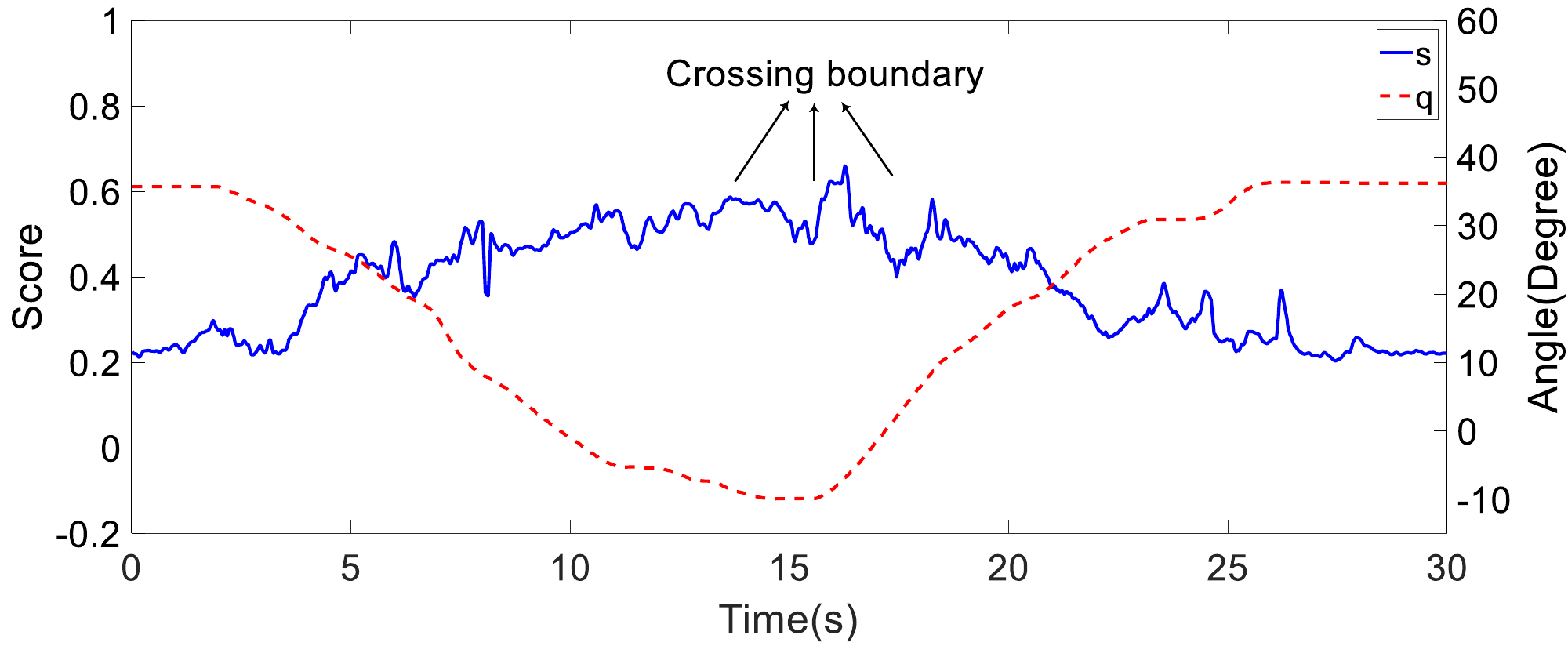}}
	\caption{(a)–(c) Snapshots of anomaly due to excessive movements; (d) anomaly score and joint position during excessive movements. The white sectors represent areas where activity is normal, while the red sectors represent areas where activity is anomalous.}
	\label{move_ano}
\end{figure}

During the experiment, a participant was required to move while wearing the upper-limb exoskeleton robot to simulate various anomalies.
Three abnormal scenarios were considered: an excessive movement scenario (involving joint movements beyond the normal range), a balance deviation scenario (involving deviations from the relative balance position), and a simulated stroke tremors scenario.
In the excessive movement scenario, the participant initially maintained the upper-limb exoskeleton robot within the normal movement range and then lowered the arm to simulate the anomaly.
As the arm was gradually lowered, there was a decrease in the shoulder joint angle that progressively crossed the motion boundary and thus the anomaly level increased gradually. 
Subsequently, the arm was raised back to the normal motion range. 
The results are presented in Figure \ref{move_ano} and reveal that as the joint positions gradually approached and then crossed the boundary, the anomaly score increased.
These results demonstrate that the our anomaly detector effectively detected deviations from the normal motion range (i.e., the motion range of the collected data).

\begin{figure}[!h] 
	\centering 
	\vspace{-0.3cm} 
	\subfigtopskip=2pt 
	\subfigbottomskip=2pt 
	\subfigcapskip=-5pt 
	\subfigure[]{
		\includegraphics[width=0.31\linewidth]{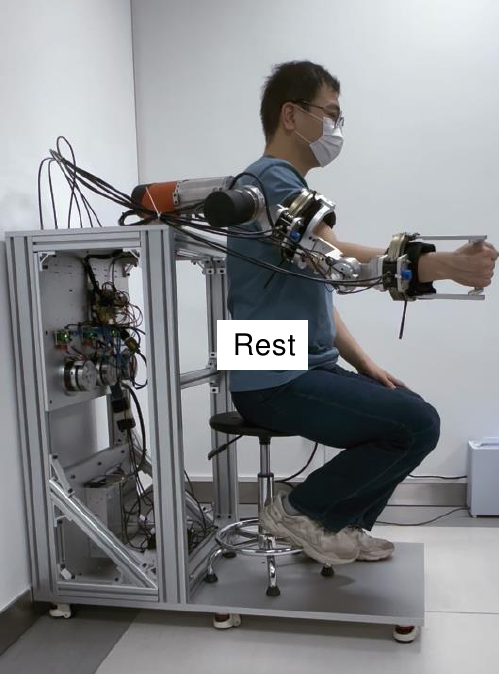}}
	\subfigure[]{
		\includegraphics[width=0.31\linewidth]
        {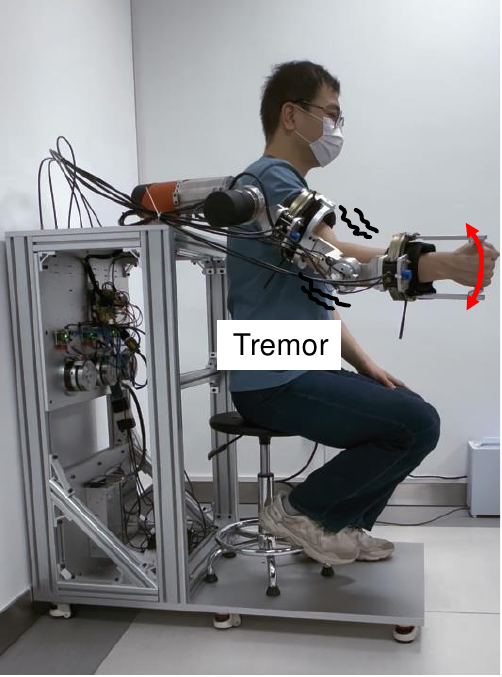}}
	\subfigure[]{
		\includegraphics[width=0.31\linewidth]{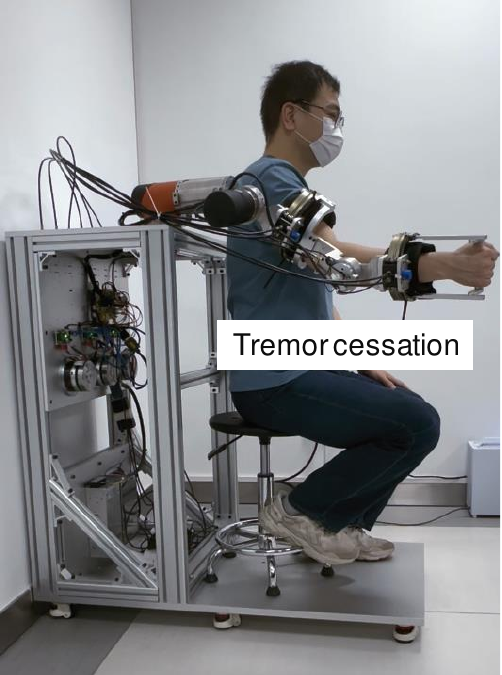}}
  \subfigure[]{
		\includegraphics[width=1\linewidth]
  {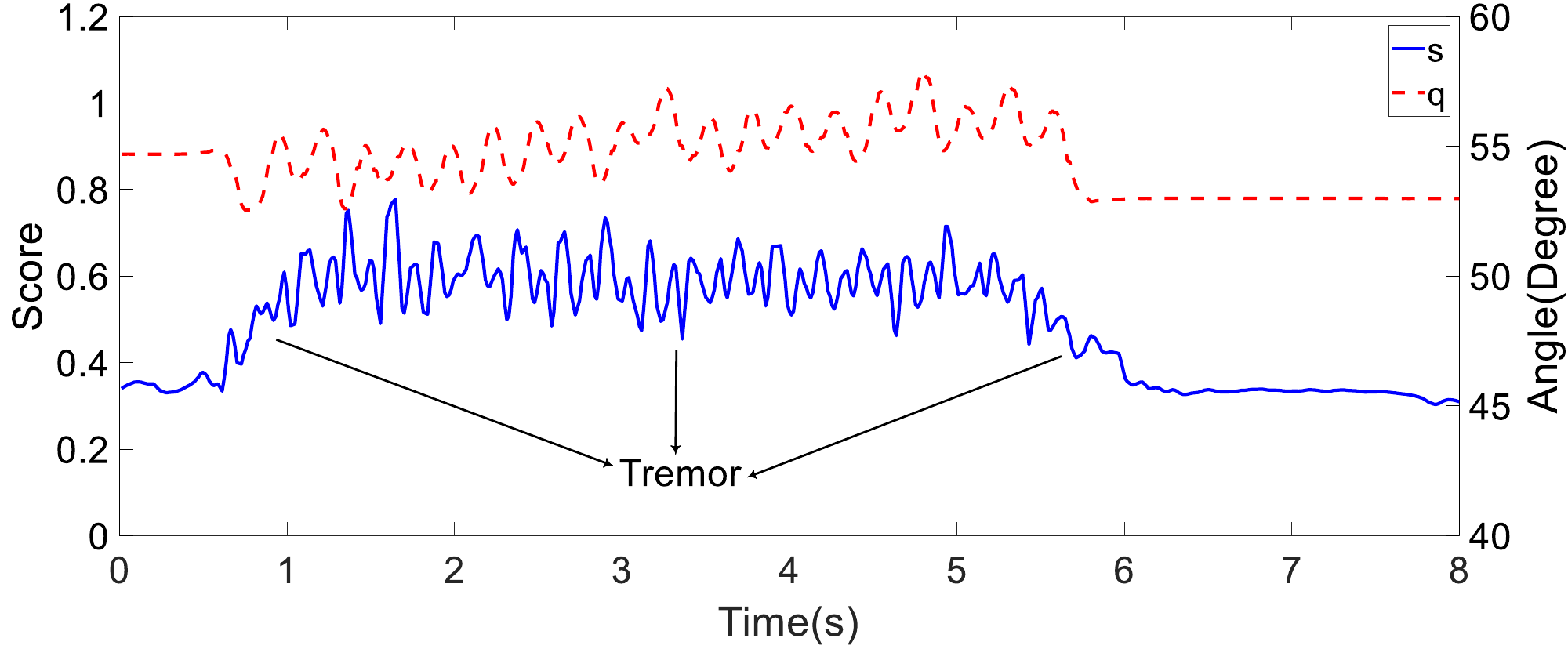}}
	\caption{(a)–(c) Snapshots of conflict due to simulated stroke
tremors; (d) anomaly score and joint position during simulated stroke tremors}
	\label{tremor_ano}
\end{figure}

In the simulated stroke tremors scenario, the participant initially maintained the upper-limb exoskeleton robot in the rest phase.
Subsequently, the participant shook the entire arm of the upper-limb exoskeleton robot to simulate tremors.
The corresponding anomaly score and joint positions are depicted in Figure \ref{tremor_ano}.
It is evident that during the simulated stroke tremors, the anomaly score generally increased, and while the tremors persisted, it remained rather high.

\begin{figure}[!h] 
	\centering 
	\vspace{-0.3cm} 
	\subfigtopskip=2pt 
	\subfigbottomskip=2pt 
	\subfigcapskip=-5pt 
	\subfigure[]{
		\includegraphics[width=0.31\linewidth]{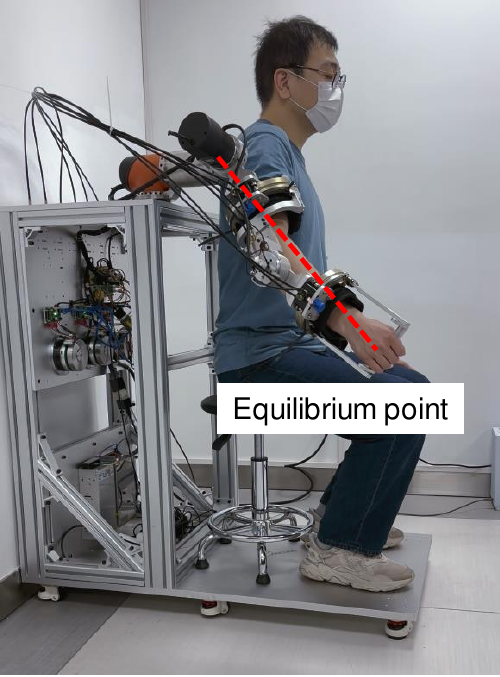}}
	\subfigure[]{
		\includegraphics[width=0.31\linewidth]
        {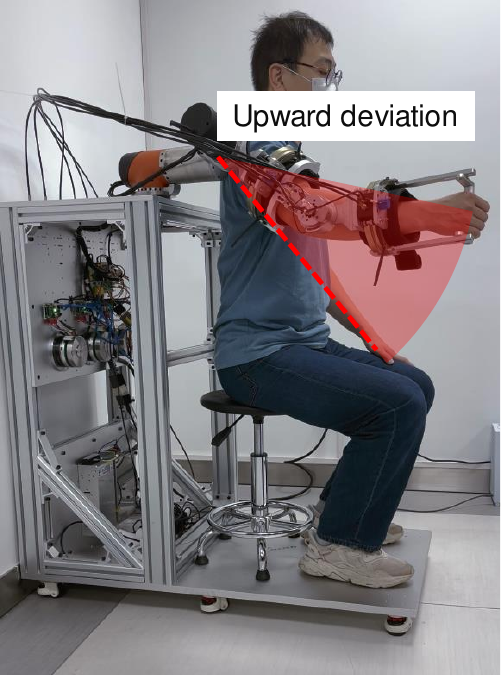}}
	\subfigure[]{
		\includegraphics[width=0.31\linewidth]{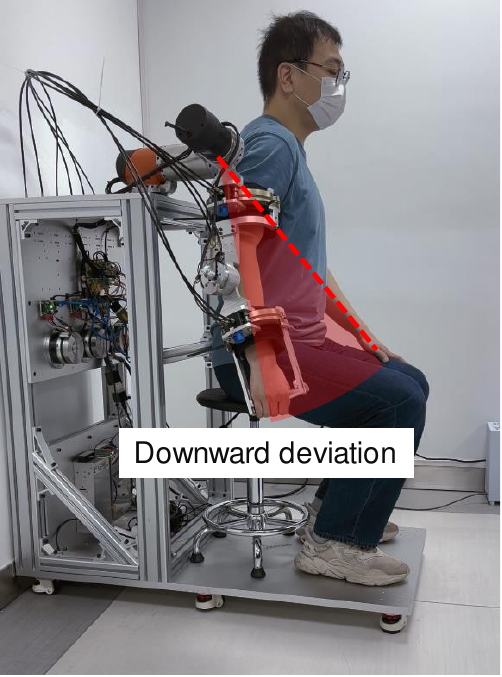}}
  \subfigure[]{
		\includegraphics[width=1\linewidth]
  {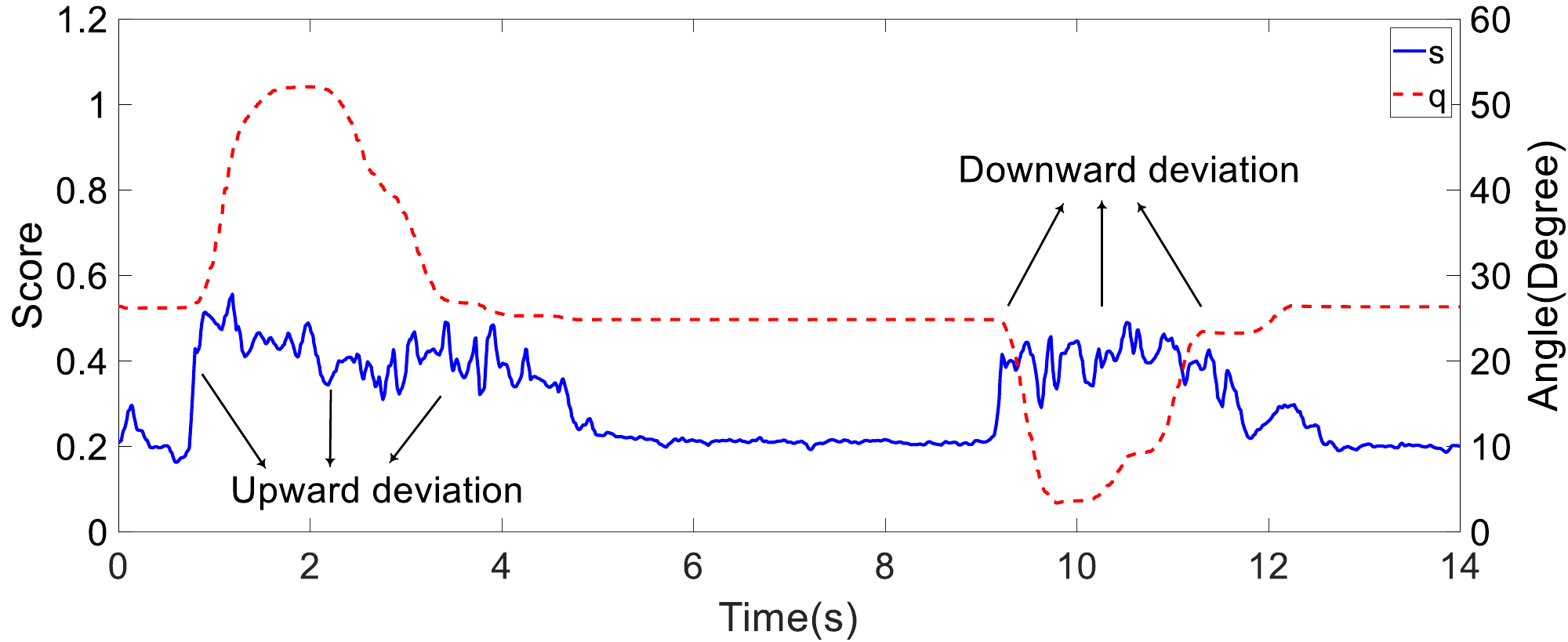}}
	\caption{(a)–(c) Snapshots of conflict due to balance deviation; (d) anomaly score and joint position during balance deviation. The red sectors represent areas where anomalies occurred.}
	\label{equib_ano}
\end{figure}

In the balance deviation scenario, the participant initially maintained the upper-limb exoskeleton robot in a static equilibrium position that represented the balance state in which the upper-limb exoskeleton robot offered assistance. 
The participant was then instructed to manipulate the upper-limb exoskeleton robot upward and downward to simulate misalignment during assistance.
The results are presented in Figure \ref{equib_ano} and reveal that there was a marked increase in anomaly scores whenever deviations occurred, regardless of the direction (i.e., regardless of whether the patient’s movement trajectory was above or below the predetermined trajectory).
This increase in scores indicates that the above-mentioned anomalies were detected.

\begin{table}[!h]
\centering
\caption{Comparison of AUCs of the Two Anomaly Detection Methods}
\begin{tabular}{|l|ll|} 
\hline
Method & VAE   & Ours   \\ 
\hline
AUC    & 0.865 & 0.999  \\
\hline
\end{tabular}
\label{AUC}
\end{table}

\begin{figure}[!h] 
	\centering 
	\vspace{-0.3cm} 
	\subfigtopskip=2pt 
	\subfigbottomskip=2pt 
	\subfigcapskip=-5pt 
		\includegraphics[width=1\linewidth]{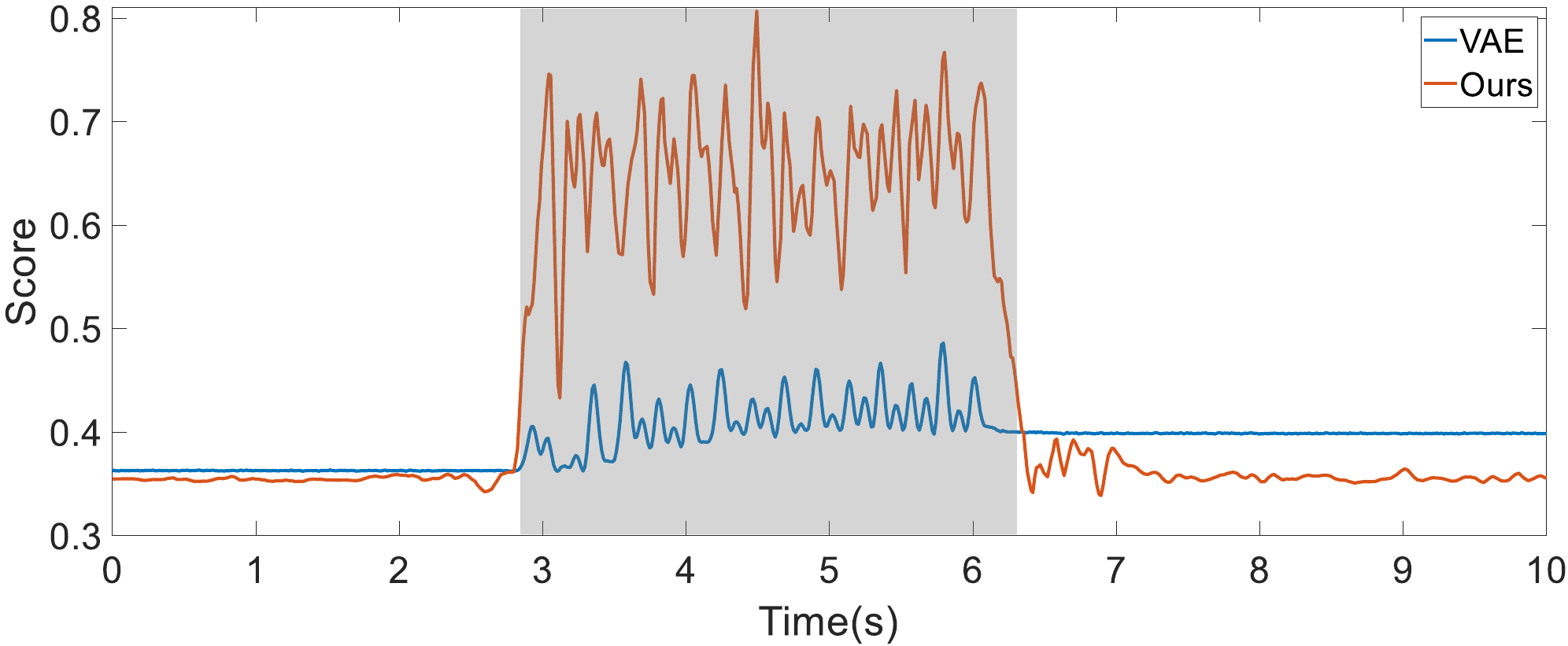}
	\caption{Anomaly score curves for the two different anomaly detection methods, with shaded areas indicating the areas where anomalies occurred.}
	\label{ROC}
\end{figure}

The performance of the our anomaly detector was experimentally assessed using a VAE-based anomaly detector \citep{zhang2023multi} as a baseline.
Specifically, detection performance in the stroke tremor scenario was evaluated using a receiver operating characteristic (ROC) curve, which illustrates the ability of a classification model to differentiate between classes.
The areas under the ROC curve for the considered models are presented in Table \ref{AUC}, and the calculated anomaly scores are depicted in Figure \ref{ROC}.
In the figure, it can be seen that the scores generated by the proposed anomaly detector markedly increased as the anomaly occurred. Furthermore, the scores generated by the VAE-based anomaly detector  failed to return to a normal level once the anomaly ceased.
These results indicate that compared with the VAE-based anomaly detector, our anomaly detector is more adaptive in classifying anomalies in different joint configurations, owing to the superior generative performance of the diffusion model.

\begin{figure*}[!h] 
	\centering 
	\vspace{-0.3cm} 
	\subfigtopskip=2pt 
	\subfigbottomskip=2pt 
	\subfigcapskip=-5pt 
	\subfigure[]{
		\includegraphics[width=0.3\linewidth]{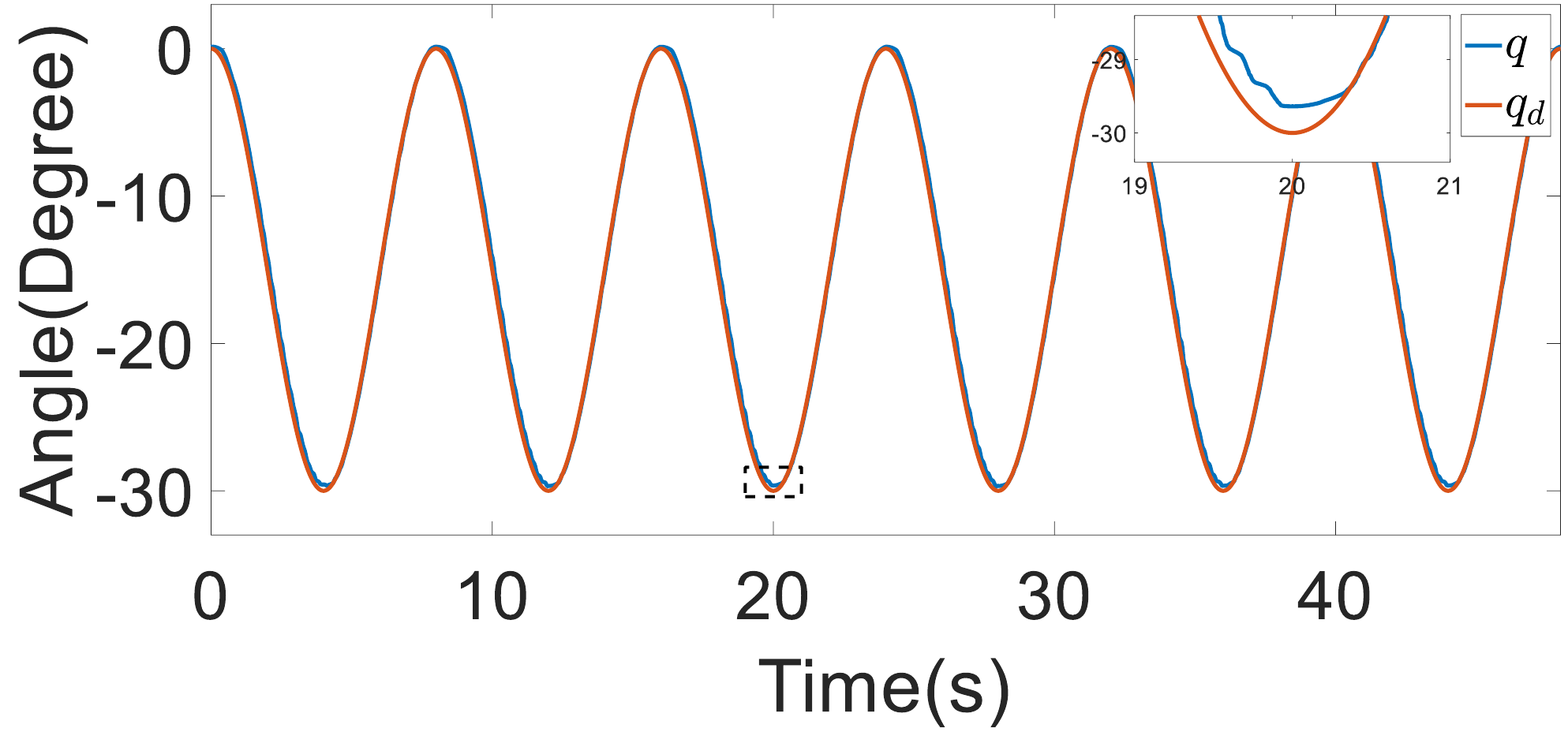}}
	\subfigure[]{
		\includegraphics[width=0.3\linewidth]
        {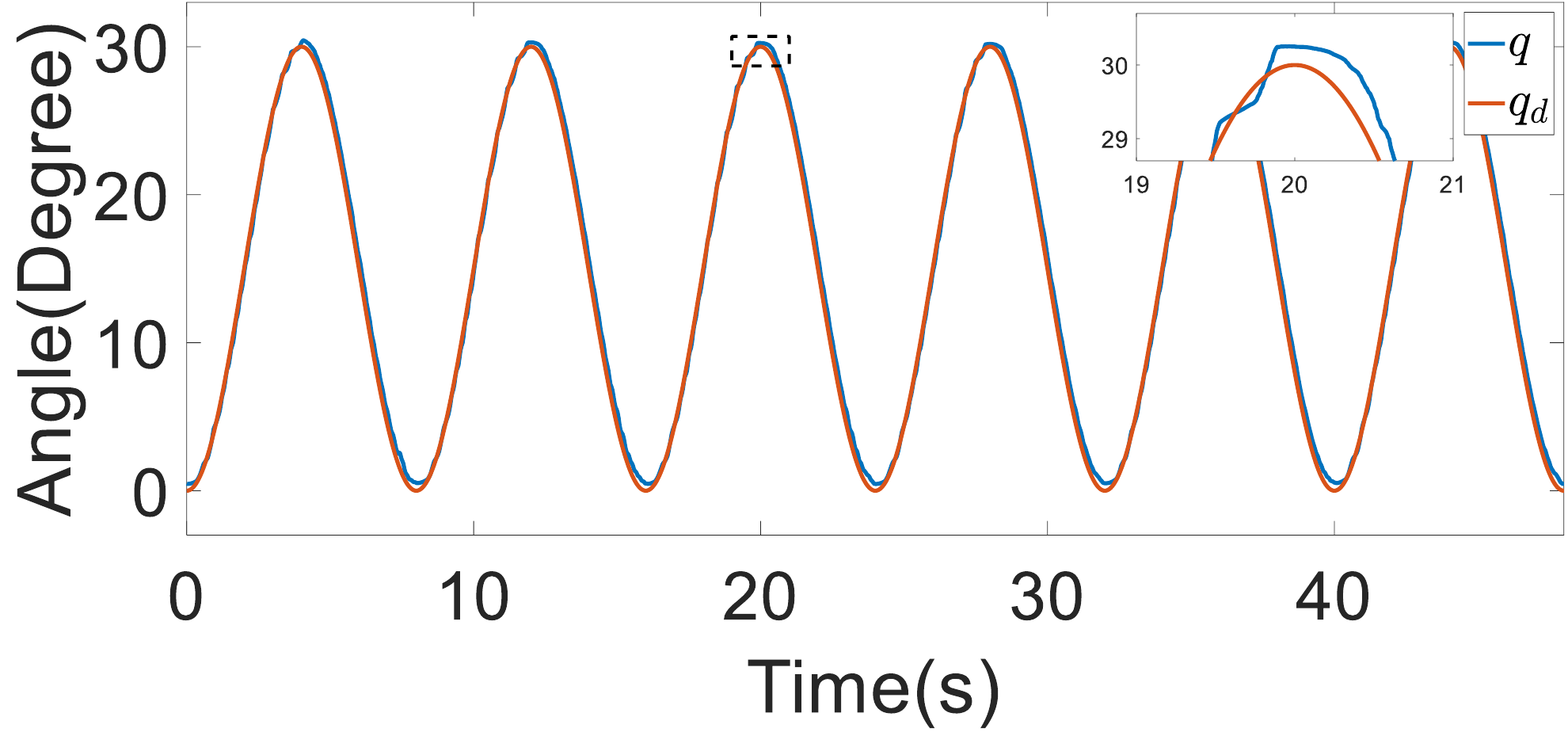}}
	\subfigure[]{
		\includegraphics[width=0.3\linewidth]{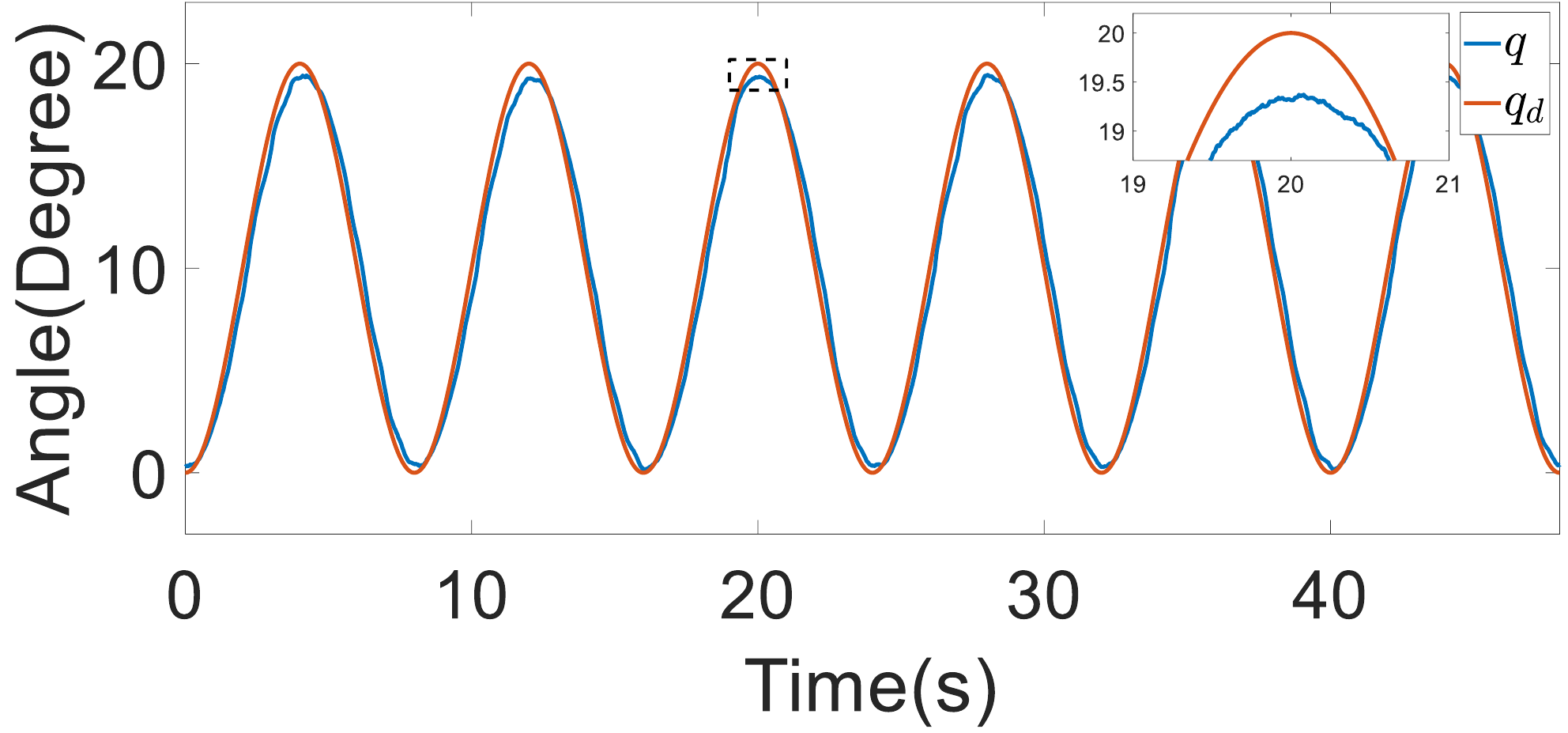}}
    \subfigure[]{
		\includegraphics[width=0.3\linewidth]{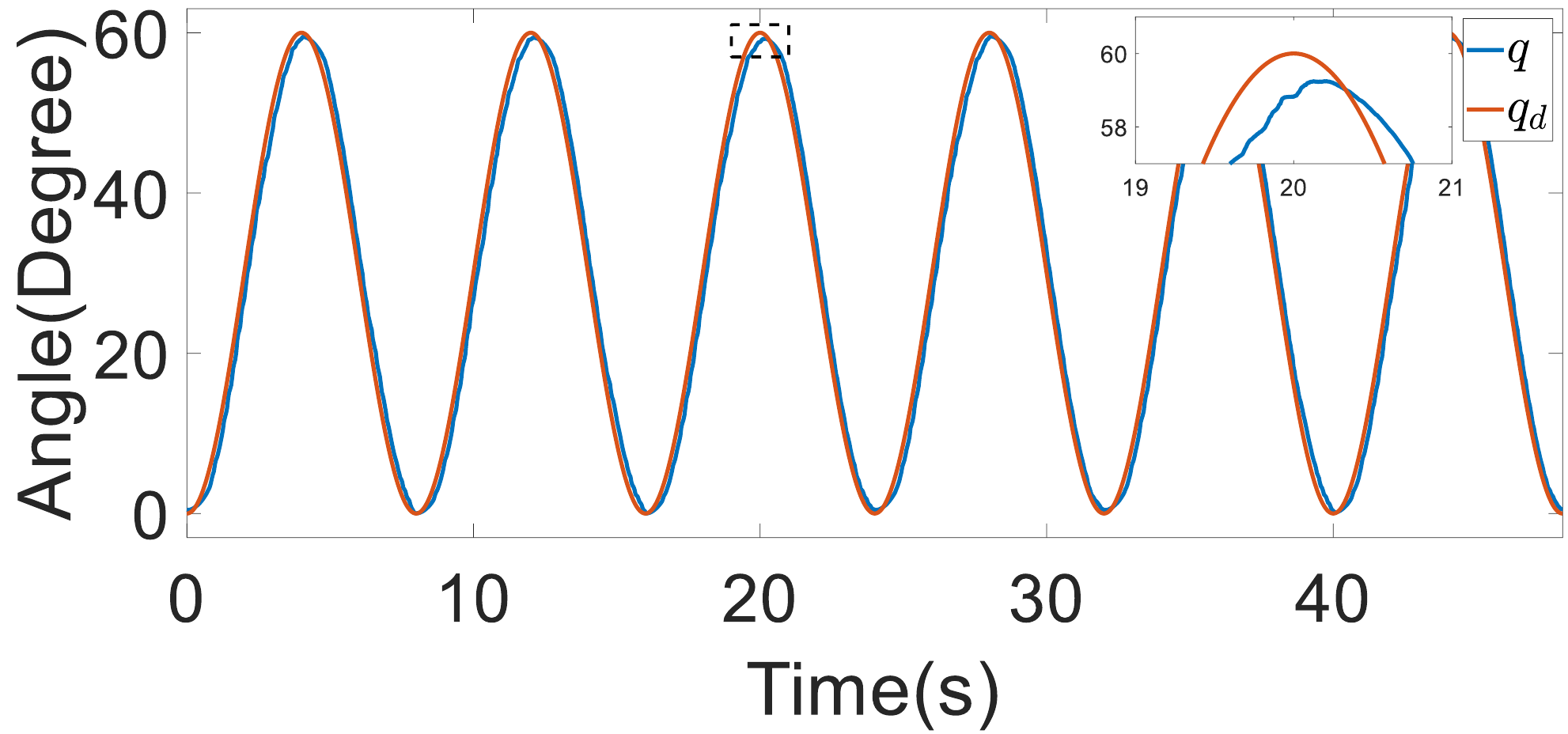}}
    \subfigure[]{
		\includegraphics[width=0.3\linewidth]{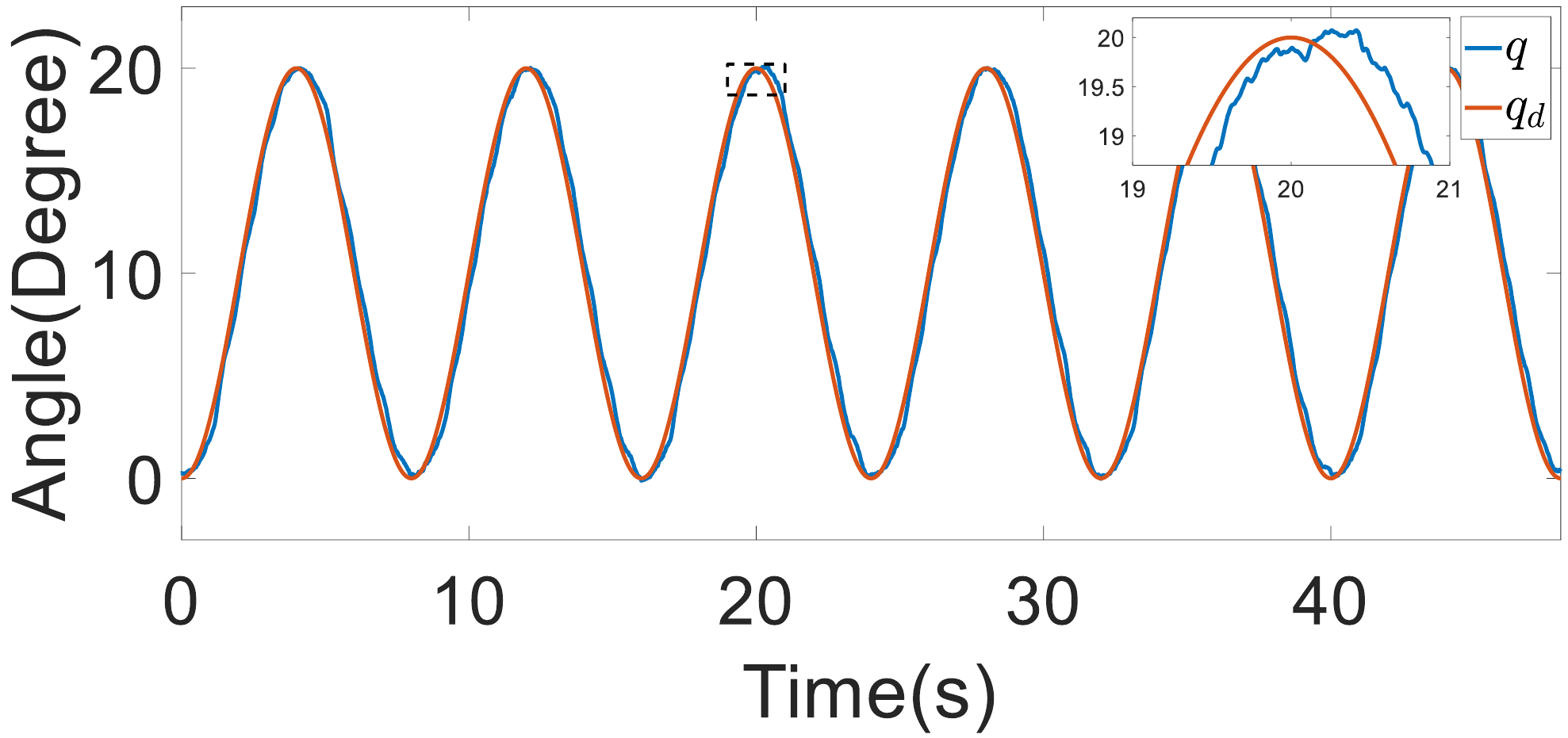}}
	\caption{(a)-(e) Trajectory tracking performance across Joints 1 through 5 of the exoskeleton.}
	\label{VIC_track}
\end{figure*}

\subsection{Interaction Control}
During rehabilitation training, the upper-limb exoskeleton robot is required to regulate human-robot interaction in a desired impedance model with consideration of friction compensation.
To this end, the upper-limb exoskeleton robot was operated to move slowly in the absence of a participant and friction was quantified.
That is, the difference between the readings from the potentiometers at the motor output and the measurements from the torque sensor at the joint end were calculated. In addition, joint velocities were recorded. 
Subsequently, polynomial fitting was applied based on the simplified friction model (\ref{friction_model}) to afford the estimated values for friction parameters presented in Table \ref{friction_param}.

\begin{table}[!h]
\centering
\renewcommand{\arraystretch}{1.2} 
\caption{Estimated Friction Model}
\label{friction_param}
    \begin{tabular}{|c|c|c|c|}
        \hline
         & $\bar a_f$ & $\bar{b}_f$ & $\bar c_f$  \\ 
         \hline
        Joint 3 & 0.822 & -2.132 & 0.557\\ \hline
        Joint 4 & 1.718 & 11.441 & -3.028\\ \hline
        Joint 5 & 0.636 & -1.212 & 1.005\\ \hline
    \end{tabular}
\end{table}

Next, we implemented our impedance controller in the upper-limb exoskeleton robot, devoid of a patient, to track a pre-defined sinusoidal trajectory involving all active joints.
This experiment aimed to validate the accuracy of the friction compensation and evaluate the dynamic performance of the controller.
The impedance parameters were set as follows: $\bm C_d=10\bm I_5, \bm K_d=50\bm I_5$ where $\bm I_5$ is a $5\times 5$ identity matrix. The parameters of the weighting function were set as follows: $\lambda_1=0.5, \chi_1=0.04, \chi_2=8.75$ and $\lambda_2=1.5$. The control parameters were set as follows: $\bm K_v=1.1\bm I_3$ and $\bm K_z=diag(1.5,0.6,0.7,4,1.8)$. 
The experimental results demonstrate that the impedance controller effectively compensated for friction, enabling the upper-limb exoskeleton robot to accurately follow the desired trajectory, as illustrated in Figure \ref{VIC_track}.
Specifically, the RMSEs for the joints during this trajectory tracking task were $0.613^{\circ}$ for Joint 1, $0.728^{\circ}$ for Joint 2, $0.997^{\circ}$ for Joint 3, $2.143^{\circ}$ for Joint 4, and $0.948^{\circ}$ for Joint 5.






\begin{figure}[!ht] 
	\centering 
	\vspace{-0.3cm} 
	\subfigtopskip=2pt 
	\subfigbottomskip=2pt 
	\subfigcapskip=-5pt 
	\subfigure[]{
		\includegraphics[width=1\linewidth]{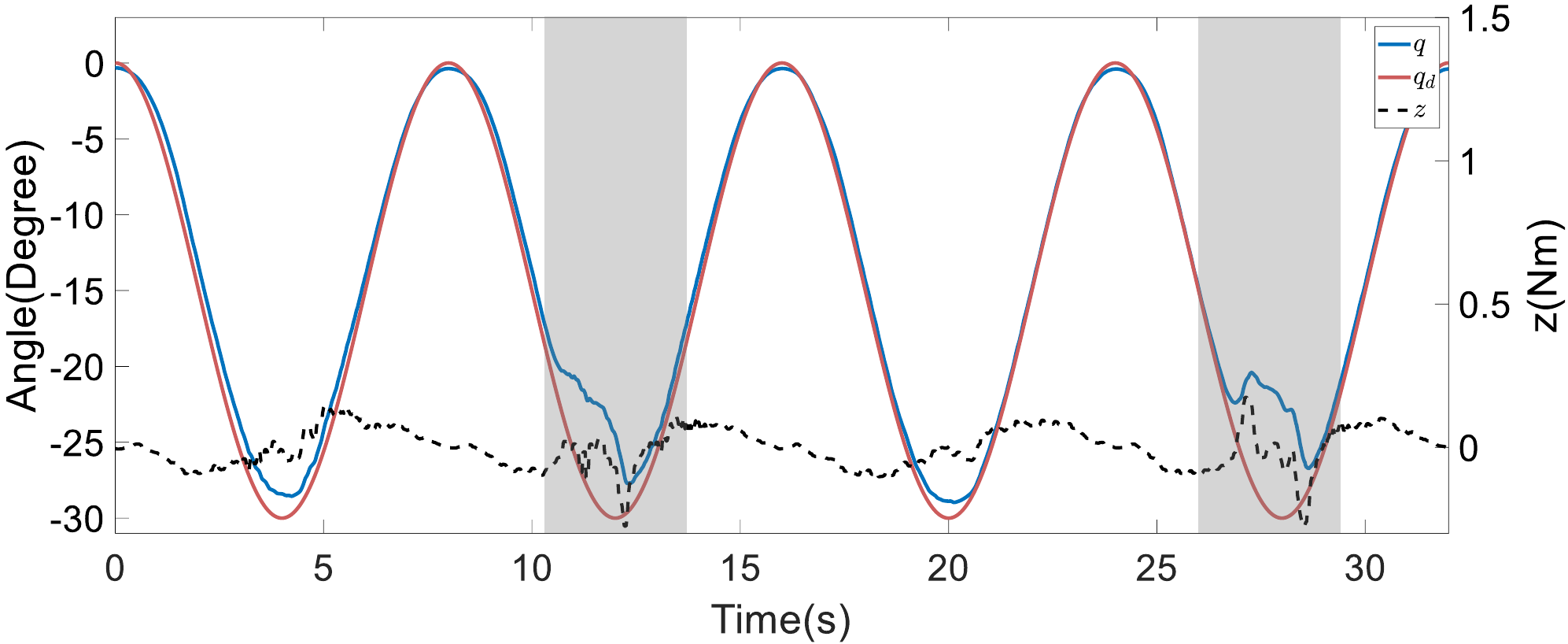}}
	\subfigure[]{
		\includegraphics[width=1\linewidth]{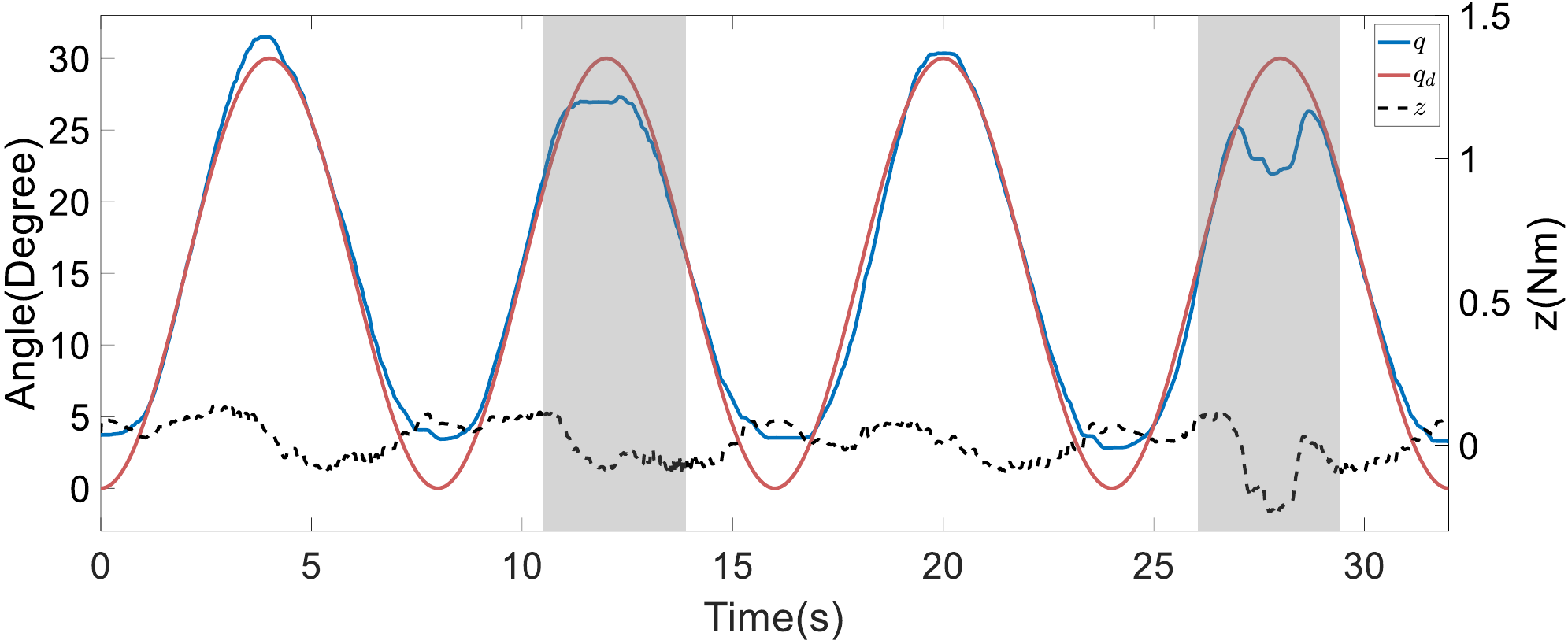}}
	\subfigure[]{
		\includegraphics[width=1\linewidth]{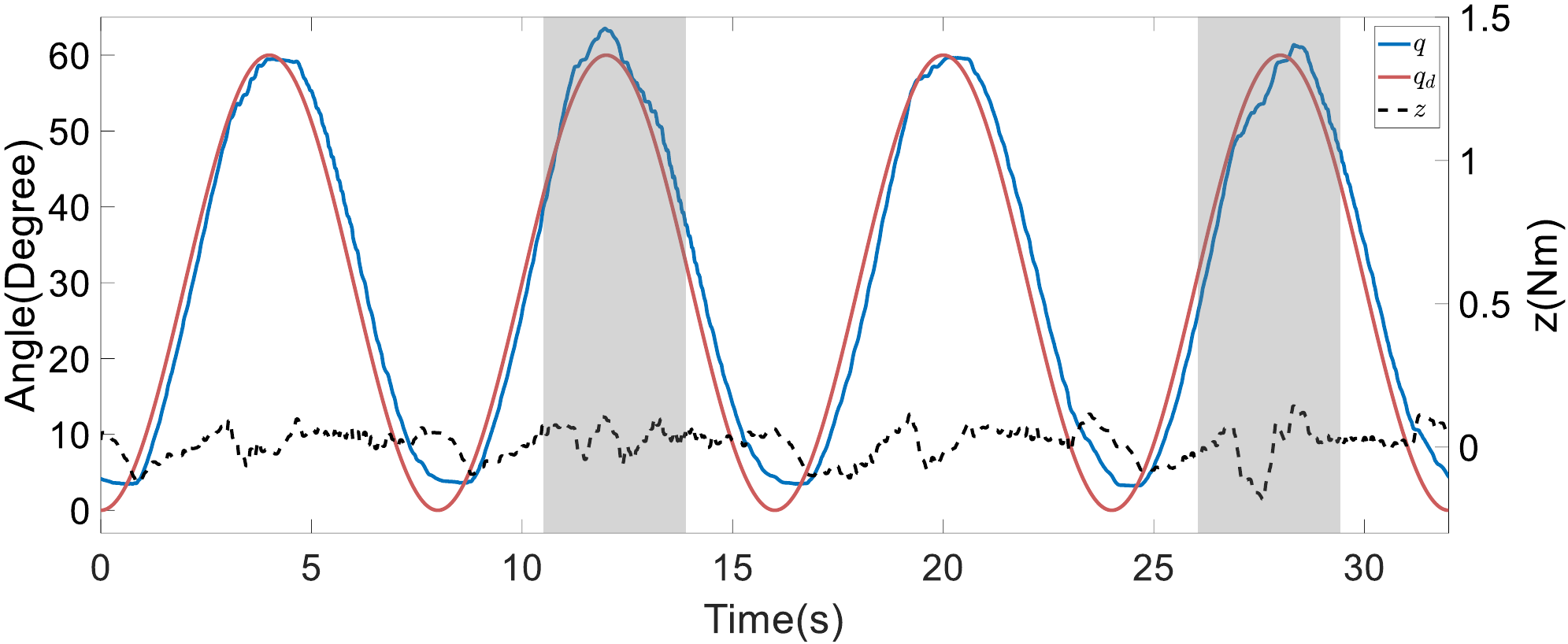}}
	\caption{(a)–(c) Performance of variable impedance controller for Joints 1, 2, and 4 of the upper-limb exoskeleton robot in the presence of a patient. The shaded areas indicate the periods when anomalies occurred.}
	\label{VIC_var}
\end{figure}

Additionally, we assessed the performance of our impedance controller under conditions involving anomalies. 
This was achieved by having the upper-limb exoskeleton robot guide the patient in trajectory tracking while the patient held the upper-limb exoskeleton robot in positions that simulated anomalies.
The results of this experiment are depicted in Figure \ref{VIC_var}.
It can be seen that throughout the experiment, the impedance vector remained close to zero, indicating that the desired variable impedance model was effectively maintained despite human involvement and the occurrence of anomalies.

\begin{figure}[!ht] 
	\centering 
	\vspace{-0.3cm} 
	\subfigtopskip=2pt 
	\subfigbottomskip=2pt 
	\subfigcapskip=-5pt 
	\subfigure[]{
		\includegraphics[width=1\linewidth]{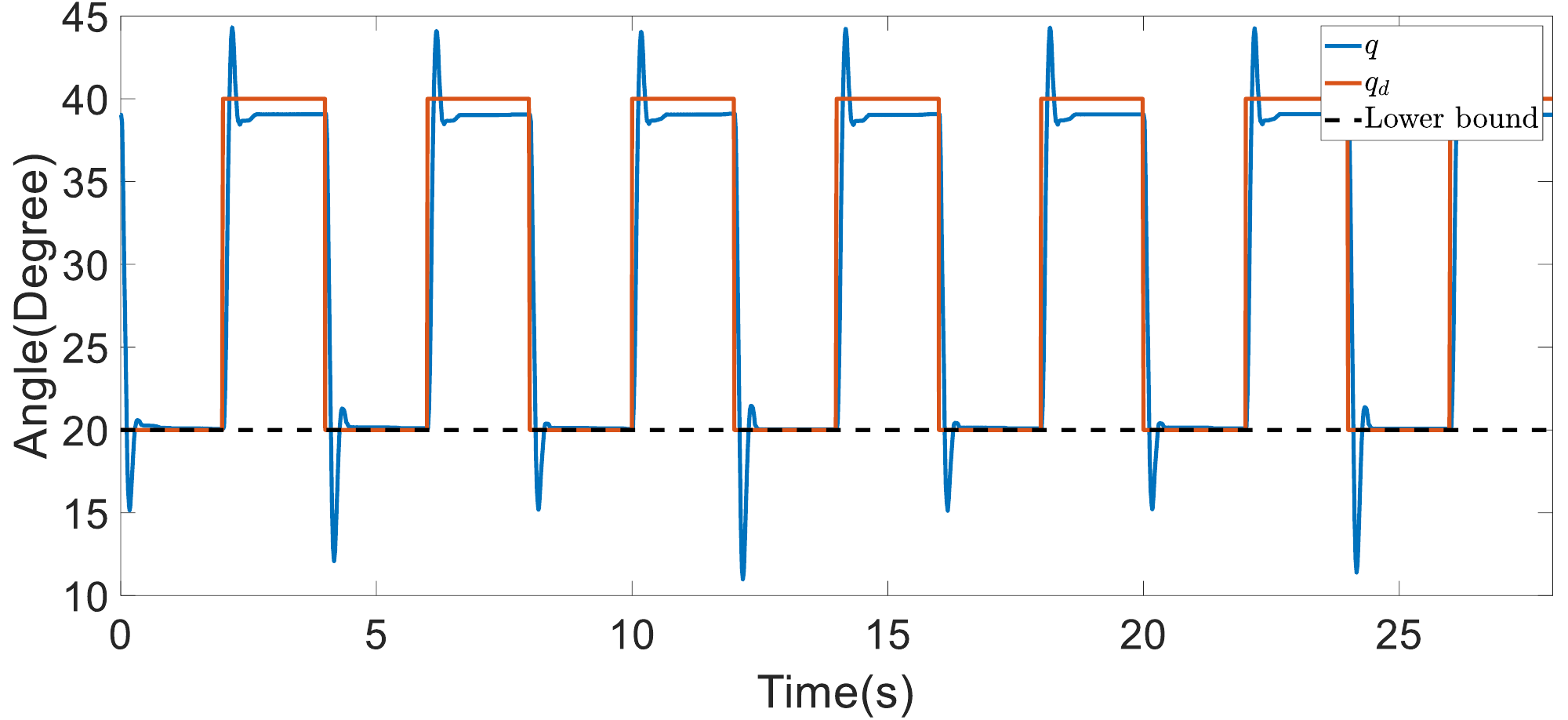}}
	\subfigure[]{
		\includegraphics[width=1\linewidth]{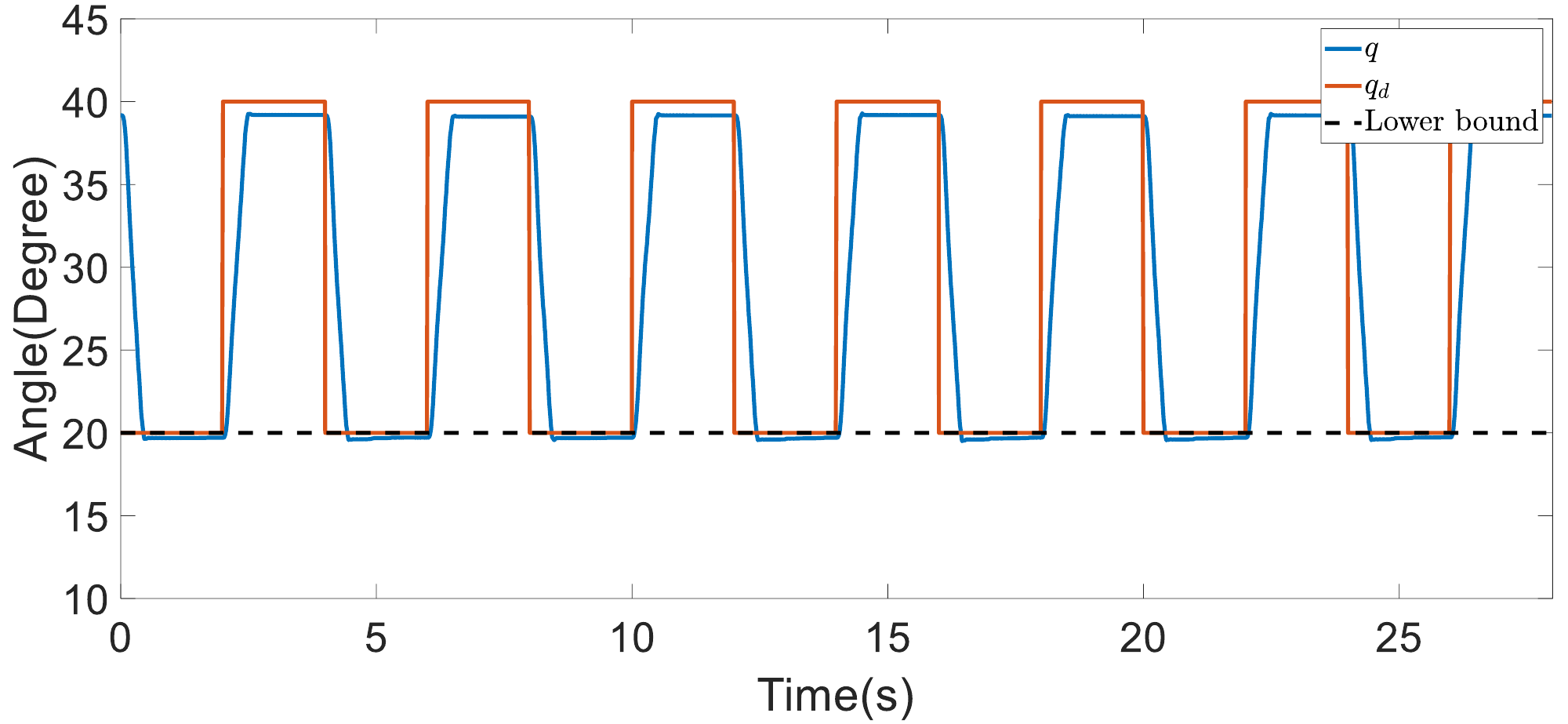}}
	\caption{Test of ability to handle unexpected external impacts on Joint 2 (a) without online trajectory refinement; and (b) with online trajectory refinement}
	\label{active_impact}
\end{figure}

\subsection{Active Mirroring Training}
In active mirroring training, the movements of the upper-limb exoskeleton robot were aligned with the motion intentions of the unaffected side of the body of the patient (as illustrated in Figure \ref{exp_setup}).
In addition, online trajectory refinement was used to smoothen movement commands and enhance safety through dynamic constraints. The details of the experiments are given below.
\begin{enumerate}
\item[-] To evaluate the capability of the proposed method to handle unexpected external impact, we conducted an experiment on Joint 2 of the upper-limb exoskeleton robot’s shoulder.
\item[-] We simulated sudden external disturbances by abruptly altering the desired position during the experiments.
\item[-] To represent the lower constraint of the online trajectory refinement, we imposed a lower bound on the position command at $20^{\circ}$, thereby simulating a tendency to exceed the acceptable movement range.
\item[-] During comparative trials, we eliminated motion command inconsistencies caused by feedback from the unaffected side of the body by deactivating the positional feedback on this side. 
Therefore, we instead relied on the proprioceptive sensors of the upper-limb exoskeleton robot (specifically, its encoders) and conducted the experiment without a patient.
\end{enumerate}

\begin{figure}[!h]
  \centering
    \includegraphics[width=1\linewidth]{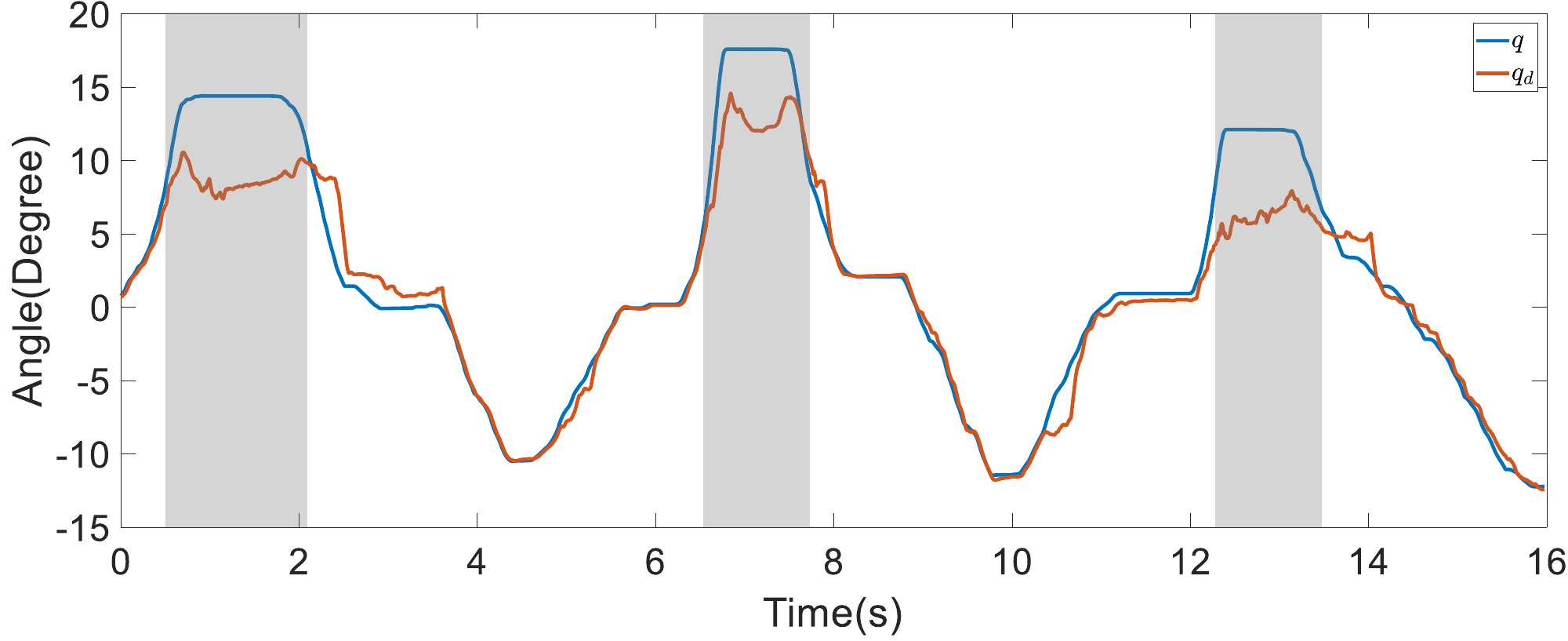}
  \caption{Performance of the system in guiding the upper-limb exoskeleton robot to areas with lower anomaly scores than other areas through online trajectory refinement, with shaded areas indicating anomalies caused by abnormal adduction angles.}
  \label{active_score} 
\end{figure}

The position commands and joint positions with and without the implementation of online trajectory refinement, respectively, are illustrated in Figure \ref{active_impact}.
This figure reveals that without online trajectory refinement, the joint position surpassed the set movement boundary.
However, when online trajectory refinement was activated, the joint movement responded effectively to the dynamic constraints and remained close to the boundary.
Specifically, without trajectory refinement, the movement exceeded the established boundary by approximately $9^{\circ}$. 
In contrast, with trajectory refinement, the excess was significantly reduced to just $0.5^{\circ}$.
The trajectory refinement is designed to adjust the trajectory to adhere to constraints and to mitigate the extent to which joint position violates movement boundaries.
Therefore, even with trajectory refinement, the joint position may slightly cross the boundary, as observed with the $0.5^{\circ}$ transgression in this experiment.

\begin{figure}[!h] 
	\centering 
	\vspace{-0.3cm} 
	\subfigtopskip=2pt 
	\subfigbottomskip=2pt 
	\subfigcapskip=-5pt 
	\subfigure[]{
		\includegraphics[width=1\linewidth]{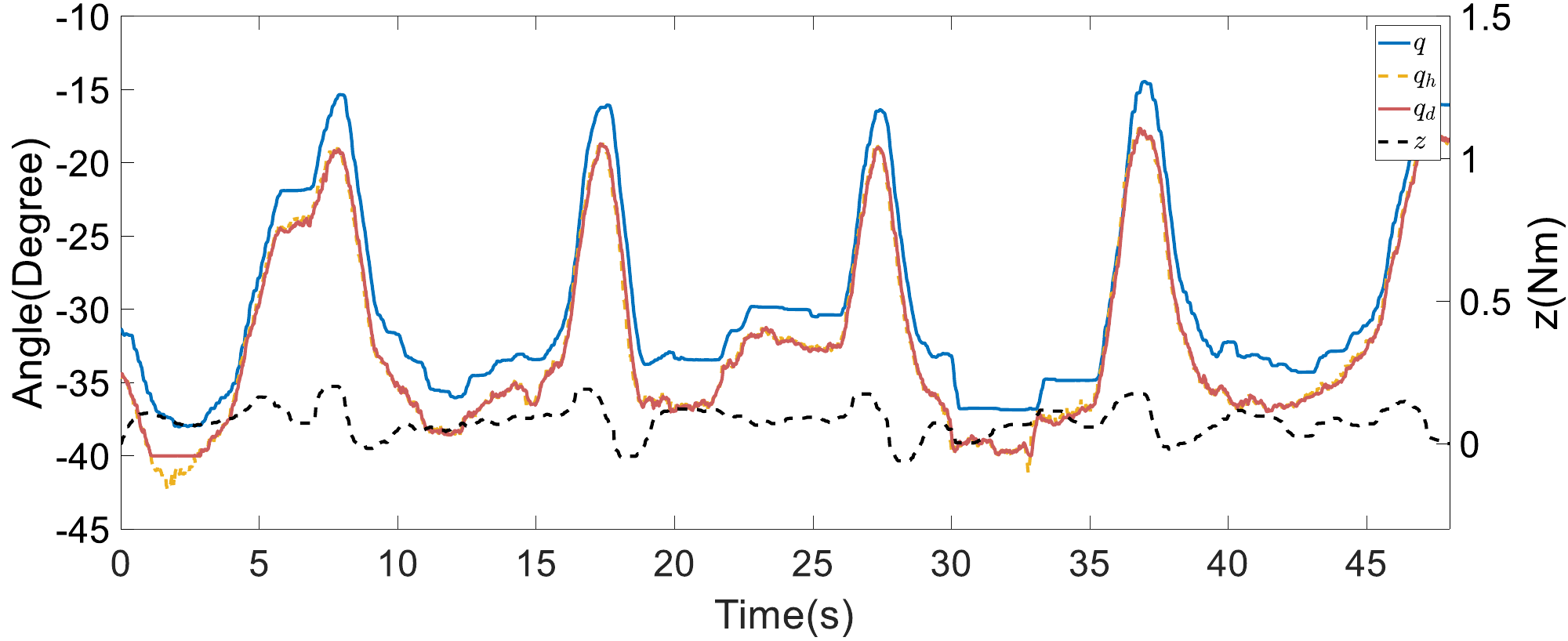}}
	\subfigure[]{
		\includegraphics[width=1\linewidth]{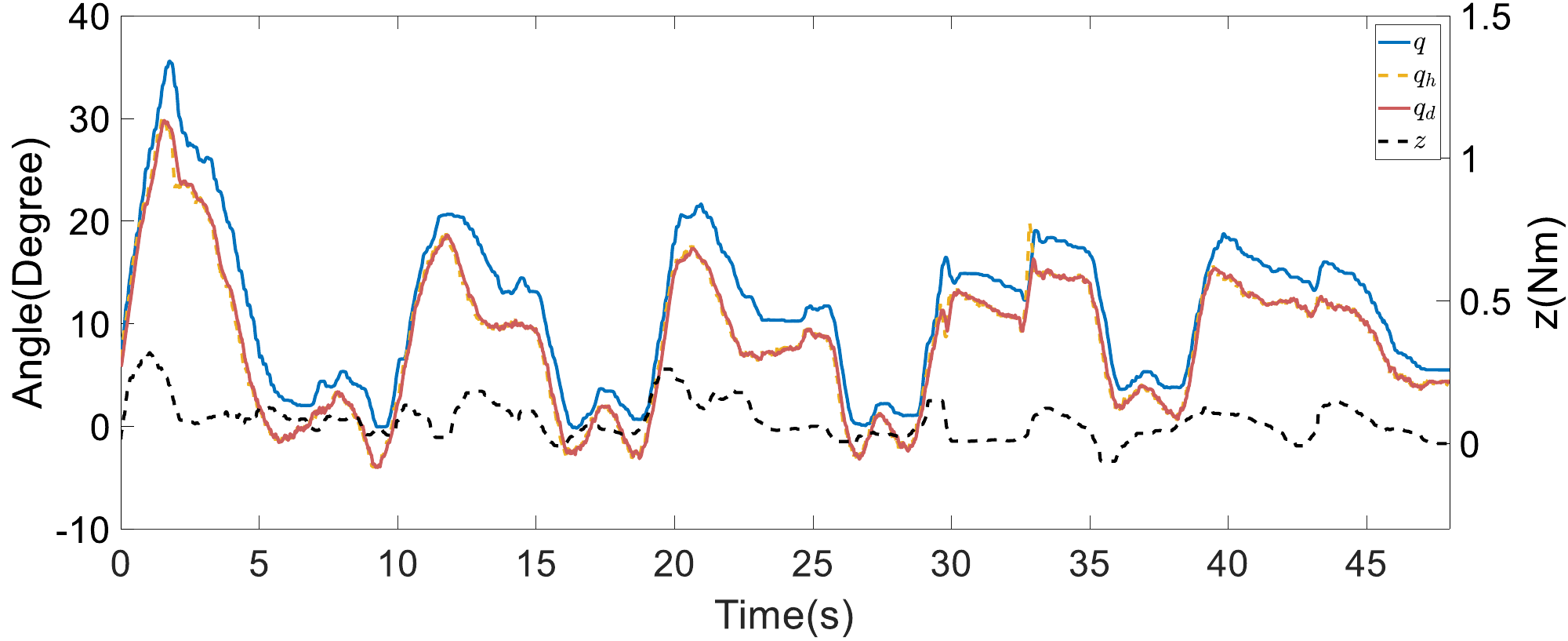}}
	\subfigure[]{
		\includegraphics[width=1\linewidth]{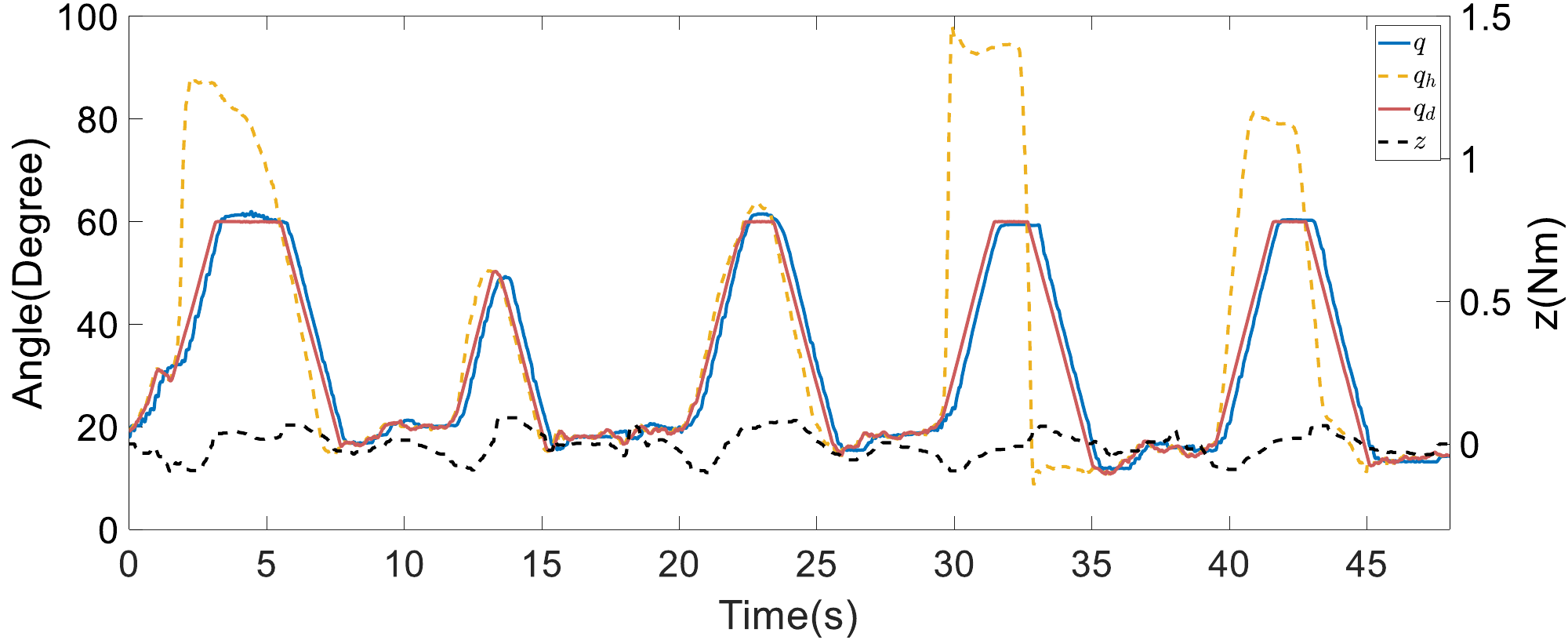}}
	\caption{(a)–(c) Performance of active mirroring training on Joints 1, 2, and 4 of the upper-limb exoskeleton robot}
	\label{active_mirror}
\end{figure}

Next, we experimented with Joint 1 to verify that the proposed method identified anomaly regions and guided the planned trajectory toward safer areas with lower anomaly scores than the current area.
Ideally, this joint’s normal operational range should not significantly exceed zero.
In particular, the shoulder adduction angle should remain small during typical activities.
Thus, as a motion capture system may yield inaccurate adduction angle estimates due to marker obstruction, we instead utilized the encoder feedback from the upper-limb exoskeleton robot to obtain adduction angle estimates.
The results are displayed in Figure \ref{active_score}.
In the figure, it can be seen that as the shoulder adduction angle increased, the motion commands generated by the online trajectory refinement decreased the adduction angle of movements rather than ensuring that they strictly adhered to the target trajectory.
Moreover, when the joint angle returned to the normal range, i.e., when the shoulder moved into the abduction space, the trajectory refinement resumed its focus on aligning the trajectory with the target trajectory within the dynamic constraints. 
These results confirm that the proposed method is capable of tracking changes in human motion intention within the normal activity range and refining the trajectory accordingly.

\begin{figure*}[!h] 
	\centering 
	\vspace{-0.3cm} 
	\subfigtopskip=2pt 
	\subfigbottomskip=2pt 
	\subfigcapskip=-5pt 
	\subfigure[]{
		\includegraphics[width=0.3\linewidth]{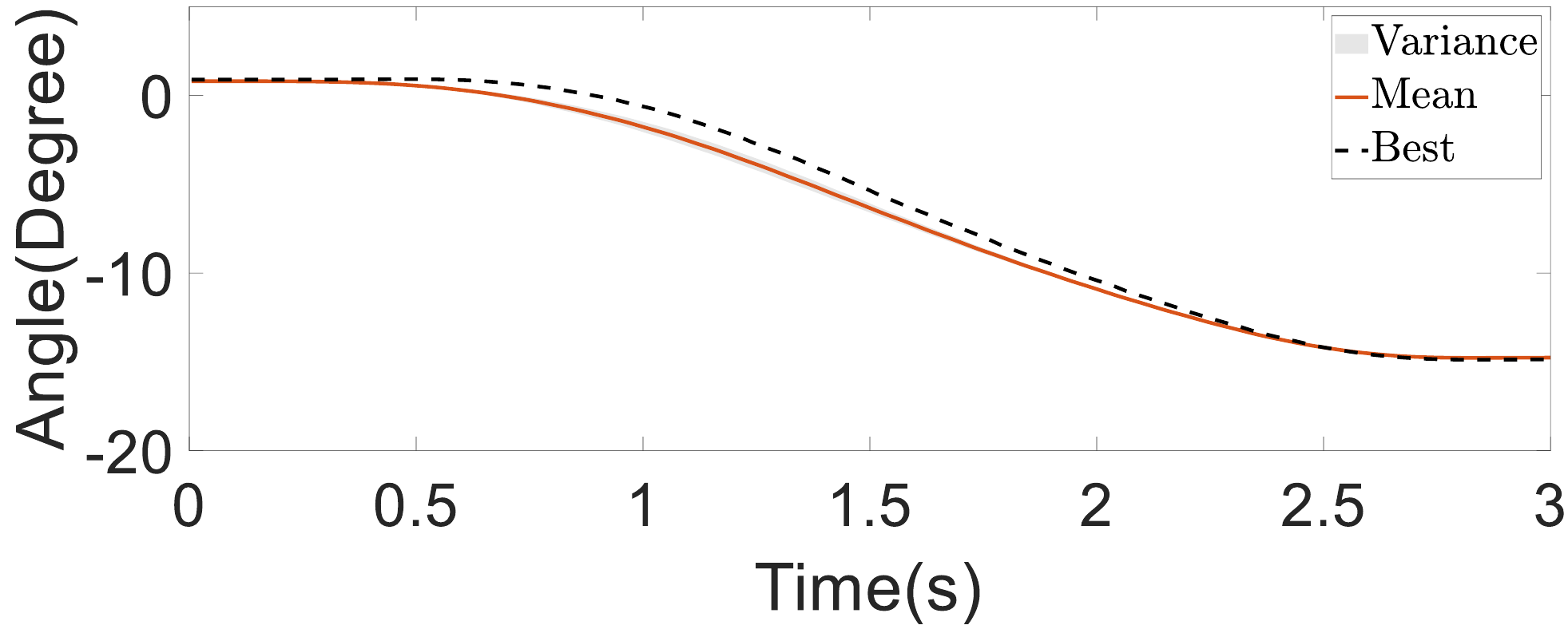}}
	\subfigure[]{
		\includegraphics[width=0.3\linewidth]{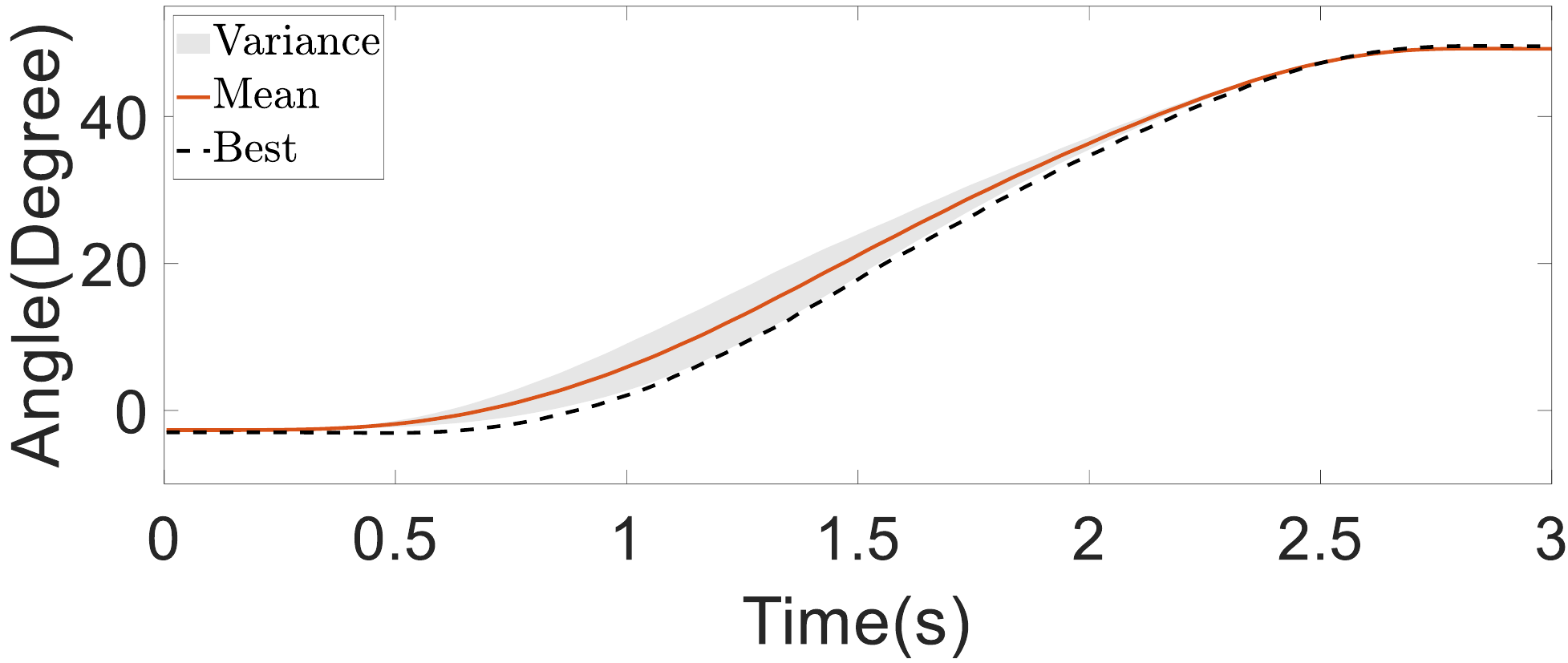}}
	\subfigure[]{
		\includegraphics[width=0.3\linewidth]{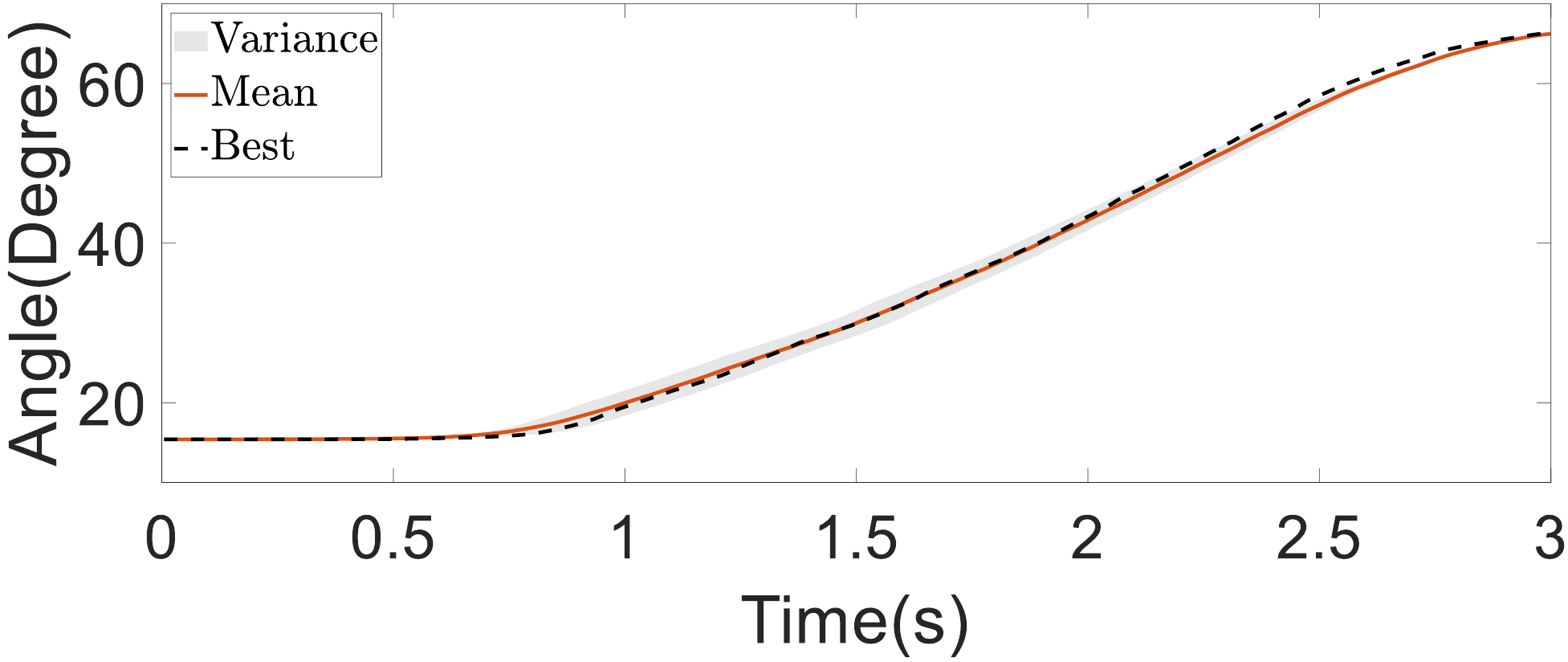}}
    \subfigure[]{
		\includegraphics[width=0.3\linewidth]{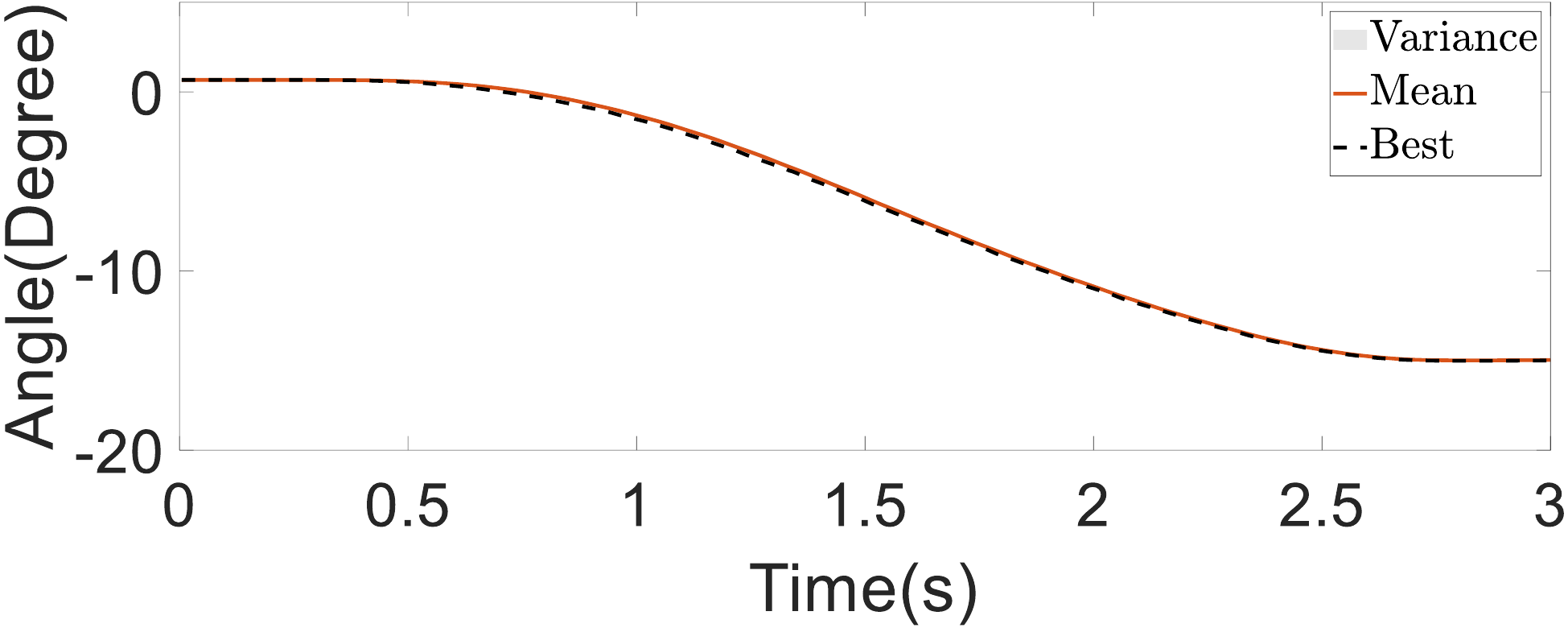}}
    \subfigure[]{
		\includegraphics[width=0.3\linewidth]{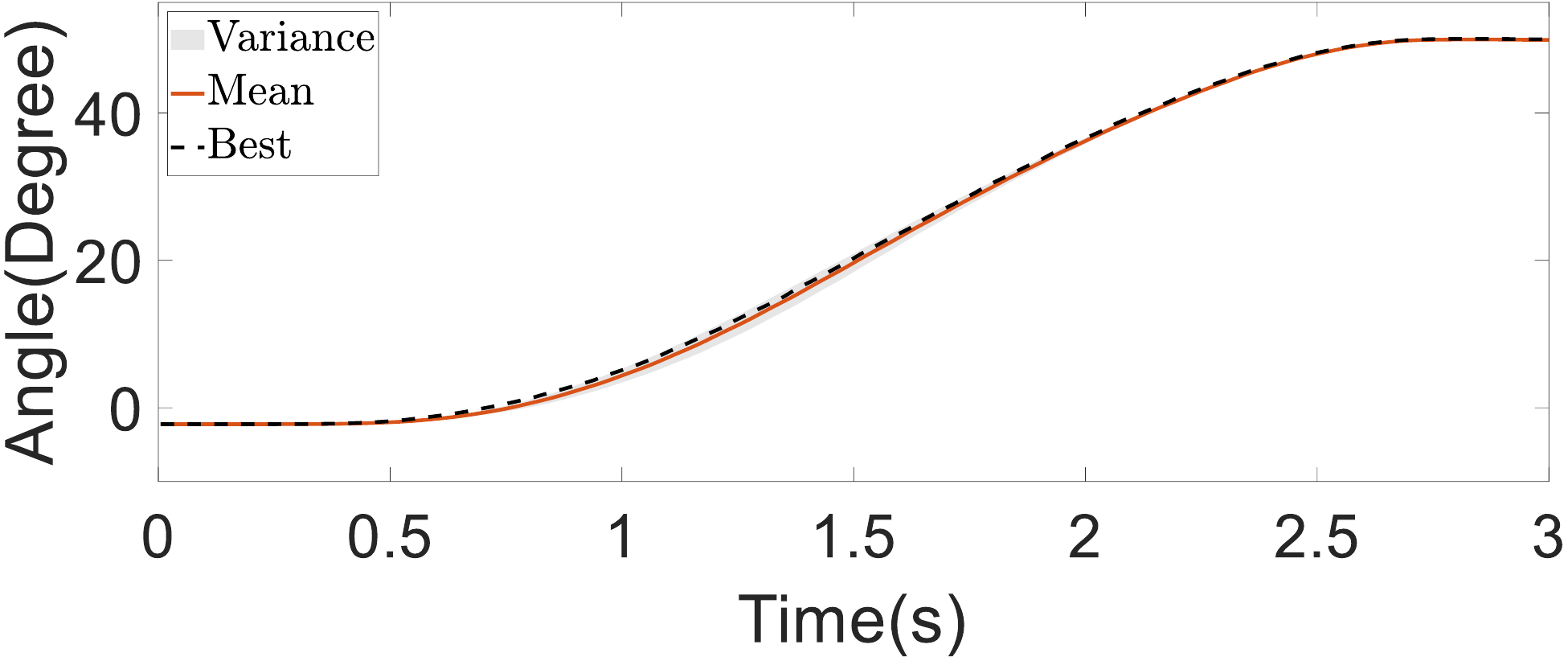}}
    \subfigure[]{
		\includegraphics[width=0.3\linewidth]{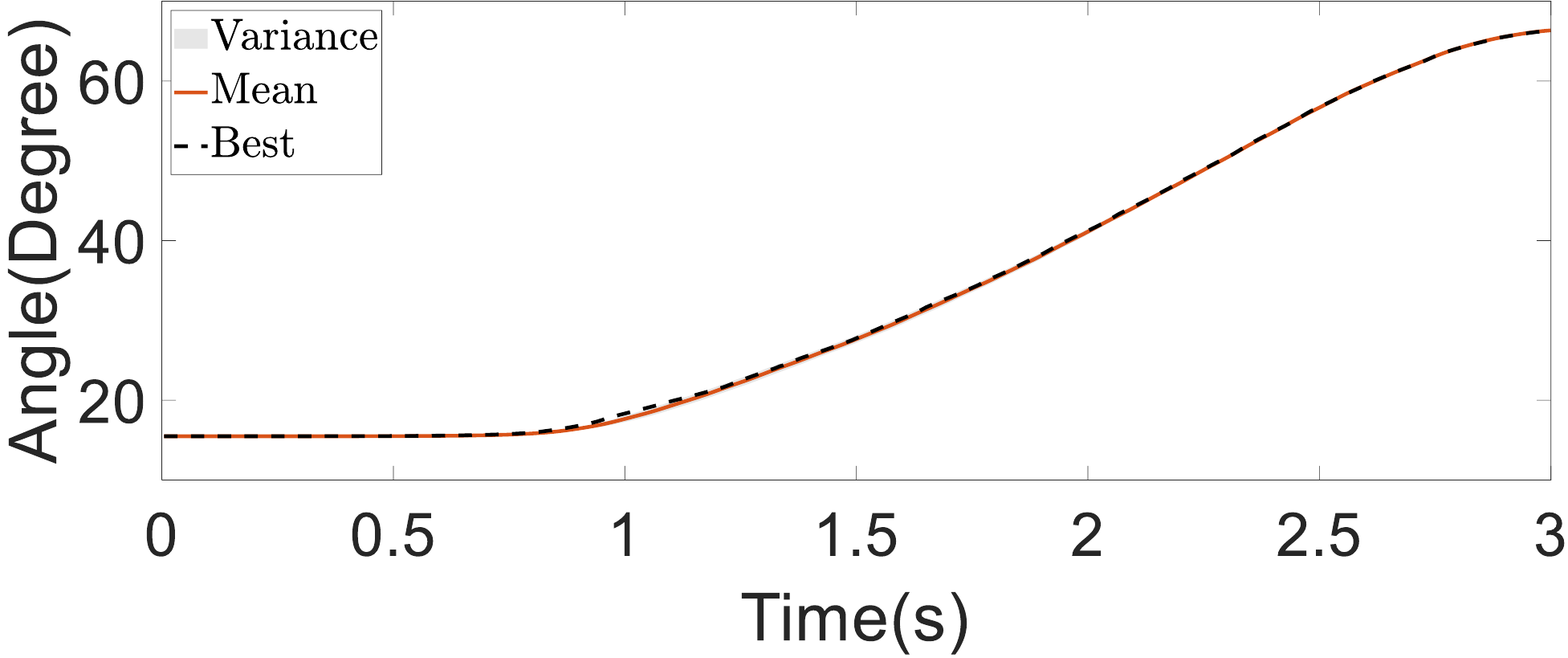}}
	\caption{Optimized assistance distribution (a)–(c) without online refinement and (d)–(f) with online refinement.}
	\label{passive_wwo}
\end{figure*}

\begin{figure}[!h] 
	\centering 
	\vspace{-0.3cm} 
	\subfigtopskip=2pt 
	\subfigbottomskip=2pt 
	\subfigcapskip=-5pt 
	\subfigure[]{
		\includegraphics[width=0.45\linewidth]{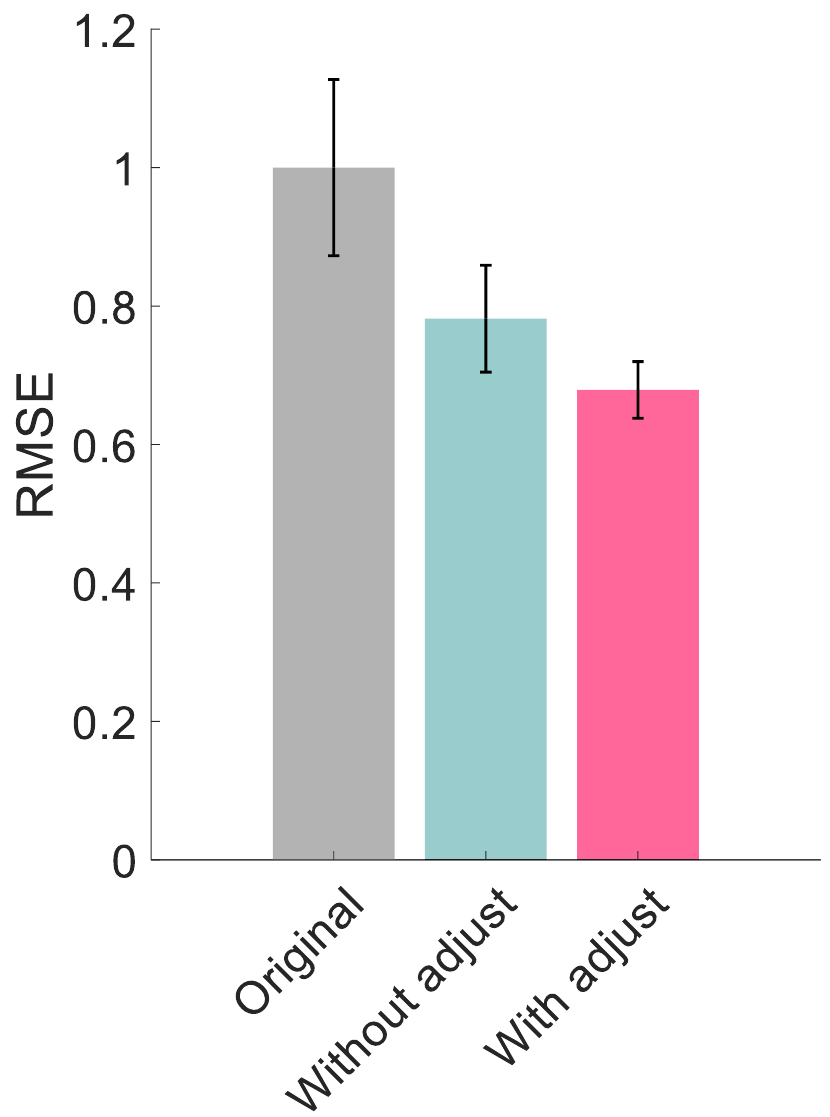}}
	\subfigure[]{
		\includegraphics[width=0.45\linewidth]{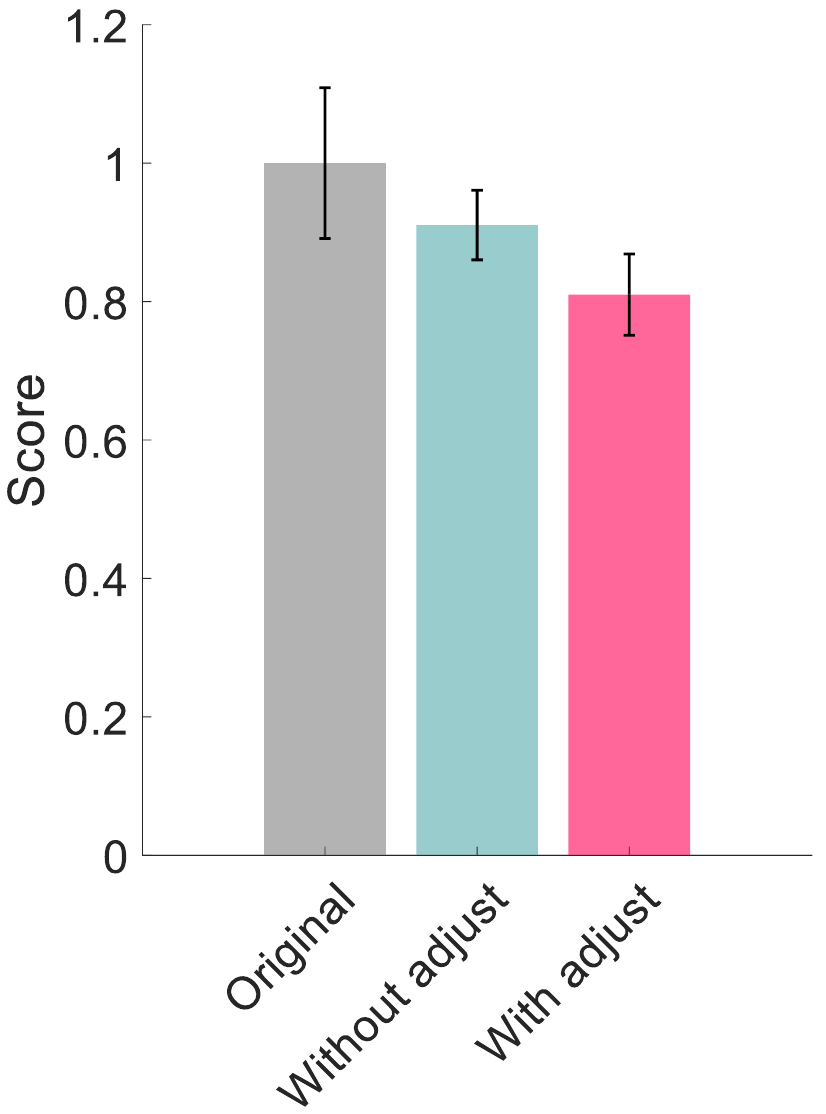}}
	\subfigure[]{
		\includegraphics[width=0.45\linewidth]{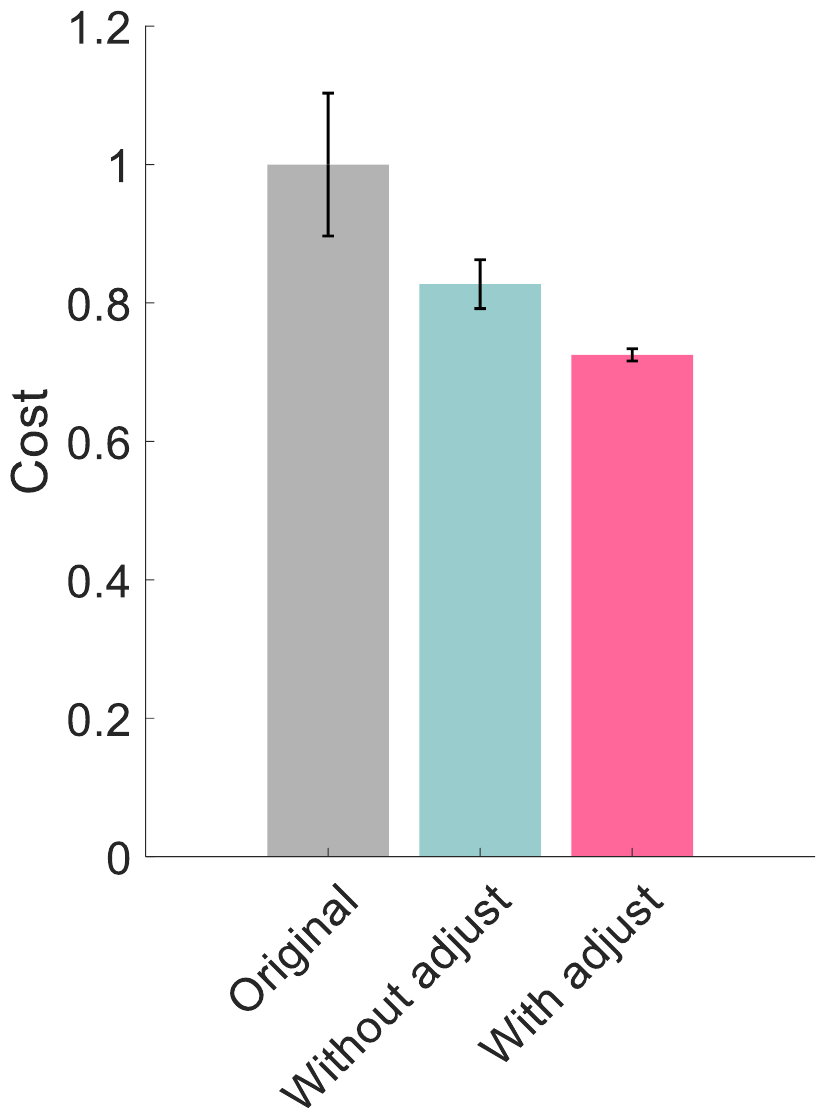}}
    \subfigure[]{
		\includegraphics[width=0.45\linewidth]{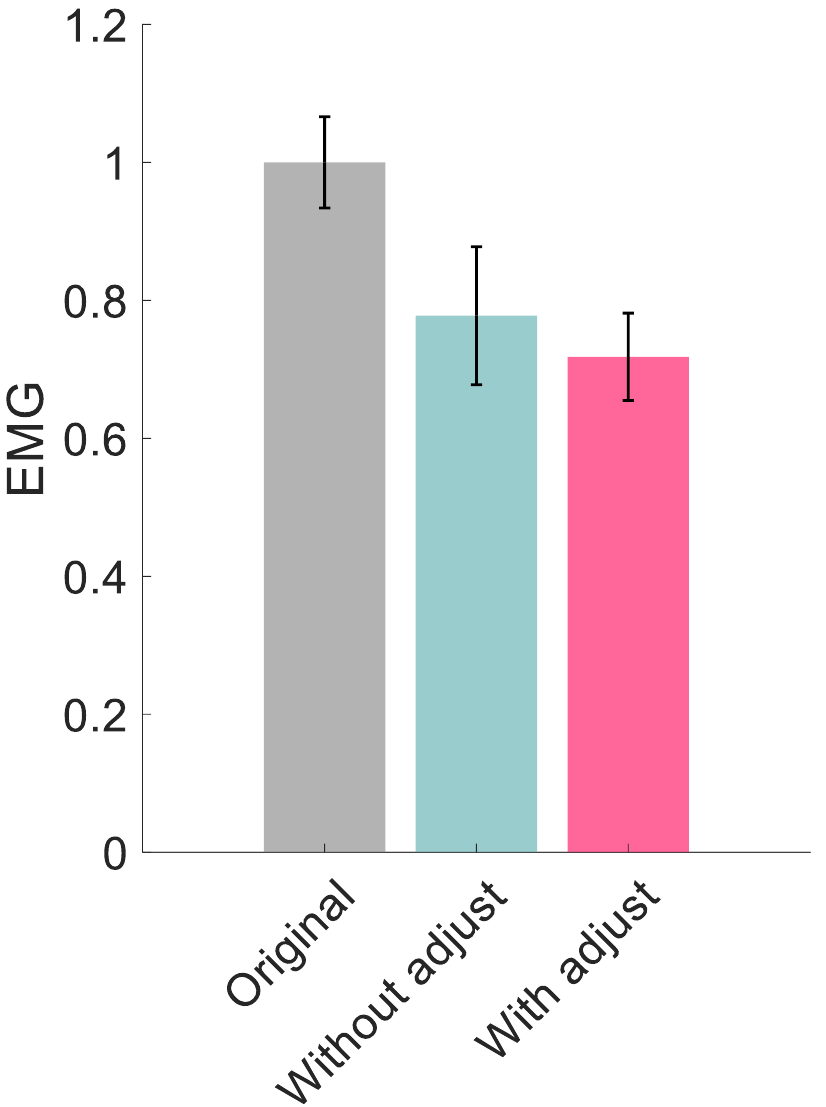}}
	\caption{(a)-(d) Evaluation results of passive following training showing the metrics: RMSE, anomaly score, cost, and EMG signal level of the biceps brachii. All metrics were averaged over one cycle and normalized relative to the original assistance.}
	\label{passive_evalua}
\end{figure}

A motion capture system must be installed in a hospital for deploying active mirroring training in clinical trials, and such a system introduces additional limitations and inconvenience.
Thus, we did not conduct clinical trials of active mirroring training. Instead, we evaluated our method by using the upper-limb exoskeleton for active mirroring training in a healthy subject by mapping the motion of the left arm (i.e., the healthy side) to that of the right arm (i.e., the mock stroke side).
Acquiring precise angles of human upper limbs through an optical motion capture system requires a custom-made suit, which was beyond the scope of this study.
Instead, we employed a self-manufactured brace embedded with key markers and gloves fitted with markers to estimate the angles at three limb joints, specifically those corresponding to Joints 1, 2, and 4.
However, due to brace deformation and marker obstruction, the accuracy of the upper-limb joint angles was limited to within a certain range.
Thus, to ensure safety and demonstrate the capability of our method to impose constraints on upper-limb movements, we set position constraints for Joints 1, 2, and 4 as $ [ -40^{\circ},10^{\circ} ] $, $ [ -10^{\circ},80^{\circ} ] $, and $[ 0^{\circ},60^{\circ} ]$, respectively.

As shown in Figure \ref{active_mirror}, the desired trajectory was consistently aligned with the trajectory of the human limb $\bm q_h$ and was refined according to dynamic constraints.
Moreover, when the unaffected side of the body moved rapidly to a position outside the established movement boundary, the proposed trajectory refinement was swiftly adjusted.
That is, a predetermined maximum speed was implemented to prevent the set joint position boundaries being exceeded. 
When the movements of the joints on the unaffected side of the body remained within these dynamic constraints, the upper-limb exoskeleton robot tracked the trajectory of the healthy limb.
Simultaneously and throughout the training session, our variable impedance controller effectively sustained the human-robot interaction within the desired impedance model.

\subsection{Passive Following Training}
We conducted a series of experiments to validate the efficiency of the proposed individualization framework on the passive following training task. 
These experiments were centered on a typical task in upper-limb rehabilitation: raising the upper limb to a fixed point.
This task requires coordination between the shoulder and elbow, specifically involving Joints 1, 2, and 4. 
During the cost calculation, the anomaly score was scaled to have the same magnitude as the tracking error, and the hyperparameter $\lambda_p$ was set to $0.003$.
Moreover, as the exploration part of assistance individualization largely consists of initial trajectory sampling based on a demonstration, we set $N_s=40$.
The training process was designed to stop when the RMSE of the mean trajectory of the distribution between two consecutive iterations decreased to less than $0.1^{\circ}$ or when the maximum number of iterations was reached.
Furthermore, the sampling space was greatly reduced by the demonstration data.
The number of samples set was considered to be sufficient for exploration in the current task based on practical experience.

First, we conducted ablation studies with a healthy patient present to validate the efficacy of the online refinement module in passive following training scenarios.
In addition, five restarts were executed to bypass local optima.
The results of these studies are depicted in Figure \ref{passive_wwo}. Within the figure, the black dashed line illustrates the trajectory that achieved the lowest cost during the exploration phase, while the red solid line represents the mean of the assistive trajectory distribution.
Without online trajectory refinement, the experiment spanned 67 iterations, whereas with online trajectory refinement, the experiment spanned 55 iterations.
The experimental results reveal that without online refinement in passive following training individualization, the converged assistive trajectory distribution exhibited a large variance, and the trajectory with the lowest cost significantly deviated from the mean of the distribution.
This indicates that the trajectory distribution with similar costs was rather wide.
In contrast, with online refinement in passive following training individualization, the converged assistive trajectory distribution had a smaller variance, suggesting that there was a more concentrated set of trajectories with similar costs.
This concentration is largely attributable to the use of sensor feedback and anomaly scores in the online refinement, thereby facilitating real-time adjustments to the assistive trajectory.
This process effectively reduced the uncertainty of motion intentions and minimized conflicts in human–robot interaction.
Thus, the precision with which the cost function evaluated and distinguished the performance of different trajectories was increased, leading to the overall effectiveness of the training being enhanced.

The effectiveness of the individualized assistance was assessed using four metrics: the tracking RMSE, the anomaly score, the cost (\ref{cost_qr}), and the EMG signal level of the biceps brachii.
The results of this evaluation are displayed in Figure \ref{passive_evalua}. 
To allow comparison, all metrics were averaged over one cycle and normalized relative to the original assistance.
Compared with the other assistance (i.e., the mean trajectory of the demonstrations, and the optimized assistance without online refinement), the assistance incorporating online refinement demonstrated the best performance in improving tracking accuracy and reducing the anomaly score during motion.
Furthermore, implementing online refinement resulted in the lowest cost following passive training and a reduction in the EMG signal level.
These outcomes substantiate the efficacy of the proposed individualization framework for passive following training.

\begin{table*}
\centering
\caption{Attributes of Participants}
\begin{tblr}{
  cell{2}{1} = {r=3}{},
  cell{5}{1} = {r=3}{},
  vline{2} = {1-7}{},
  hline{1,8} = {-}{0.08em},
  hline{2,5} = {-}{},
}
Group        & Subject & Gender & Age(y) & Weight(kg) & Height(cm) & Arm length(cm) & Diagnosis           \\
Control      & 1       & Male   & 46     & 70         & 176        & 61             & Cerebral hemorrhage \\
             & 2       & Male   & 60     & 65         & 167        & 57             & Stroke              \\
             & 3       & Male   & 62     & 75         & 170        & 58             & Stroke              \\
Experimental & 4       & Male   & 44     & 67         & 169        & 59             & Stroke              \\
             & 5       & Male   & 51     & 77         & 177        & 64             & Stroke              \\
             & 6       & Male   & 66     & 66         & 167        & 57             & Stroke              
\end{tblr}
\label{patient_attr}
\end{table*}

We also recruited seven patients for clinical trials to demonstrate the effectiveness of the proposed method.
As one participant was transferred to another hospital, only six participants completed the trial. All the patients had a healthy side that did not move naturally, due to the effects of their stroke. Hence, passive following training was employed. Each participant signed an informed consent form, and all experiments were approved by the ethics committee created by Shenzhen MileBot Robotics Co., Ltd in May 2023.
The participants were allocated to either a control group or an experimental group and their details are summarized in Table \ref{patient_attr}.
Both groups engaged in regular daily rehabilitation exercises, and the experimental group participated in an additional 14 days of passive following training.

\begin{figure}[!h] 
	\centering 
	\vspace{-0.3cm} 
	\subfigtopskip=2pt 
	\subfigbottomskip=2pt 
	\subfigcapskip=-5pt 
	\subfigure[]{
		\includegraphics[width=0.3\linewidth]{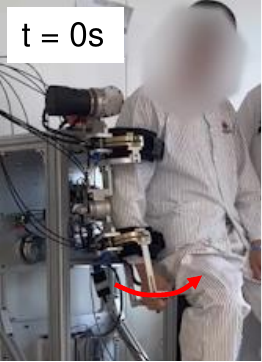}}
	\subfigure[]{
		\includegraphics[width=0.3\linewidth]{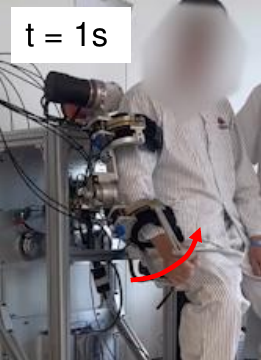}}
	\subfigure[]{
		\includegraphics[width=0.3\linewidth]{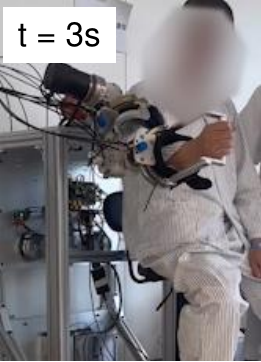}}
    \subfigure[]{
		\includegraphics[width=0.3\linewidth]{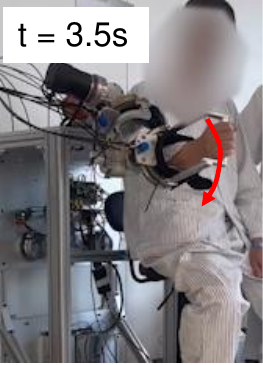}}
	\subfigure[]{
		\includegraphics[width=0.3\linewidth]{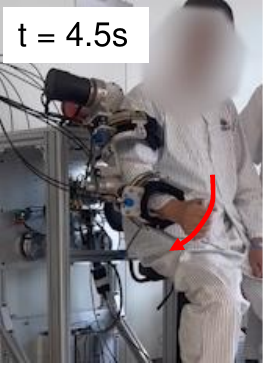}}
	\subfigure[]{
		\includegraphics[width=0.3\linewidth]{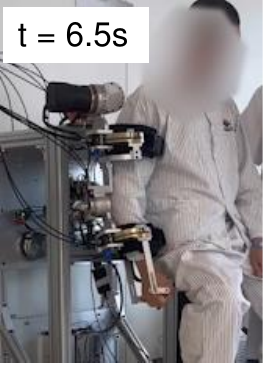}}
    \subfigure[]{
		\includegraphics[width=0.95\linewidth]{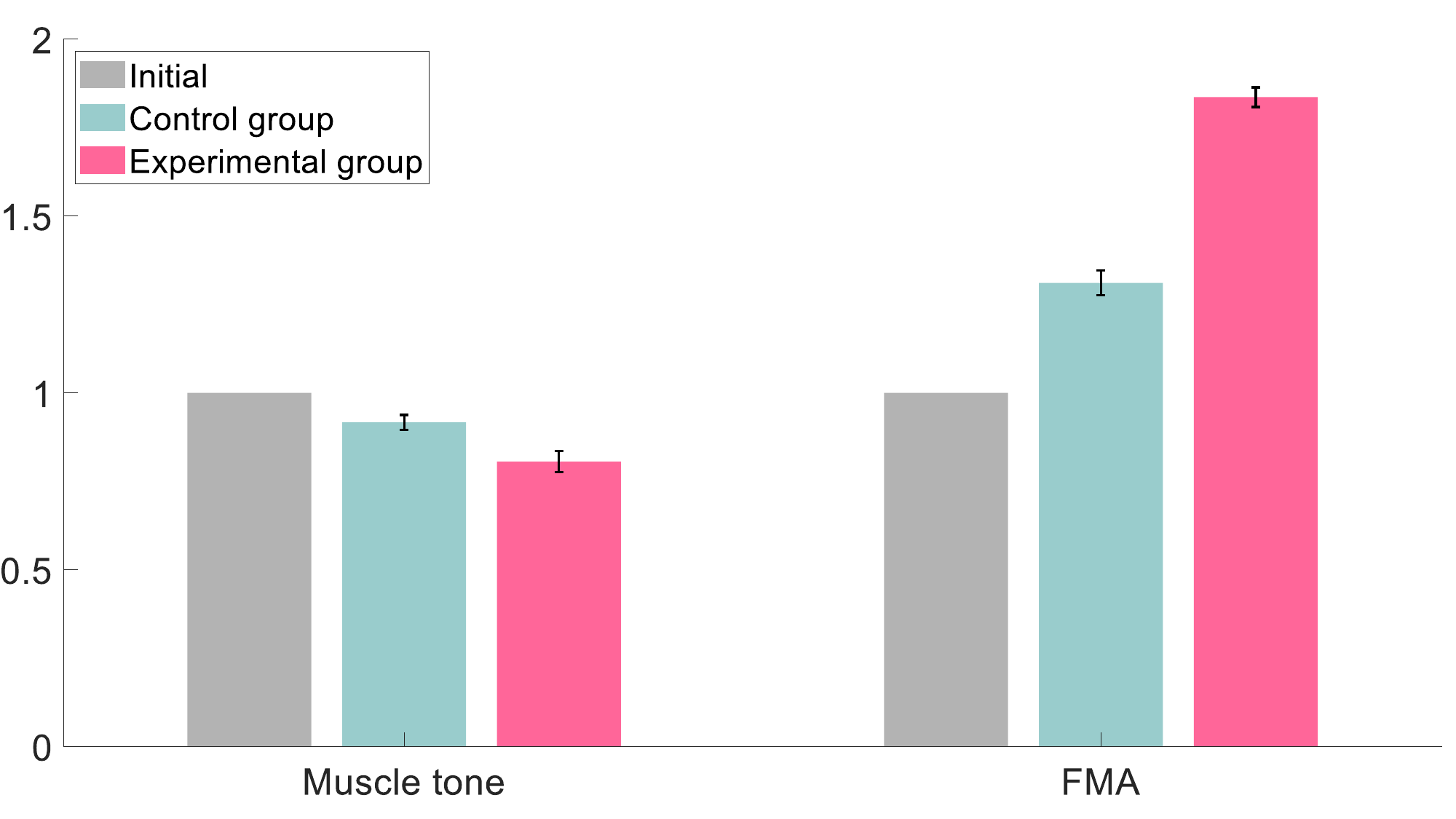}}
	\caption{(a)–(f) Snapshots of upper-limb movement during
training, with red arrows indicating the direction of movement;
(g) comparative results of the clinical trial, normalized to the initial evaluation.}
	\label{clinical_res}
\end{figure}

The rehabilitation task was set to be the same as the previously described task of raising the arm, involving the coordinated movement of Joints 1, 2, and 4, as illustrated in Figure \ref{clinical_res}.
Interactive information was recorded during the limb-lifting phase, which was considered as the training task, while the limb-lowering phase, which began at $t=3.5s$, was used to return to the initial position.
The motor abilities of both groups were evaluated and scored by professional healthcare personnel using specific evaluation metrics. These metrics were the muscle tone level and the Fugl-Meyer assessment (FMA) score \citep{fugl1975method}.
A low level of muscle tone and a high FMA score are indicative of good upper-limb motor ability.
Both groups were subjected to a motor function assessment before treatment. During the treatment phase, the experimental group participated in daily passive following training sessions, each of which lasted approximately 15 minutes.
Both groups underwent reassessment 2 weeks after the start of the treatment.
The experimental results normalized to the initial evaluation are displayed in Figure \ref{clinical_res}, and detailed evaluation results are included in the Appendix. 
Unlike the control group, the experimental group exhibited significant improvements in all metrics compared with their initial assessments before the treatment. 
Thus, compared with the results of the control group, the results of the experimental group indicate that the passive following training with the upper-limb exoskeleton robot accelerated the recovery of motor functions.
Therefore, this training could enhance the effectiveness of treatment for conditions such as stroke and cerebral hemorrhage.

\section{Conclusion and Discussion}
\subsection{Conclusion}
\noindent\textbf{Overall}: This paper introduces a dual-mode individualization framework that incorporates generative models. This framework incorporates an intention predictor and an anomaly detector, which are used to capture the motion intentions of the unaffected side of the patient and to assess the human–robot interaction in real time during rehabilitation tasks.
In active mirroring mode, the assistance reflects the patient’s original motion intentions. In passive following mode, the assistance is tailored to the patient based on interactive feedback.
Trajectories in both modes are integrated within an online trajectory refinement framework, ensuring that they are smooth, adhere to dynamic constraints, and are individualized, thereby effectively supporting the patient’s rehabilitation.

\noindent\textbf{Details}:
The online trajectory refinement integrates both training configurations and utilizes generative models to achieve personalized assistance.
Specifically, in active mirroring mode, the reference trajectory is derived from the unaffected limb, with the intention predictor providing a predicted trajectory distribution that is preemptively tuned to mitigate potential risk movements.
Conversely, in passive following mode, the reference trajectory is pre-defined based on human demonstrations.
Additionally, the anomaly detector plays a crucial role in guiding the online refinement process to enhance the naturalness of movements in real time.
This detector assesses the deviation between the current interaction data and standard demonstration data obtained from healthy individuals, thereby facilitating performance evaluation during passive following training.
In passive following mode, ProMPs are implemented for specific training tasks, with each movement of the patient weighted according to a cost function. This approach significantly enhances the effectiveness of the generated assistance distribution.

\noindent\textbf{Performance}:
We conducted a series of experiments, including a clinical trial,  to validate each of the proposed modules and demonstrate their effectiveness in enhancing assistance and ensuring safety.
In terms of prediction accuracy, the intention predictor outperformed alternative methods, namely forward integration and a CNN-LSTM.
Furthermore, the anomaly detector accurately identified anomalies across different scenarios. 
Moreover, the performance of the variable impedance controller was validated in trajectory tracking and in assisting the patient when anomalies occurred. 
During active mirroring training, online refinement effectively reduced the degree of constraint violations in the presence of unexpected impacts.
It was also capable of identifying abnormal regions within the movement space and guiding the upper-limb exoskeleton robot to result in a decreased anomaly score. 
This active mirroring training approach was tested under a motion capture system, which validated its effectiveness.
In passive following mode, testing was conducted using healthy individuals and in a clinical trial, respectively. The results confirm that the approach provided personalized assistance to healthy participants and significantly accelerated the recovery of motor functions in stroke participants.
Specifically, the clinical trial data indicate that the experimental group, which had participated in passive following training, showed improvements in various performance metrics after completing the treatment protocol.

\subsection{Discussion}
\noindent\textbf{Limitations}: 
The current trajectory generation method exhibits three main limitations, as detailed below.
\begin{enumerate}
\item[1)] The performance of the intention predictor and anomaly detector depends on the size and quality of the dataset. Expanding the dataset to include more subjects would significantly improve the performance of the generative models. This would increase the accuracy of predictions of patient motion intentions and the precision of detection of abnormal interactions during movements, thereby enhancing the personalization of the training modes.
\item[2)] The clinical trial included only six participants and focused exclusively on the rehabilitation effects of the passive following mode. Conducting a clinical trial with more participants and incorporating active mirroring mode into the rehabilitation process would provide a more comprehensive evaluation of the proposed dual-mode individualization framework.
\end{enumerate}
Efforts to address these limitations in the manner described will form the basis of our future research and development activities.

\noindent\textbf{Intellectual Merits}: 
We have developed an innovative dual-mode individualization framework that incorporates generative models, thereby establishing a new benchmark for adaptive rehabilitation systems.
This novel framework can switch between active mirroring and passive following modes based on the patient’s needs and thus offers personalized assistance and enhanced rehabilitation outcomes.
Key features of this framework are its real-time intention prediction and anomaly detection capabilities.
Specifically, the intention predictor captures motion intentions from the unaffected side of the patient, while the anomaly detector evaluates human–robot interactions in real time, ensuring immediate adaptation and response to the patient’s movements.
Additionally, the framework integrates online trajectory refinement that unifies trajectories from both active mirroring and passive following modes to ensure they are smooth, dynamically constrained, and individualized.
Thus, the framework provides more natural and effective assistance than other frameworks. The application of generative models to personalize assistance based on interactive feedback ensures that the rehabilitation process is effectively responsive to individual patient conditions and needs.

\noindent\textbf{Potential Impacts}:
The development of a dual-mode individualization framework that integrates generative models represents a significant advancement that could enhance the deployment and effectiveness of rehabilitation technologies in both clinical and homecare environments. 
Specifically, as this innovative framework delivers personalized and adaptive assistance tailored to real-time feedback and the specific motion intentions of the patient, it has the potential to revolutionize the rehabilitation process.
To the best of the authors’ knowledge, this study is the first to integrate generative models into an upper-limb exoskeleton robot and perform a clinical trial. 
Our pioneering approach not only enhances the functionality of rehabilitation devices but also contributes to a potential impact on the field by merging artificial intelligence with rehabilitation medicine.
That is, our approach could effectively bridge the gap between AI and rehabilitation medicine, thereby facilitating the translation of advancements in AI into practical medical applications. 
This study exemplified the power of interdisciplinary research, as it involved a combination of principles from the fields of robotics, control systems, machine learning, and clinical rehabilitation.
This led to advances in each field and set a precedent for future studies aiming to develop comprehensive and adaptive rehabilitation systems.
Furthermore, the framework devised in this study addresses the broader societal challenge posed by an aging population. Specifically, the framework offers methods that could be used in advanced rehabilitation solutions that are applicable in both healthcare facilities and home settings.

\section{Funding}
This work was supported in part by the Science and Technology Innovation 2030-Key Project under Grant 2021ZD0201404, in part by
the Institute for Guo Qiang, Tsinghua University, and in part by the National Natural Science Foundation of China under Grant U21A20517 and 52075290. 








\bibliographystyle{SageH}
\bibliography{ref/ref}

\begin{table*}
\centering
\caption{Detailed Clinical Evaluation}
\begin{tblr}{
  cell{2}{1} = {r=3}{},
  cell{5}{1} = {r=3}{},
  vline{2} = {1-7}{},
  hline{1,8} = {-}{0.08em},
  hline{2,5} = {-}{},
}
Group        & Subject & Initial muscle tone level & Post-trial muscle tone level & Initial FMA & Post-trial FMA \\
Control      & 1       & 1.5                       & 1.5                          & 8           & 12             \\
             & 2       & 1.5                       & 1.5                          & 31          & 35             \\
             & 3       & 2                         & 1.5                          & 23          & 30             \\
Experimental & 4       & 1.5                       & 1                            & 8           & 16             \\
             & 5       & 2                         & 1.5                          & 19          & 35             \\
             & 6       & 2                         & 2                            & 3           & 5              
\end{tblr}
\label{clinical_table}
\end{table*}
\section{Appendix}
\subsection{Stability Analysis}
To prove the stability of the whole system, we substitute equations (\ref{vectorz}) and (\ref{slow_controller}) into (\ref{substitute_SP4sec}), resulting in the following system dynamics:
\begin{align}
(\bm M(\bm q)+\bar{\bm B})\dot{\bm z}+\bm C(\dot{\bm q}, \bm q)\bm z=-\bm K_z\bm z-\bm S_2^{\mathsf{T}}\Tilde{\bm\tau}_f-k_g{\bm{sgn}}(\bm z),\label{global_stability_pre}
\end{align}
where $\Tilde{\bm\tau}_f = \hat{\bm\tau}_f - {\bm\tau}_f$ represents the estimation error of the friction.

Subsequently, we propose the following candidate Lyapunov function:
\begin{eqnarray}
&V = \frac{1}{2}\bm z^T(\bm M(\bm q)+\bar{\bm B})\bm z.
\label{Lyapunov_candidate}
\end{eqnarray}

By differentiating (\ref{Lyapunov_candidate}) with respect to time and substituting the dynamics with those from (\ref{global_stability_pre}), we derive the following expression:
\begin{eqnarray}
\begin{array}{*{20}{l}}
\dot V = -\bm z^T\bm K_z\bm z - \bm z^T\bm S_2^{\mathsf{T}}\Tilde{{\bm\tau}}_f-k_g\bm z^T\bm{sgn}(\bm z).
\end{array}
\label{diff_Lyapunov_candidate}
\end{eqnarray}

Assuming that $\Vert \bm S_2^{\mathsf{T}}\Tilde{{\bm\tau}}_f \Vert \hspace{-0.05cm} \leq \hspace{-0.05cm} \kappa$, the upper bound for $\dot V$ is derived as follows:
\begin{align}
\dot V \leq& -\bm z^T\bm K_z\bm z -(k_g-\kappa)\Vert \bm z \Vert,
\end{align}

If $k_g$ is adequately large such that $k_g>\kappa$, the inequality simplifies to
\begin{align}
\dot V \leq& -\bm z^T\bm K_z\bm z<0,
\end{align}

Given that $V>0$ and $\dot V < 0$, the \textit{quasi-steady-state system} is exponentially stable. 
Considering that the \textit{boundary-layer system} can be made intrinsically stable by appropriate tuning of $\bm K_1$ and $\bm K_2$, the stability of the closed-loop system is assured according to \cite{tikhonov1952systems}, ensuring convergence to the desired impedance vector.

\subsection{Weight Setting in Passive Following Mode}

For a specific training task involving human-robot interaction, we assume that the patient intends to synchronize movement with the upper-limb exoskeleton robot.
However, physical factors such as the randomness of human motion intentions may impact the tracking performance of the upper-limb exoskeleton robot.
Given the challenge of explicitly accounting for such disturbances within the dynamics, we model this interference as additional noise in trajectory planning. The dynamics are described as follows:
\begin{align}
    \bm x_p^{(t+1)} &= g_p(\bm x_p^{(t)},\bm q_{r}),\label{sample_dynamic}\\
    \bm q_{r} &\sim \mathcal{N}(\bm q_{rp},\bm \Sigma_p),
\end{align}
where $\bm x_p = [\bm x_d^{\mathsf{T}},s]^{\mathsf{T}}$ is an augmented state vector, $g_p(\cdot)$ is a nonlinear time-variant function that integrates online refinement, the impedance controller, and deterministic components of human motion intention with human-robot interaction. 
$\bm q_{rp}$ is the mean of the assistance distribution, and $\Sigma_p$ encapsulates the overall stochastic disturbance, which includes the randomness of human motion intentions and the sampling variability of generative models.

According to \cite{williams2017information}, to minimize the following cost
\begin{align}
\hat{\mathcal{S}}(\bm q_{r}) = \int_{T_r} \{\|\bm q_{d}(\bm q_{r}) - \bm q\|_{\bm Q}^2+s^2 + \frac{\lambda_p}{2}\bm q_{rp}^{\mathsf{T}}\Sigma_p^{-1}\bm q_{rp}\}dt,
\label{new_cost}
\end{align}
where the optimal control input is structured in a cost-decoupled manner as follows:
\begin{align}
\label{optimal_weight}
\hat{w}(\bm q_{r}) &= \frac{1}{\hat{\eta}_k}\exp(-\frac{1}{\lambda_p}(\hat{\mathcal{S}}(\bm q_r) + \gamma_p\sum \Tilde{\bm q}_{rp}^{\mathsf{T}}\Sigma_p^{-1}\bm q_{r})), \\
\hat{\bm q}_{rp}^* &= \mathbb{E}[\hat{w}(\bm q_{r})\bm q_{r}].
\end{align}
Here, $\gamma_p = \lambda_p(1-\alpha_p)$ is the decoupled temperature parameter, defined with $\alpha_p\in[0,1]$, and $\Tilde{\bm q}_{rp}$ is the mean of the initial trajectory estimates.

In this approach, the optimal assistance is derived from the expected value of the current trajectory distribution.
Each sampling iteration is based on the previously improved trajectory distribution, which may lead to inconsistencies in the sampling space.
To address these potential inconsistencies and ensure adequate exploration, we opt for a large sample size $N_s$ in Algorithm \ref{ProMP_ALG}.
Additionally, we set $\alpha_p = 1$ to mitigate the influence of inconsistencies in $\Tilde{\bm q}_{rp}$ across different samples and thereby enhance the robustness and reliability of the optimal control solution.

Given that the last term in (\ref{new_cost}) is unexpected in the rehabilitation process, the ideal set is $\lambda_p = 0$, leading to the following definition of optimal assistance:
\begin{align}
\hat{\bm q}_{rp}^* &= \mathop{\arg\min}\limits_{\bm q_r}\hat{\mathcal{S}}(\bm q_{r}).
\end{align}
In this scenario, the optimal assistance is identified as the sampled trajectory with the lowest cost, and all other sampled outcomes are disregarded. 
However, this formulation disregards the distribution of the optimal assistance, rendering the iterative improvement mechanism nonviable.
Thus, to maintain the feasibility of iterative improvements and also minimize the impact of the undesirable term, the parameter $\lambda_p$, which governs the tightness of the solution, is set to a small value. 
This adjustment ensures that $\hat{\mathcal{S}}(\bm q_{r}) \rightarrow {\mathcal{S}}(\bm q_{r})$, and the optimal weight formula in (\ref{optimal_weight}) simplifies to the configuration used in (\ref{promp_weight}).
This setting balances the need to minimize the undesired term with the need to maintain a practical and effective iterative improvement process.

\subsection{Clinical Evaluation Results}
In this clinical trial, muscle tone levels were assessed using a rating scale with grades 0, 1, 1+, 2, 3, and 4, where grade 0 indicates normal muscle tone. 
For the purposes of numerical analysis, grade 1+ is quantified as 1.5. The evaluation results are documented in Table \ref{clinical_table}.
All assessments were exclusively focused on the upper limb. In the upper limb segment of the FMA, the maximum score, indicating normal function, is 66 points.

\end{document}